\documentclass[10pt,twocolumn,letterpaper]{article}

\usepackage{cvpr}              %

\usepackage[dvipsnames]{xcolor}

\usepackage{graphicx}
\usepackage{multirow}
\usepackage{pifont} %
\usepackage{xcolor}
\usepackage{colortbl}
\usepackage{amsmath,amssymb,stmaryrd}   
\usepackage{siunitx}
\usepackage{array}
\usepackage{subcaption}
\usepackage{tabularx}
\usepackage{wrapfig}
\usepackage{ifthen}
\usepackage{listings}

\usepackage{xspace}
\usepackage{algorithmic}
\usepackage{algorithm}
\usepackage{tabularx}
\usepackage{mathrsfs}

\usepackage{bibunits}

\definecolor{cvprblue}{rgb}{0.21,0.49,0.74}
\usepackage[pagebackref,breaklinks,colorlinks,citecolor=cvprblue]{hyperref}

\usepackage[capitalize]{cleveref}
\crefname{section}{Sec.}{Secs.}
\Crefname{section}{Section}{Sections}
\Crefname{table}{Table}{Tables}
\crefname{table}{Tab.}{Tabs.}

\newenvironment{packed_itemize}
{\begin{itemize}
    \setlength{\itemsep}{1pt}
    \setlength{\parskip}{0pt}
    \setlength{\parsep}{0pt}
}{\end{itemize}}

\newcommand{\filluptopage}[1]{%
  \clearpage
  \loop\ifnum\value{page}<#1\relax
    \null\clearpage
  \repeat
  \loop\ifnum\value{page}=#1\relax
    \null\clearpage
  \repeat
}

\makeatletter
\def\blfootnote{\xdef\@thefnmark{}\@footnotetext}
\makeatother

\usepackage[accsupp]{axessibility}

\newcommand{\rowA}{\rowcolor{gray!15}}
\newcolumntype{M}[1]{>{\centering\arraybackslash}p{#1}}
\definecolor{myblue}{RGB}{1,137,157}
\definecolor{mygreen}{RGB}{177,196,77}
\definecolor{mypink}{RGB}{178,34,34}
\definecolor{mypurple}{RGB}{102,153,255}

\def\paperID{3287} %
\def\confName{CVPR}
\def\confYear{2025}

\newcommand{\methodname}{GigaHands\xspace}
\title{\methodname: A Massive Annotated Dataset of Bimanual Hand Activities}

\author{Rao Fu$^{1 *}$\quad Dingxi Zhang$^{2 *}$ \quad Alex Jiang$^{1}$ \quad Wanjia Fu$^{1}$ \\
\quad Austin Funk$^{1}$ \quad Daniel Ritchie$^{1}$ \quad Srinath Sridhar$^{1 \dag}$ \\
\small{$^{*}$ Equal contribution\quad $^{\dag}$ Corresponding author}\\
$^{1}$Brown University \quad
$^{2}$ETH Zurich\\
\small{\url{https://ivl.cs.brown.edu/research/gigahands.html}}
}

\begin{document}
\twocolumn[{
\maketitle

\centering
\vspace{-0.3in}
\makebox[\textwidth][c]{\includegraphics[width=1.06\linewidth]{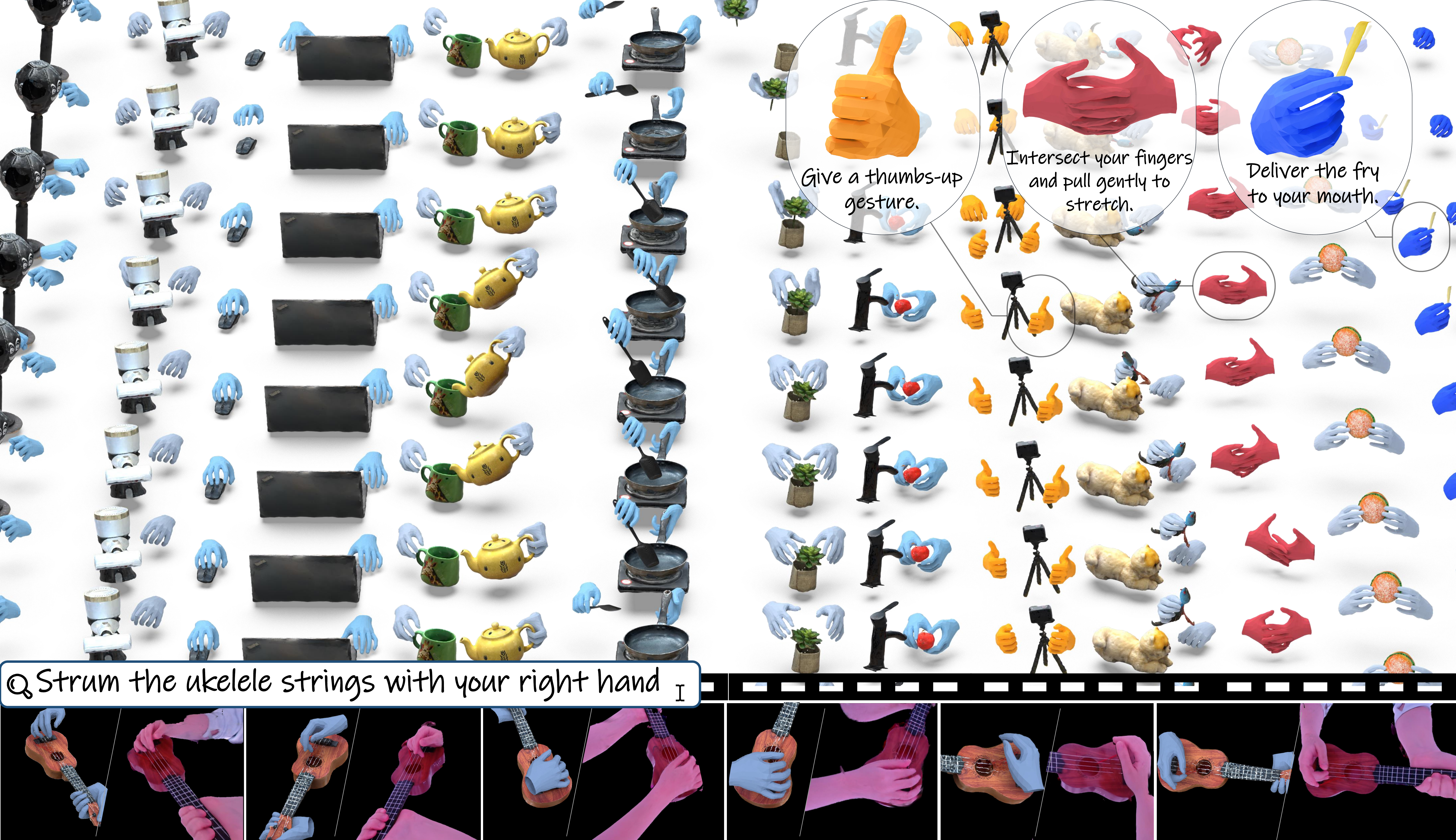}}
\vspace{-0.2in}
\captionof{figure}{\textbf{\methodname} is a massive dataset of human bimanual activities with paired text annotations.
Each column above shows an activity sequence from the dataset.
The dataset covers diverse 3D hand activities, including hand-object interactions (\textcolor{blue}{blue}) across object scales, gestures (\textcolor{orange}{orange}), and self-interactions (\textcolor{red}{red}). Each clip is paired with descriptive text and 51 camera views, enabling radiance field reconstruction.
The bottom row show other annotations in the dataset including hand shape, object shape and pose (left half images).
The right half images show novel views from dynamic radiance field fitting.
\vspace{0.15in}
}
\label{fig:teaser}
}]

\begin{abstract}
\vspace{-1em}
Understanding bimanual human hand activities is a critical problem in AI and robotics.
We cannot build large models of bimanual activities because existing datasets lack the scale, coverage of diverse hand activities, and detailed annotations.
We introduce \methodname, a massive annotated dataset capturing 34 hours of bimanual hand activities from 56 subjects and 417 objects, totaling 14k motion clips derived from 183 million frames paired with 84k text annotations.
Our markerless capture setup and data acquisition protocol enable fully automatic 3D hand and object estimation while minimizing the effort required for text annotation.
The scale and diversity of \methodname enable broad applications, including text-driven action synthesis, hand motion captioning, and dynamic radiance field reconstruction. 
\end{abstract}
\vspace{-0.2in}
    
\vspace{-1em}
\section{Introduction}
\label{sec:intro}
\noindent\emph{``The \textbf{human hand} is a marvel of evolution, whose intricate structures and capabilities have allowed us to manipulate the environment in ways no other species can.''} 
\begin{flushright}
-- F.R. Wilson~\cite{wilson1999hand}
\end{flushright}
From the skillful manipulation involved in cooking a meal, to the rapid movement of fingers to type this sentence, hands are always busy shaping our environments and communicating with others.
Enabling machines to understand and replicate the remarkable capabilities of human hands is one of the grand challenges of AI and robotics.
One approach to achieve this goal could be to train models on massive datasets, the strategy followed by recent large models that learn from trillions of text tokens~\cite{dubey2024llama}, 
billions of text-image pairs~\cite{schuhmann2022laion},
tens of millions of 3D shapes~\cite{deitke2024objaverse},
or millions of robot trajectories~\cite{o2023open,brohan2022rt,brohan2023rt,kim2024openvla}.
However, datasets of this scale are currently unavailable for natural hand activities.

Sourcing large-scale 3D hand activities data is challenging.
The two most common methods for data acquisition involve using cameras to capture hand manipulations in the wild or in controlled studio settings.
In-the-wild data includes monocular internet videos~\cite{shan2020understanding}, egocentric videos captured using wearable cameras~\cite{pirsiavash2012detecting,damen2018scaling,li2021eye,damen2022rescaling,grauman2022ego4d,liu2022hoi4d}, or multi-view videos from third-person cameras~\cite{goyal2017something,miech2019howto100m,shan2020understanding}.
However, this data is sparse, hard to calibrate, and noisy, resulting in limited 3D motion reconstruction accuracy, especially for objects.
Alternatively, studio settings bring subjects into camera-rich environments and could employ markers for accurate reconstruction.
However, the staged setup and lack of real-world context limits data diversity, and marker-based tracking~\cite{liu2022beat,garcia2018first,taheri2020grab,yang2022oakink,fan2023arctic,liu2024taco,zhan2024oakink2,banerjee2024hot3d,lv2025himo} inhibits natural interactions.

We address the data sourcing problem preventing the scaling up of hand datasets by introducing \textbf{\methodname}, a diverse, massive, and fully-annotated 3D bimanual hand activities dataset (see \Cref{fig:teaser}).
To our knowledge, \methodname is the largest bimanual hand activities dataset, with over \textbf{183 million} unique image frames with two hands each (0.37$\times$10$^9$ unique hand poses, hence \underline{Giga}Hands).
We used a multi-camera markerless capture system to acquire accurate bimanual hand-object interaction activities, gestures, and self-contacts. We overcome the studio capturing limitation by designing procedural activity elicitation protocols to include as many actions as in the real-world in-the-wild setting.
To ensure diversity that reflects the real world, we captured data from \textbf{56 subjects} interacting with \textbf{417 real-world objects}.
All images in our dataset are fully annotated with: detailed activity text descriptions; 3D hand shape and pose; MANO~\cite{romero2022embodied} hand meshes; 3D object shape, pose and appearance; hand/object segmentation masks; 2D/3D hand keypoints; camera pose.
We minimized manual annotation effort using a procedural \emph{instruct-to-annotate} strategy by guiding subjects with detailed instructions to reduce annotation effort post-capture.
In total, we collected \textbf{34 hours} of text-annotated bimanual hand activities, \textbf{13k} 3D motion sequences, and over \textbf{14k} post-processed 3D motion clips
with \textbf{84k} detailed atomic-level text description annotations.
Our 3D hand motion clips exceed the combined size of all existing 3D bimanual hand activities datasets~\cite{liu2022hoi4d,liu2024taco,zhan2024oakink2}, and the range of verbs in our text annotations is larger than any other hand dataset, even those captured in the wild~\cite{grauman2022ego4d,grauman2024ego}.

\methodname can unlock new capabilities in applications including motion generation, robotics, and dynamic 3D reconstruction~\cite{pumarola2021d}.
To demonstrate, we show improvements enabled by our dataset on text-driven action synthesis and hand motion captioning, along with examples dynamic radiance field reconstruction.
To sum up our contributions:
\begin{packed_itemize}
    \item We present \textbf{\methodname}, a massive, diverse and annotated 3D bimanual activities dataset.
    It includes \textbf{34 hours} of activities, \textbf{14k} hand motions clips paired with \textbf{84k text annotation}, and over \textbf{183M} unique hand images.
    \item A \textbf{procedural data acquisition} strategy that ensures activity diversity and detailed atomic-level text descriptions while minimizing manual effort.
    We also built an accurate and fully-automated pipeline for estimating 3D hand/object shape and pose, segmentation, and camera pose.
    \item We show \textbf{applications} enabled by our dataset's scale, including 3D hand motion generation, hand motion captioning,
    and dynamic semantic scene reconstruction.
\end{packed_itemize}

\section{Related Work}

\subsection{Hand Motion Data Sourcing}
Hand motion data has four primary sources, and each presents its own advantages and challenges concerning realism, diversity, accuracy, and practicality.

\noindent\textbf{Static poses} involve motion planning \cite{amor2012generalization,bai2014dexterous,brahmbhatt2019contactgrasp,she2022learning}, policy learning \cite{mandikal2021learning,christen2022d}, or motion synthesis \cite{zhang2024artigrasp,christen2024diffh2o,zhang2024graspxl} from individual static positions.
While these methods provide reliable 3D data, it lacks semantic richness,
resulting in motions disconnected from real-world activities and not fully representing diverse human movement. 

\noindent\textbf{Synthetic data} from game engines \cite{zimmermann2017learning}, augmented reality \cite{mueller2017real}, virtual reality~\cite{han2022umetrack}, or simulations \cite{miller2004graspit,hasson2019learning,turpin2022grasp} allow for controlled and accurate 3D motion sequences. Despite their reliability, these synthetic motions often fail to capture the intricacies of genuine human movements, limiting their applicability in modeling natural hand behaviors. 

\noindent\textbf{Real-world in-the-wild settings} involves using wearable or portable sensors to capture hand interactions from egocentric \cite{pirsiavash2012detecting,damen2018scaling,li2021eye,damen2022rescaling,grauman2022ego4d,liu2022hoi4d}, third-person \cite{goyal2017something,miech2019howto100m,shan2020understanding} or both types of views~\cite{sigurdsson2018charades,grauman2024ego}, yielding realistic and natural motions.
However, obtaining reliable 3D motion data for both hands and objects is challenging due to occlusions resulting from limited sensor deployment and accuracy.
Additionally, annotating such data is labor-intensive, and certain self-contact motions are difficult to record accurately. 

\noindent\textbf{Studio environments} place participants in controlled settings equipped with extensive sensors, enabling reliable capture of detailed information such as 3D hand motions \cite{joo2018total,zhang2022egobody,liu2022beat,ohkawa2023assemblyhands,lin2024motion,zhang2024both2hands}, object movements \cite{garcia2018first,hampali2020honnotate,chao2021dexycb,kwon2021h2o,hampali2022keypoint,yang2022oakink,banerjee2024hot3d,liu2024taco,zhan2024oakink2,lv2025himo}, contact regions \cite{taheri2020grab,fan2023arctic,pokhariya2024manus}, audio \cite{lee2019talking,yoon2022genea,jin2024audio,ng2024audio}, and tactile data \cite{buscher2015flexible,pham2017hand,sundaram2019learning,grady2022pressurevision}. However, collecting data in studios is arduous and may not reflect the diversity of real-life scenarios. Participants might find it challenging to perform natural motions in an unfamiliar environment, especially when encumbered with wearable sensors. For example, motion capture (MoCap) systems \cite{liu2022beat,garcia2018first,taheri2020grab,yang2022oakink,fan2023arctic,liu2024taco,zhan2024oakink2,banerjee2024hot3d,lv2025himo} require markers attached to the body, reducing visual realism and inhibiting movements like self-contact between hands, necessitating additional post-processing to restore a realistic appearance. On the other hand, markerless motion capture systems \cite{joo2018total,hampali2020honnotate,chao2021dexycb,kwon2021h2o,hampali2022keypoint,zhang2022egobody,ohkawa2023assemblyhands,lin2024motion,pokhariya2024manus,zhang2024both2hands} offer more realistic visuals and encourage natural motions, but they may trade off capture accuracy. \methodname{} balances accuracy, diversity, realism, and practicality by replicating in-the-wild settings during activity elicitation and estimating accurate 3D motion from marker-less motion capture.

\subsection{Hand Datasets and Annotations}
Hand datasets vary in sources and scales, each constructed for specific purposes and enriched with various annotations.

\noindent\textbf{Hand motion} is the most common type of annotation, represented as bounding boxes \cite{pirsiavash2012detecting,goyal2017something,shan2020understanding}, segmentation masks \cite{narasimhaswamy2019contextual,li2021eye,darkhalil2022epic}, 2D keypoints \cite{andriluka20142d,mckeenz,jin2020whole,sener2022assembly101}, 3D keypoints \cite{qian2014realtime,tang2014latent,sun2015cascaded,tzionas2016capturing}, or parameters of parametric models \cite{sridhar2013interactive,liu2022beat,pavlakos2024reconstructing}. Depending on the data source and desired detail, annotations are obtained through manual labeling \cite{tzionas2016capturing}, synthesis \cite{sharp2015accurate,zimmermann2017learning,mueller2017real,mueller2018ganerated}, markers and sensors \cite{tompson2014real,yuan2017bighand2,garcia2018first,joo2018total,zhang2022egobody,lin2024motion,zhang2024both2hands}, cross-view bootstrapping \cite{simon2017hand,moon2020interhand2}, iteratively trained keypoint extraction networks \cite{zimmermann2019freihand,ohkawa2023assemblyhands}, multi-view annotation \cite{ohkawa2023assemblyhands}, or audio refinement \cite{jin2024audio}. Semi-automatic and automatic labeling techniques significantly improve scalability.

\noindent\textbf{Object annotations} are provided given the frequent interaction between hands and objects. These can be acquired from manual labeling \cite{sridhar2016real,chao2021dexycb,liu2022hoi4d}, synthesis \cite{hasson2019learning,corona2020ganhand}, MoCap \cite{ye2021h2o,yang2022oakink,fan2023arctic,banerjee2024hot3d,liu2024taco,zhan2024oakink2,zhao2024m,lv2025himo}, hand-object reconstruction or retrieval~\cite{cao2021reconstructing,xie2023hmdo,min2024genheld}, or multi-view RGB-D data \cite{hampali2020honnotate,kwon2021h2o,hampali2022keypoint}. We demonstrate that multi-view RGB us sufficient for automatic annotation, enhancing scalability.

\noindent\textbf{Text annotations.} To capture the rich semantic meanings in human hand actions, certain datasets provide text annotations. These annotations may include action types \cite{pirsiavash2012detecting,goyal2017something,dreher2019learning,krebs2021kit,grauman2022ego4d}, atomic action descriptions
 \cite{damen2018scaling,li2021eye,damen2022rescaling,sener2022assembly101,grauman2024ego}, activity narration \cite{damen2018scaling,miech2019howto100m,damen2022rescaling,grauman2024ego}, activity commentary \cite{grauman2022ego4d,grauman2024ego,zhan2024oakink2,cha2024text2hoi}, object affordance \cite{yang2022oakink,jian2023affordpose,zhan2024oakink2,liu2024taco}, or body dynamics \cite{zhang2024nl2contact,zhang2024both2hands,yu2024signavatarslargescale3dsign}. Typically, annotations are manually provided, especially when data are collected in uncontrolled, unscripted environments \cite{damen2018scaling,damen2022rescaling,grauman2022ego4d,grauman2024ego}. Even in studio settings, post-processing is often required to extract meaningful action clips \cite{li2021eye,zhan2024oakink2}. We introduce a procedural instruct-and-annotate pipeline that maximizes semantic interpretability and minimizes the annotation effort.

\noindent\textbf{Contact} are derived from manual annotation \cite{shan2020understanding}, video grounding \cite{nagarajan2019grounded}, thermal sensors \cite{brahmbhatt2019contactdb,brahmbhatt2020contactpose} or geometry analysis \cite{taheri2020grab,zhu2024contactart,pokhariya2024manus} based on captured data. Our dataset enables us to derive contact regions using geometry analysis on both surface mesh and density fields.

\begin{table}[t!]
\centering
\small
\tabcolsep 1pt
\caption{Comparisons of 3D bimanual motion datasets. Dataset names are highlighted with different colors if it has \textcolor{darkgray}{no text annotations (gray)}, \textcolor{mygreen}{action type (green)}, \textcolor{mypink}{sparse description (red)}, and \textcolor{mypurple}{dense description (blue)}.
See the supp.~document for the full table.
}
\begin{tabular}{l|ccccccc}
\hline
 Name&  mins &  motions & poses & views & frames & sub. & obj. \\ \hline
  \rowA \multicolumn{8}{l}{{w/o Object Annotation}} \\
\textcolor{mygreen}{AssemblyHands}~\cite{ohkawa2023assemblyhands}&  630 & 62 & 203k & 12 & 3.03M & 34 & $/$ \\
 \textcolor{mypink}{Ego-Exo4D}~\cite{grauman2024ego}& $/$ & $/$ & \textbf{4.4M} & 5-6 & $/$ & \textbf{740} & $/$ \\
 \rowA \multicolumn{8}{l}{{w/ Object Annotation}} \\
 \textcolor{mygreen}{HOI4D}~\cite{liu2022hoi4d}& 1,333 & 4k & 1.2M & 1 & 2.4M & 4 & \textbf{800}\\
 \textcolor{darkgray}{ARCTIC}~\cite{fan2023arctic}& 121  & 339 & 218k  & 9 & 2.1M & 10 & 11\\
 \textcolor{mygreen}{TACO}~\cite{liu2024taco}& 202 & 2.3k & 363k & 13 & 4.7M & 14 &  196\\
 \textcolor{mypink}{OakInk2}~\cite{zhan2024oakink2}& 557 & 2.8k & 993k & 4 & 4.01M & 9& 75\\
  \textcolor{darkgray}{HOT3D}~\cite{banerjee2024hot3d}&  833  & 4.1k & 1.7M & 2-3 & 3.7M & 19 & 33\\
 \textcolor{mypurple}{\methodname (Ours)} & \textbf{2,034} & \textbf{13.9k} & {3.7M} & \textbf{51} & \textbf{183M} & {56} & 417\\ \hline
\end{tabular}
\label{tab:stats}
\vspace{-1em}
\end{table}

\begin{figure*}[t!]
     \centering
     \begin{subfigure}[b]{0.30\textwidth}
         \centering
         \includegraphics[width=\textwidth]{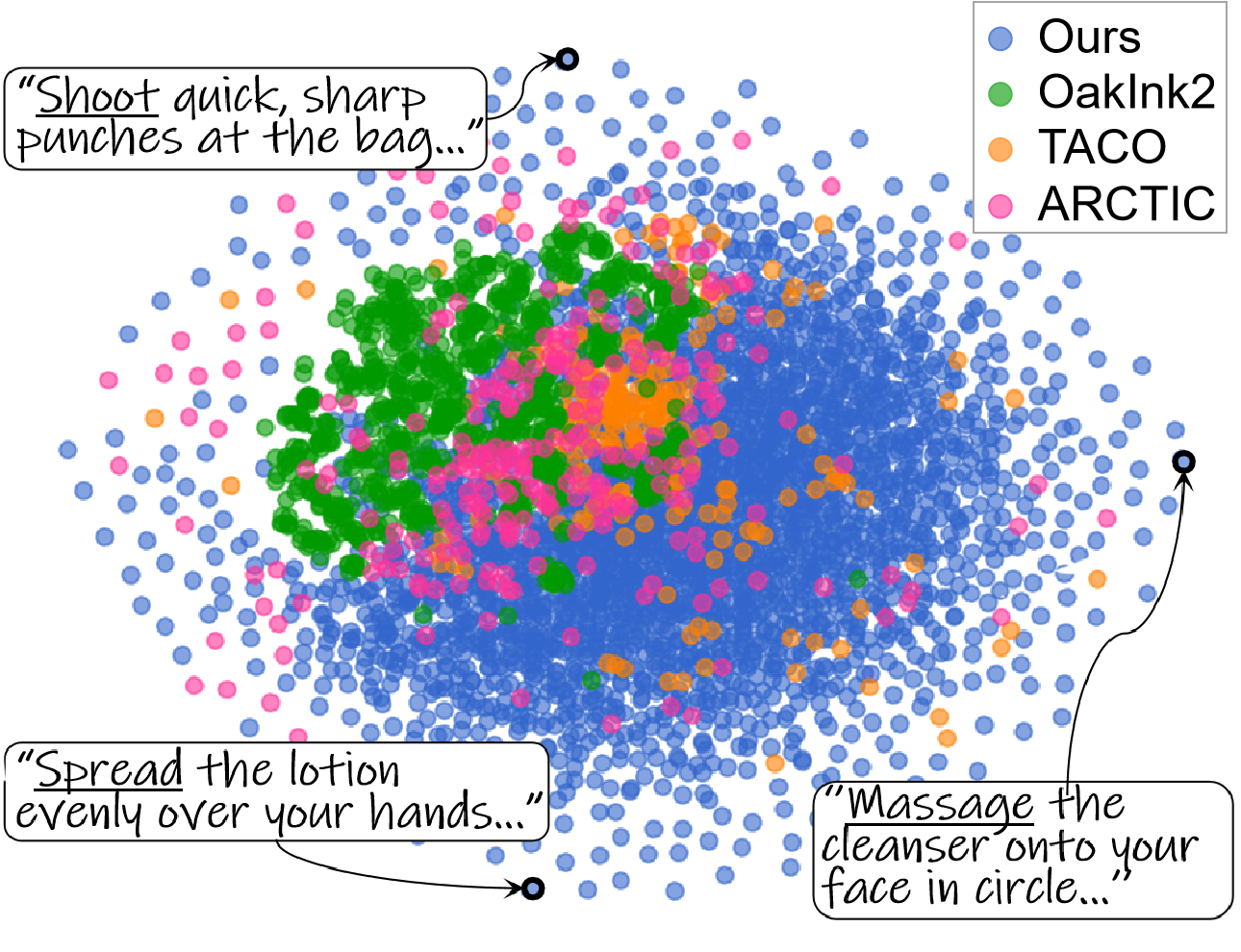}
         \label{fig:pose}
     \end{subfigure}
     \begin{subfigure}[b]{0.30\textwidth}
         \centering
         \includegraphics[width=\textwidth]{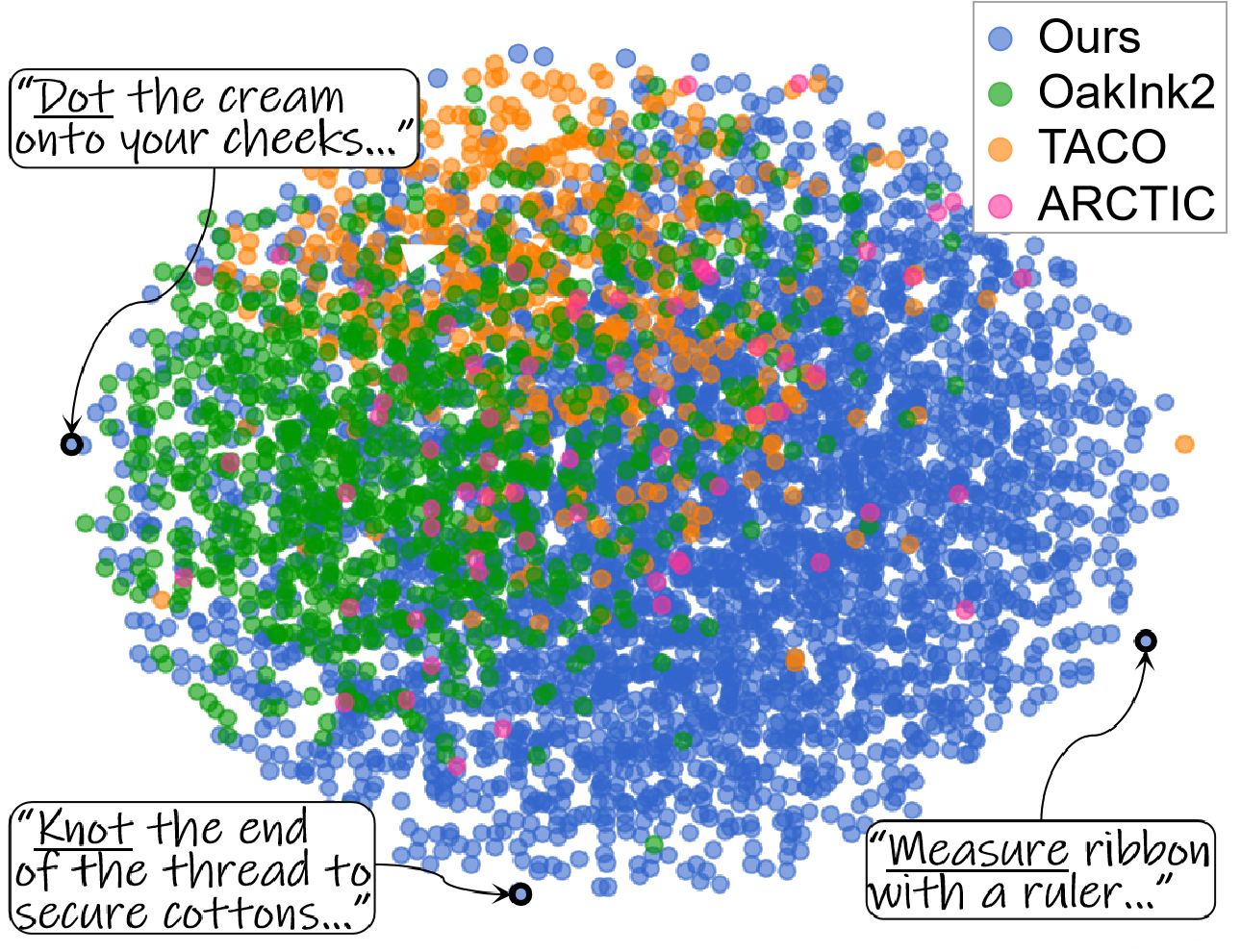}
         \label{fig:motion}
     \end{subfigure}
     \begin{subfigure}[b]{0.38\textwidth}
         \centering
         \includegraphics[width=\textwidth]{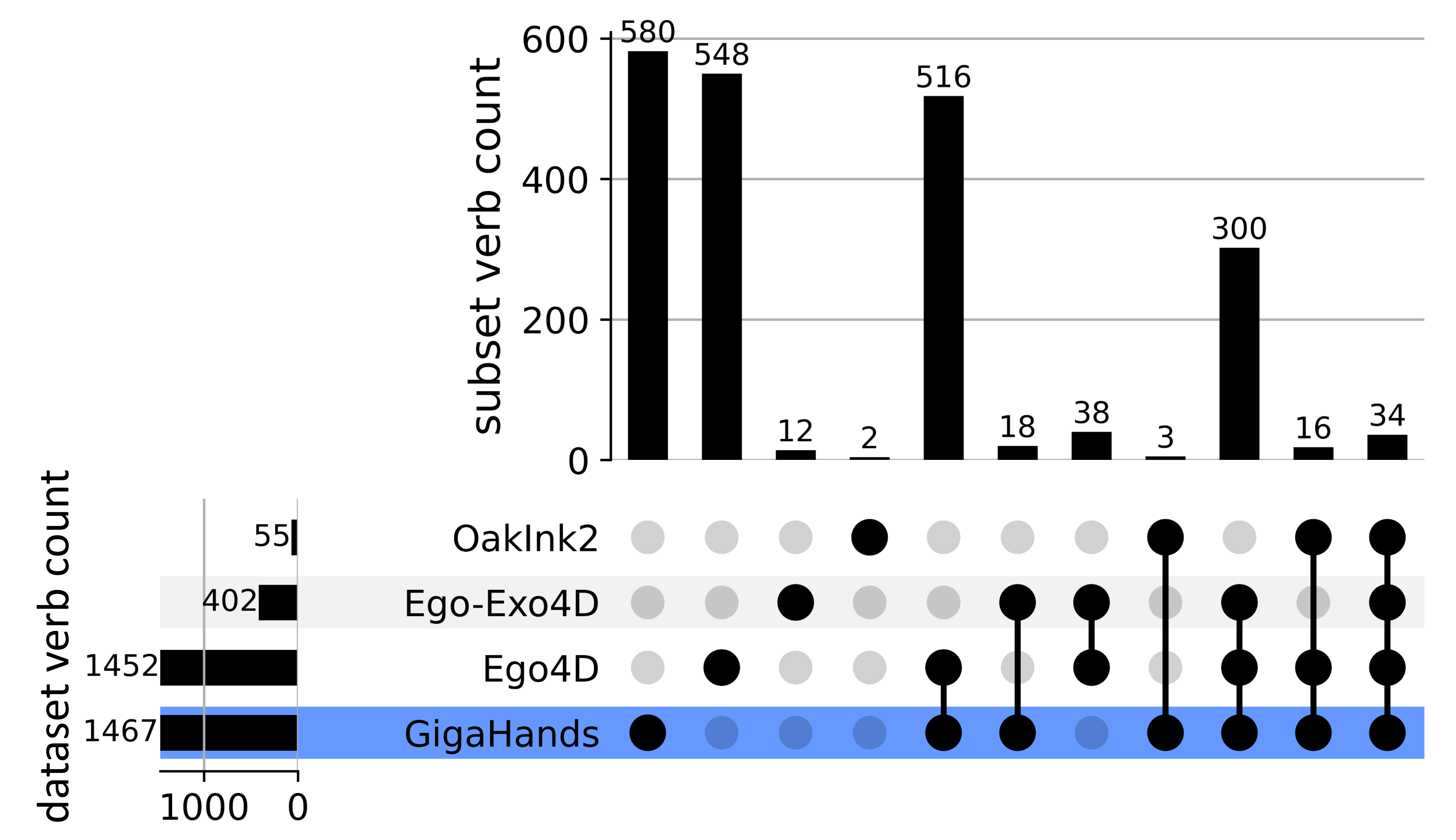}
         \label{fig:verbs}
     \end{subfigure}
     
        \caption{ \textbf{Dataset Diversity.} The left and middle figures illustrate the diversity of pose and motion variations in \methodname{}, visualized using t-SNE embeddings. Some points along the convex hull are highlighted with their corresponding text instructions, showcasing unique motions captured in our dataset. The right figure compares the verb sets among different datasets using an UpSet visualization~\cite{lex2014upset}. Each column represents the number of verbs exclusive to specific subsets of datasets, indicated by the connected dots below the columns. The rows indicate the total verb count in each dataset. \textcolor{mypurple}{\methodname{}} contains more verbs and more exclusive verbs compared to other datasets.
        }

\label{fig:diversity}
\vspace{-1em}
\end{figure*}

\section{The \methodname{} Dataset}
\paragraph{Dataset Characteristics.}
\methodname is a large and diverse dataset of bimanual hand activities with detailed annotations.
It encompasses a wide range of scenarios, including hand-object interactions, gesturing, and self-contacts performed by 56 subjects who collectively used 417 objects (see \Cref{tab:stats}).
The dataset contains 2,034 minutes of bimanual hand activities, surpassing the length of any existing 3D hand motion dataset.
Compared to other datasets that are unannotated~\cite{fan2023arctic,banerjee2024hot3d}, contain motion type annotations~\cite{ohkawa2023assemblyhands,liu2022hoi4d,liu2024taco}, or provide sparse motion descriptions~\cite{grauman2024ego,zhan2024oakink2}, \methodname offers detailed text annotations for all captured activities.
With a total of 13k instructed motion sequences, 14k annotated motion clips, and 84k augmented text descriptions, it is larger than any other dataset. 
Furthermore, \methodname includes 3.7 million bimanual 3D hand poses, represented by both 3D keypoints and MANO~\cite{romero2022embodied} hand meshes, comparable to the scale of Ego-Exo4D~\cite{grauman2024ego} -- the current largest annotated 3D hand pose dataset. However, unlike Ego-Exo4D, which is annotated from selected, non-continuous frames, \methodname provides continuous bimanual 3D hand poses and shapes.
Each motion clip was captured using 51 camera views, resulting in 183M RGB frames (and 366M unique hand images).
This multi-view setup enables new applications such as dynamic 3D radiance field reconstruction.
Additionally, the dataset contains annotations for hand/object segmentation masks, 3D object shape, pose and appearance, 2D/3D keypoints, and camera poses.

\paragraph{Diverse Hand Pose and Motion.}
\Cref{fig:diversity} (left) compares hand pose diversity across datasets using t-SNE~\cite{van2008visualizing} clustering of 3D keypoints of both hands.
Similarly, \Cref{fig:diversity} (middle) shows hand motion diversity through t-SNE clustering of the latent codes of a Variational Autoencoder trained for motion reconstruction.
\methodname exhibits significantly greater diversity than existing datasets in both bimanual hand pose and motion.
Additionally, datapoints near the boundary of our dataset seem to correspond to text annotations that are unique in our datasets, demonstrating that diverse verbs contribute to diverse poses and motions.

\paragraph{Diverse Text Annotation.}
\methodname contains activities that are more diverse than existing datasets~\cite{zhan2024oakink2,grauman2022ego4d,grauman2024ego} as measured using verb counts in
\Cref{fig:diversity} (right).
Since each verb corresponds to an activity, a diverse set of verb indicates activity diversity.
\methodname contains the most verbs (1467) with 580 of them being unique to our dataset.
The key to achieving this diversity is our \emph{instruct-to-annotate} strategy (see \Cref{sec:annotation}) for automated labeled activity sourcing.
Please refer to the supp.~document for the source of verbs and detailed verb comparison.

\begin{figure}[t!]
     \centering

         \includegraphics[width=0.46\textwidth]{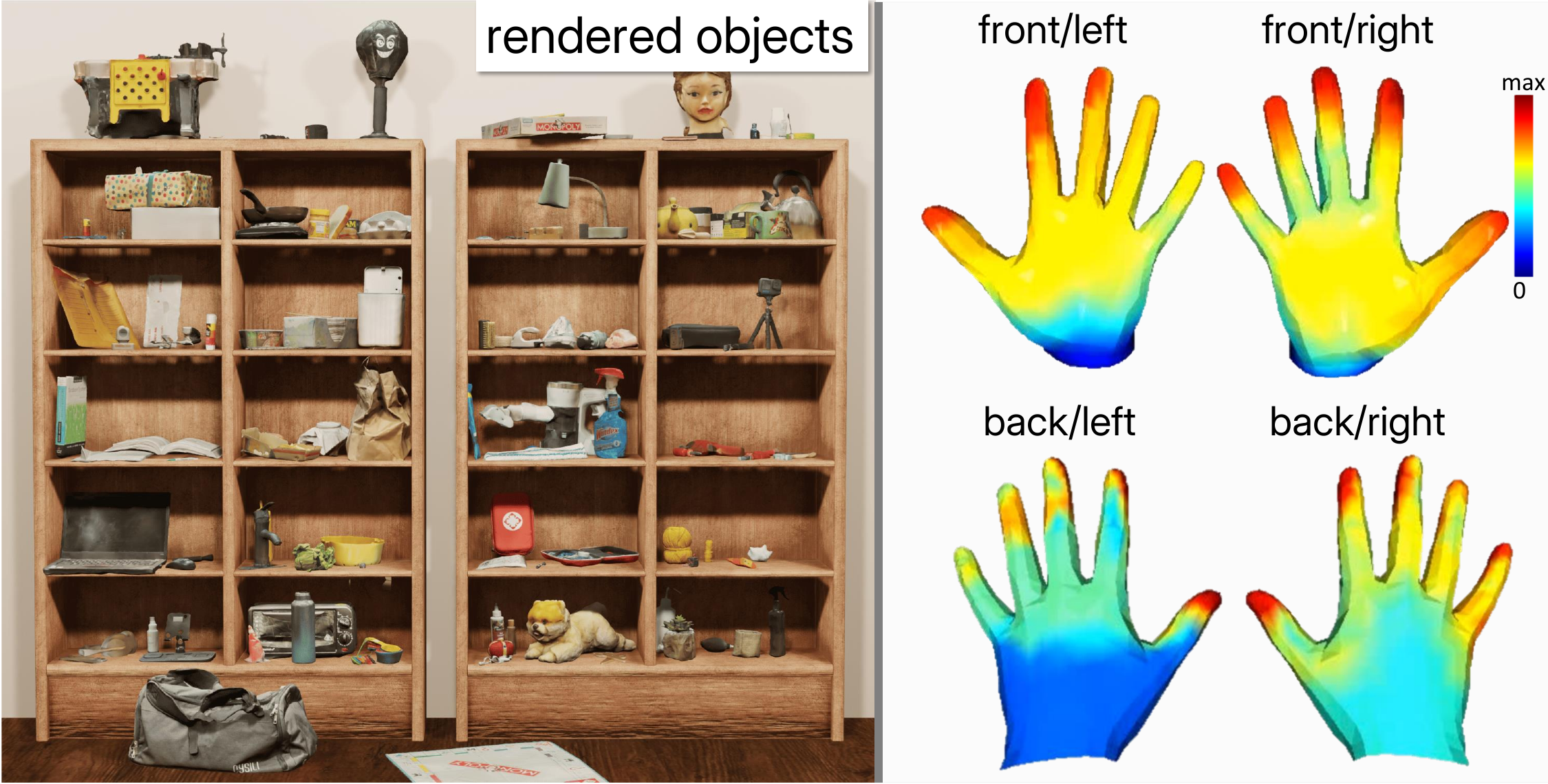}
        
        \caption{\textbf{Diverse Objects and Frequent Hand Contact Regions.} \methodname{} provide objects (left) spanning diverse scenarios, including cooking, office working, crafting, entertainment, and housework. The diverse activities result in contact regions (right) spanning both the front and back of both hands.
        }
        \label{fig:objects_n_contacts}
        \vspace{-1em}
\end{figure}

\begin{figure*}[t!]
    \centering
    \includegraphics[width=\textwidth]{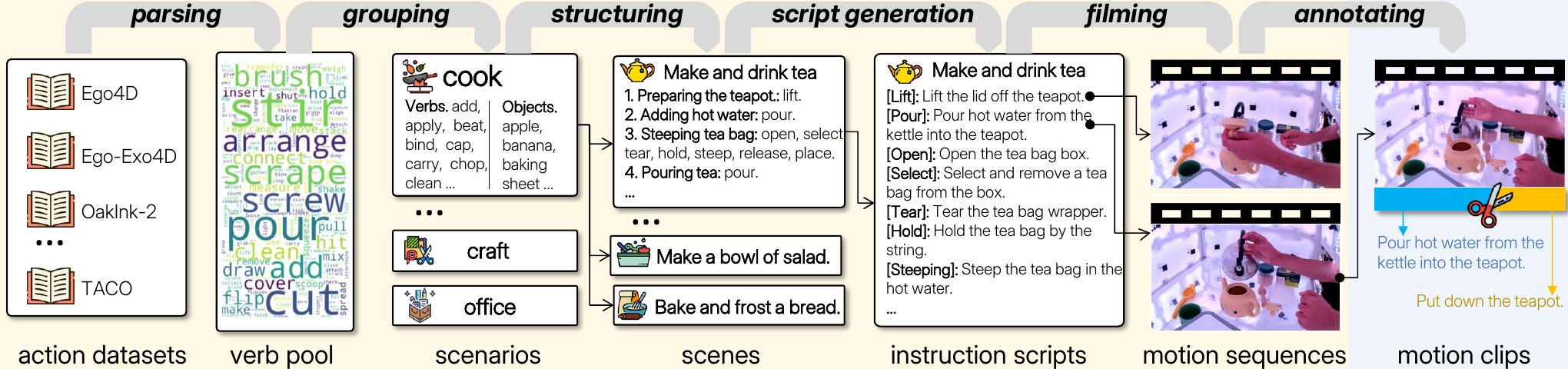}
    \caption{\textbf{Instruct-to-Annotate Pipeline.} The instruction elicitation process (left yellow block) creates atomic action-level instruction scripts in a temporally smooth order, structured within scenes. This is achieved by parsing action datasets, grouping verbs into a pool, structuring scenarios, and generating scene scripts. During filming, subjects act according to these scripts, producing recorded motion sequences. Annotators then process these sequences (right blue block) by segmenting them into clips and annotating unscripted motions.
    }
    \label{fig:instruct_and_annotate}
    \vspace{-1em}
\end{figure*}

\paragraph{Diverse Objects.}
\Cref{fig:objects_n_contacts} shows some of the tabletop objects included in \methodname.
We provide 3D meshes \cite{arcodeCodeAugmented}
with associated textures for rigid objects and multi-view segmentation masks for non-rigid objects.
Different from most datasets, we also include small-scale objects (\eg pens) that are difficult to capture.
We employ single-view reconstruction~\cite{meshyMeshyDocs} to obtain meshes for these small objects.
In total, the dataset includes 417 objects, with 310 multi-view scanned meshes and 31 single-view generated meshes.

\paragraph{Diverse Contacts.}
Since we have 3D meshes both hands and objects, we can use them to estimate \textbf{contact maps} on both hands, following the approach in~\cite{taheri2020grab}.
\Cref{fig:objects_n_contacts} shows the accumulated contact regions from randomly sampled frames across the dataset.
The results reveal diverse contact areas, including regions between fingers and on the back of the hands (e.g. punching uses the backs of hands).
The right hand interacts with objects more frequently than the left, since a majority of our subjects were right-handed.

\section{Dataset Acquisition}
We present our data acquisition pipeline, which balances realism, diversity, and accuracy while reducing annotation effort.
Our procedural pipeline, called \emph{Instruct-to-Annotate} (see \Cref{fig:instruct_and_annotate}), consists of several key components:
\begin{packed_itemize}
    \item \textbf{Procedural instruction elicitation} (Section~\ref{sec:instruction}) for generating instructions to guide participants during data capture.
    \item \textbf{Filming} (Section~\ref{sec:capturing}) using a markerless multi-camera system to record high-quality hand activity sequences.
    \item \textbf{Motion annotation and text augmentation} (Section~\ref{sec:annotation}) to refine sequences by breaking them into shorter clips with accurate annotations.
    \item \textbf{Hand motion} (Section~\ref{sec:hand_motion}) and \textbf{object motion estimation} (Section~\ref{sec:object_motion}) for detailed 3D hand and object motions from the captured multi-view RGB videos.
\end{packed_itemize}

\subsection{Instruction Elicitation}
\label{sec:instruction}
To ensure diverse hand activities and reduce annotation effort, the \emph{Instruct-to-Annotate} pipeline starts with a procedural instruction elicitation protocol.
We began by sourcing atomic actions from multiple datasets~\cite{grauman2022ego4d,grauman2024ego,zhan2024oakink2,liu2024taco}, extracting a pool of verbs corresponding to those actions.
To elicit subjects to perform these verbs (actions), we manually associated 
each verb with multiple objects and, with the assistance of an LLM~\cite{achiam2023gpt}, grouped the verbs and objects into different scenarios such as cooking and eating, office work, crafting, entertainment, and housework.
We then structured these scenarios into scenes where the objects could co-occur. Using LLM, we organized the scenes into lists of activities that utilized the objects and verbs in a temporally smooth order and automatically generated detailed instruction scripts.
These scripts comprise 5 scenarios, 25 scenes, 191 activities, and 1370 instructions containing a total of 533 verbs. 
Please see the supp.~document for details.

\subsection{Filming}
\label{sec:capturing}
\paragraph{Hardware.} To capture our dataset, we use a custom-designed multi-camera tabletop capture system
that consists of 51 RGB cameras uniformly arranged within a cubic capture volume, with each face of the cube containing a 3$\times$3  grid of cameras evenly illuminated by LED lights.
Inside the cube, a transparent glass surface serves as a supportive platform for objects. Each camera records at 30fps with a resolution of 1280$\times$720.
Cameras are software-synchronized, with the temporal phase misalignment being less than 3~ms.
Camera intrinsics and extrinsics are obtained using COLMAP~\cite{schoenberger2016sfm,schoenberger2016mvs} aided by fiducial markers.

\paragraph{Filming Process.}
During filming, subjects perform actions according to the provided instructions, keeping both hands within the filming area.
Instructions are converted to audio and played sequentially to guide the participants, while an operator controls the capture by playing each audio instruction and recording each motion sequence.
If the ending state of a performance does not align with the next instruction, a corrective instruction will be given and re-recorded to ensure smooth transitions and reduce annotation efforts.
This approach ensures all recorded sequences correspond to a pre-scripted or recorded instruction.

\subsection{Action Annotation \& Augmentation}
\label{sec:annotation}
Though each filmed motion sequence was paired with an instruction, manual annotation was still necessary for two reasons. First, the instructions generated by the LLM occasionally contained inconsistencies or hallucinations.
Second, participants might misunderstand the instructions or add unintentional actions during filming.
To encourage subjects to act freely and naturally, we intentionally retained these recordings.
Annotators then split these sequences into individual clips and annotated any actions not included in the original instructions.
This process resulted in a more accurate dataset with matched (description, clip) pairs.
In total, we refined the 13k motion sequences into 14k motion clips.

To further enhance the text descriptions, we used the LLM to rephrase each description 5 times, providing multiple textual variations for each motion clip. 
This expanded our 14k motion clips into 84k motion-text pairs containing 1,467 unique verbs.
The augmentation prompts and examples are provided in the supp.~document.

\subsection{Hand Motion Estimation}
\label{sec:hand_motion}
\methodname provides detailed hand motion data, including 2D and 3D hand keypoints and MANO~\cite{romero2022embodied} meshes for both hands.
Since existing hand shape and pose estimation did not work well enough, we built our own hybrid method.
We begin by obtaining bounding boxes for the hands across videos using YOLOv8~\cite{reis2023real}.
Then, 2D keypoints are extracted from the MANO meshes estimated with HaMeR~\cite{pavlakos2024reconstructing} and handedness (left or right) is determined with ViTPose~\cite{xu2022vitpose}. HaMeR meshes cannot directly be used since they lacks accurate depth. 
With camera parameters, 2D keypoints are triangulated across views~\cite{sridhar2013interactive} to obtain accurate 3D positions.
To ensure temporal smoothness in the sequences, the one-euro filter~\cite{casiez20121} is applied to the both 2D and 3D keypoints.
With the bounding box and 2D \& 3D keypoints, we fit the MANO parametric hand model~\cite{romero2022embodied} following the EasyMoCap pipeline~\cite{easymocap}.
This results in a fully automated pipeline for extracting coherent and accurate hand motions.
Details and the evaluation of each step are provided in the supp.~document.

\subsection{Object Motion Estimation}
\label{sec:object_motion}
\methodname also provides 3D object motions represented as 3D shape and 6D pose. Given target meshes obtained from pre-scanning or single-view reconstruction, we track these objects with multi-view constraints. First, we segment the target objects from the background in each view.
We detect salient objects across the video using DINOv2~\cite{oquab2023dinov2} on frames subsampled at 1 fps.
Using both object text descriptions and rendered template meshes from multiple views, we select the top-k bounding boxes most aligned with the template mesh throughout the video with Grounding DINO~\cite{liu2023grounding}.
To eliminate false positives, we use OpenCLIP~\cite{ilharco_gabriel_2021_5143773} to filter out boxes aligned with negative prompts.
With the positive bounding boxes as the prompt, we use SAM2~\cite{ravi2024sam} to segment object masks throughout the video.

Since no existing method for object pose estimation worked well~\cite{ornek2023foundpose}, we decided to build our own robust method that exploits our dense multi-view setting.
We use a differentiable rendering approach~\cite{ravi2020pytorch3d} supervised by the multi-view masks.
To initialize the translation and rotation, we first build a radiance field using Instant-NGP~\cite{muller2022instant}.
We initialize the object's translation using the center of the density field.
To address potential object symmetries, we use object appearance to initialize the rotation.
Following FoundPose~\cite{ornek2023foundpose}, we render the template mesh with multiple initial rotations and use DINOv2 features to find the best match, aggregating cross-view information.
This process yields precise object motion estimates even in complex, cluttered scenes. 
More details and the evaluation of each step are provided in the supp.~document.

\section{Applications \& Experiments}
We demonstrate the utility of \methodname in several applications that require large-scale data,
including text-driven hand motion synthesis (\Cref{sec:text2motion}), motion captioning for our dataset (\Cref{sec:caption_within}) and for in-the-wild datasets (\Cref{sec:caption_inthewild}), and dynamic radiance field reconstruction (\Cref{sec:reconstruction}).
More results can be found in the supp.~document.
\begin{figure*}[!t]
    \centering
    \includegraphics[width=0.95\linewidth]{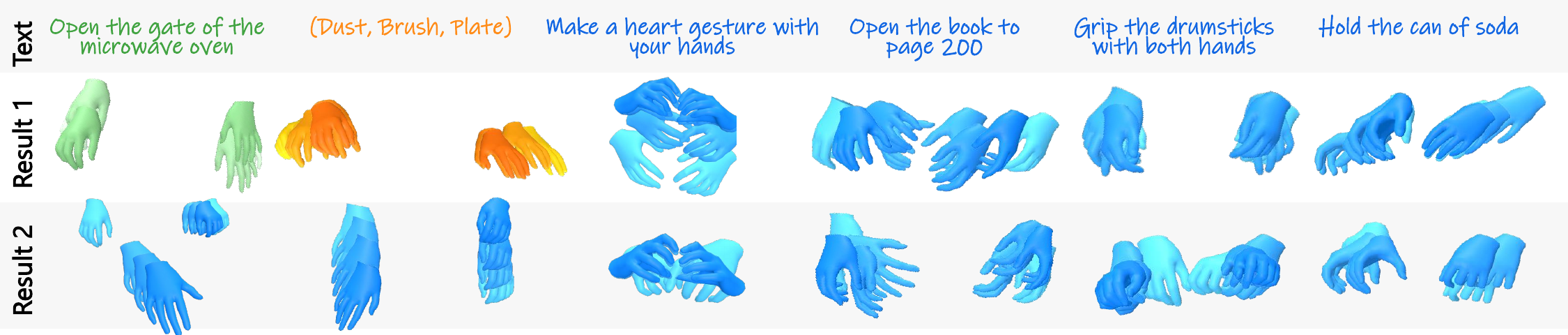}
    \caption{Generated motions from models trained on different datasets. Texts highlighted in \textcolor{green}{green}, \textcolor{orange}{orange}, and \textcolor{blue}{blue} come from the \textcolor{green}{OakInk2}, \textcolor{orange}{TACO}, and \textcolor{blue}{\methodname} datasets. In the bottom two rows, hand meshes highlighted in these colors are generated by models trained on the corresponding datasets. The model trained on \methodname can generate diverse motions from a single text (right four columns) and accurate motion with text from other datasets (left two columns). Darker color indicates later frame in the sequence.}
    \label{fig:t2m_qual}
    \vspace{-1em}
\end{figure*}

\subsection{Text-driven Hand Motion Synthesis}
\label{sec:text2motion}
Generating diverse and complex motions is crucial for training virtual agents and robotic manipulation. We use \methodname{} to train models for text-driven hand motion synthesis, overcoming the limitations of approaches that rely on strict conditions~\cite{zhan2024oakink2} or generate only simple skills~\cite{cha2024text2hoi} due to constrained data.  We demonstrate that the scale of \methodname{} leads to better performance compared to other datasets~\cite{liu2024taco, zhan2024oakink2} and smaller subsets of \methodname{}.

\paragraph{Training and Evaluation Protocol.}

We split \methodname into train, test, and val sets with a ratio of 16:3:1. From among our annotations, we chose $42$ 3D keypoints for both hands as the representation for model training (see the supp. document for an analysis of alternative representations). To evaluate the naturalness, diversity, and alignment of generated motions with textual descriptions, we use metrics from \cite{Guo_2022_CVPR}, including R-Precision, Multimodal Distance (MM Dist), Fréchet Inception Distance (FID), Diversity (Div.), and Multimodality (MM.). For computing FID and Div., we train a motion autoencoder to define a compressed motion space. For R-Precision, MM Dist, and MM., we employ contrastive learning to create a joint motion-text embedding space, following \cite{Guo_2022_CVPR}. Feature extractors are trained independently for each dataset and subset.

\begin{table}[t!]
    \footnotesize
    \tabcolsep 3pt
    \caption{\footnotesize{Quantitative results
    for text-driven motion synthesis with models trained on different datasets. \emph{upper bound} indicates performance calculated with the ground truth. We report the mean of 20 evaluations, and $\rightarrow$ means the closer to the upper bound the better. The model trained on \methodname performs best on most metrics.}}
    \vspace{-1em}
    \centering
    \begin{tabular}{l c c c c c c c}
    \toprule
     \multirow{2}{*}{Dataset}  & \multicolumn{3}{c}{R Precision(\%)$\uparrow$} & \multirow{2}{*}{MM Dist.$\downarrow$} & \multirow{2}{*}{FID$\downarrow$} & \multirow{2}{*}{Div.$\rightarrow$} & \multirow{2}{*}{MM.$\uparrow$} \\

    \cline{2-4}
       ~ & @1 & @2 & @3 \\
    
    \midrule

    {upper bound}& 64.4 & 86.2 & 89.4 & 2.86 & 0.045 & 14.2 & - \\
    TACO \cite{liu2024taco} & 18.9 & 37.7 & 52.9 & 7.39 & 11.0 & 11.1 & 6.83 \\
    \hline
    {upper bound} & 50.4 & 71.2 & 81.1 & 3.67 & 0.022 & 9.30 & - \\
    OakInk2 \cite{zhan2024oakink2} & 17.9 & 31.7 & 47.9 & 7.75 & 19.6 & 6.88 & 3.45 \\
    \hline
    {upper bound} & 77.4 & 88.8 & 91.3 & 2.96 & 0.002 & 11.9 & - \\
    \methodname{} & \textbf{31.2} & \textbf{44.7} & \textbf{53.1} & \textbf{6.68} & \textbf{4.70} & \textbf{10.5} & \textbf{9.11} \\
    \bottomrule
    \end{tabular}

    \label{tab:quant_t2m}
\vspace{-2em}
\end{table}

\paragraph{Results.}
We report text to hand motion synthesis performance using the T2M-GPT \cite{zhang2023generating} backbone in \Cref{tab:quant_t2m} (see the supp.~document for more results on GRU-based~\cite{chuan2022tm2t} and diffusion~\cite{tevet2023human} architectures).
Since the evaluation embedding spaces are trained on different test sets, we also evaluate the metrics on ground truth test data to indicate upper-bound performance for each test set. Models trained on \methodname{} outperform others on all metrics except MM Dist., with higher R-Precision due to better text-motion alignment. Compared to OakInk2 (which also includes textual descriptions for hand motions), \methodname{} achieves significantly better FID, Diversity, and Multimodality scores, likely due to its greater dataset diversity.
\Cref{fig:t2m_qual} shows hand motions generated by T2M-GPT trained with \methodname{} using text inputs from the test sets of \methodname{}, OakInk and TACO.
Even without object geometry input, our model generates reasonable hand shapes and poses for object manipulation, benefiting from the diversity of objects in our dataset. Our model can also generate reasonable motion using text from other datsasets, showing the comprehensiveness of \methodname{}. 
\begin{figure}[t!]
    \centering
    \includegraphics[width=0.96\linewidth]{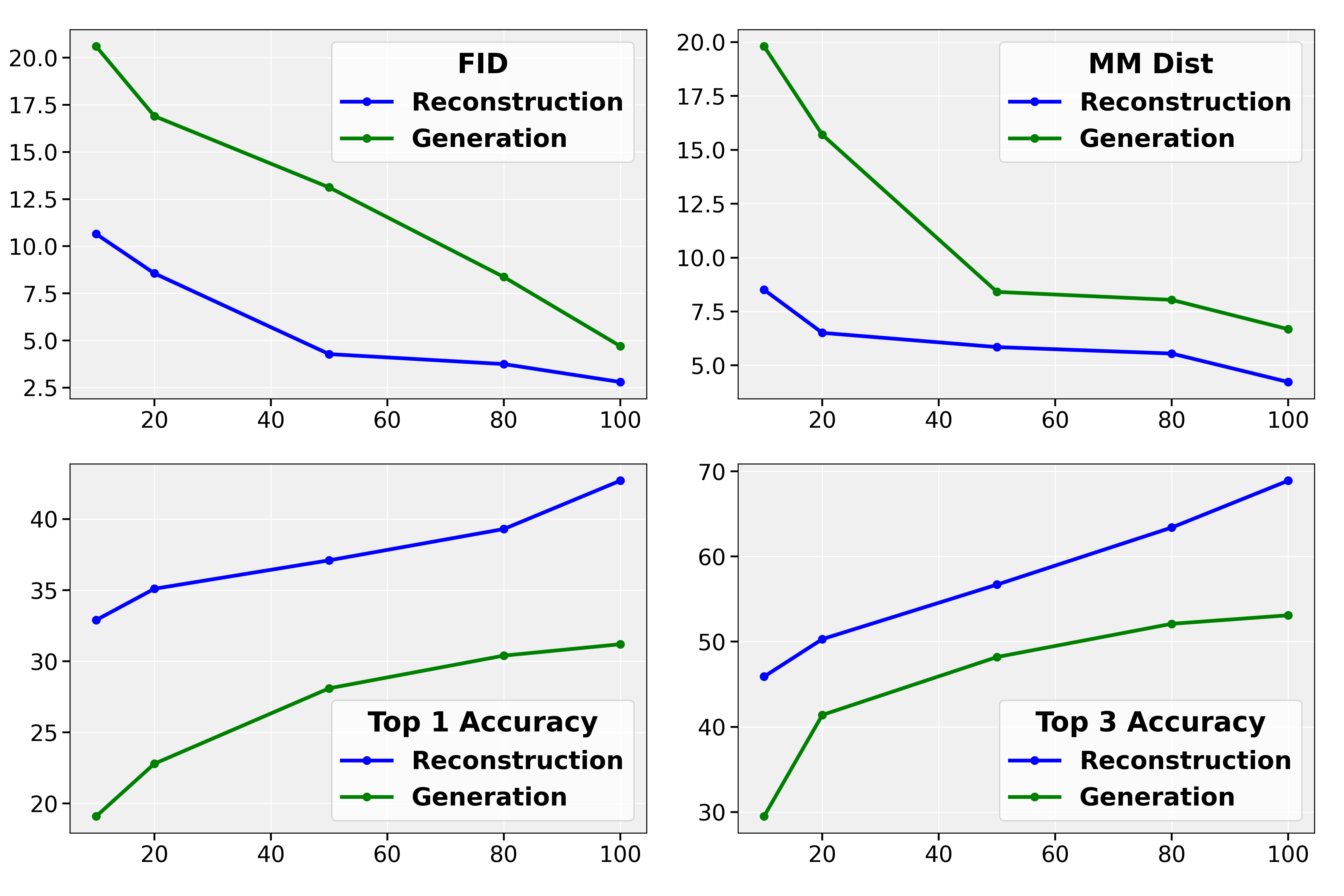}
    \caption{ Effect of dataset size on motion reconstruction and text-to-motion generation performance.
    The x-axis shows the percentage of training data used (10\%, 20\%, 50\%, 80\%, and 100\%), and the y-axis displays performance metrics: FID, MM Dist., Top-1, and Top-3 accuracy. Larger datasets consistently improve performance across all metrics, highlighting the benefits of increased data scale.
    }
    \label{fig:t2m_scale}
    \vspace{-1em}
\end{figure}

\paragraph{Effect of Data Scale.}
\Cref{fig:t2m_scale} illustrates the impact of dataset size on motion reconstruction and text-to-motion generation. Based on the T2M-GPT architecture, we train a motion VQ-VAE for reconstruction and a generative pretrained transformer model for text-to-motion generation using 10\%, 20\%, 50\%, 80\%, and 100\% of the training set, evaluating on the same test set. We evaluate FID, MM Dist., and Top 1 \& Top 3 accuracies. Most metrics continually improve with larger datasets, demonstrating the value of larger-scale data.

\subsection{Hand Motion Captioning}
Captioning human hand motion helps interpret action intent. We train a motion captioning model on \methodname and generate captions for its test set motions; we also show examples of captioning in-the-wild datasets.

\paragraph{Training and Evaluation Protocol.} 
We adopt TM2T~\cite{chuan2022tm2t} as the model backbone.
For text-motion alignment evaluation, we utilize R-Precision and MM Dist., employing the same embedding space described in \Cref{sec:text2motion}.
To evaluate alignment with the ground-truth text annotations, we use BLEU~\cite{papineni2002bleu}, ROUGE~\cite{lin2004rouge}, and BERTScore~\cite{zhang2019bertscore}.
Since our dataset has diverse verbs, we evaluate text diversity with distinct-n~\cite{li2015diversity} and Pairwise BLEU~\cite{shen2019mixture}.
Since TACO lacks text annotations, we use its triplet labels (\texttt{<action, tool, object>}) as scripts, transforming motion captioning in TACO into a classification task.

\begin{table}[ht!]
\small
\tabcolsep 2pt
    \centering
    \caption{\footnotesize{Quantitative evaluation for motion captioning with models trained on different datasets. \methodname performs best on most metrics.}}
    \vspace{-1em}
    \begin{tabular}{l c c c c }
    \toprule
     {Datasets}  & {RP@1(\%)$\uparrow$} & {RP@2(\%)$\uparrow$} &{RP@3(\%)$\uparrow$} &{MM Dist$\downarrow$} \\

    \midrule

    {upper bound} & 64.4 & 86.2 & 94.2 & 2.86\\
    TACO \cite{liu2024taco} & 43.5 & 59.6 & 68.5 & 5.84  \\
    \hline
    {upper bound} & 51.1 & 69.5 & 79.6 & 3.67 \\
    Oakink2 \cite{zhan2024oakink2} & 40.4 & 58.8 & 68.2 & \textbf{4.55} \\
    \hline
    {upper bound} & 75.3 & 89.1 & 93.9 & 2.87\\
    \methodname & \textbf{57.0} & \textbf{66.1} & \textbf{69.8} & 5.37  \\
    
    \bottomrule
    \end{tabular}

    \label{tab:quant_m2t_metric}
\vspace{-1em}
\end{table}

\begin{table}[ht!]
\small
\tabcolsep 1pt
    \centering
    \caption{\footnotesize{Pairewise-BLEU, BLEU@4, ROUGE, distinct-n, and BERTScore for motion captioning with models trained on different datasets. \methodname performs best on most metrics.}}
    \vspace{-1em}
    \begin{tabular}{l c c c c c c}
    \toprule
     {Datasets} & {\shortstack{P-B}$\downarrow$} & {B@4$\uparrow$} & {R$\uparrow$}& {dist.-1(\%)$\uparrow$}& {dist.-2$\uparrow$(\%)}& {BScore$\uparrow$}\\
    
    \midrule
    TACO \cite{liu2024taco} & 4.59 & 39.4 & \textbf{61.2} & 2.03 & 6.35 & \textbf{57.1} \\
    Oakink2 \cite{zhan2024oakink2} & 8.21 & 39.9 & 56.3 & 2.76 & 6.82 & 35.3 \\
    \methodname & \textbf{0.916} & \textbf{43.1} & 57.7 & \textbf{15.3} & \textbf{36.9} & 55.4 \\
    
    \bottomrule
    \end{tabular}

    \label{tab:quant_m2t_nlp}
\vspace{-1em}
\end{table}

\newcolumntype{P}[1]{>{\centering\arraybackslash}p{#1}}
\begin{figure*}[t!]
    \centering
    \includegraphics[width=0.95\linewidth]{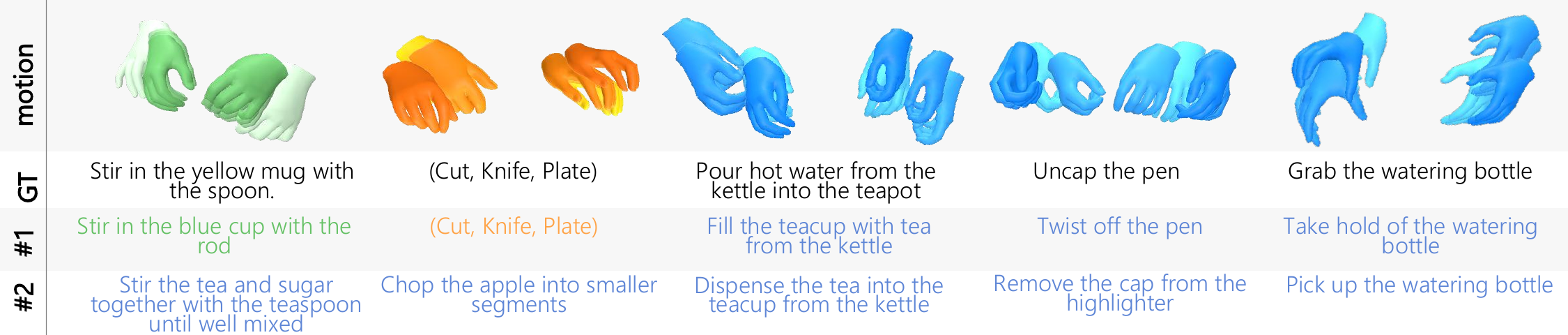}
    \caption{\textbf{Motion captioning results with different datasets.} Each column shows a motion sequence, its ground truth text description, and two generated texts. Hand motions highlighted in \textcolor{green}{green}, \textcolor{orange}{orange}, and \textcolor{blue}{blue} come from \textcolor{green}{OakInk2}, \textcolor{orange}{TACO}, and \textcolor{blue}{\methodname}, respectively. Texts highlighted in these colors are generated by models trained on the corresponding datasets. The model trained on \textcolor{blue}{\methodname} generates diverse captions from a single motion (right three columns) and accurately captions motions from other datasets (left two columns).}
    \label{fig:m2t_qual}
    \vspace{-1em}
\end{figure*}

\begin{figure}[t!]
    \centering
    \includegraphics[width=0.48\textwidth]{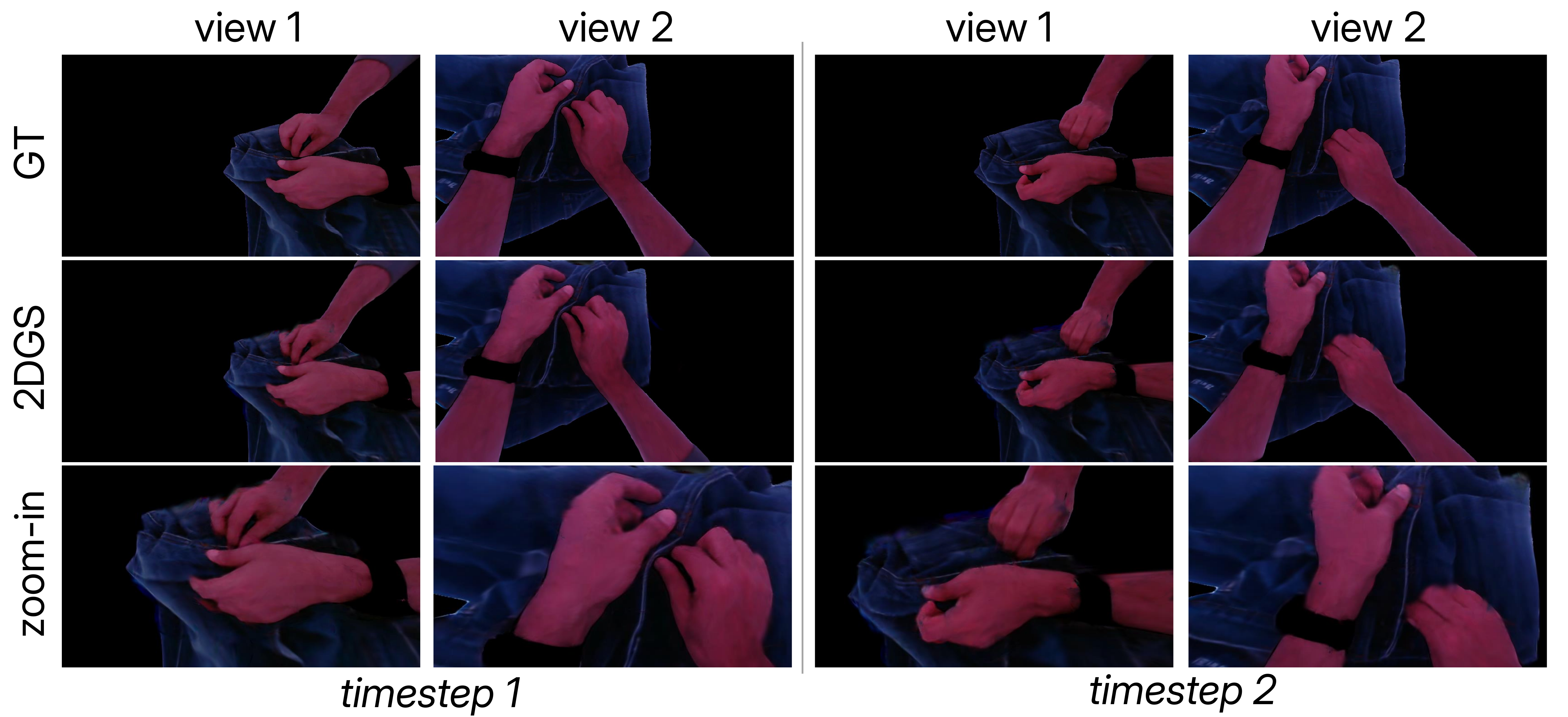}
    \caption{Synthesized test views using 2DGS for the motion `zip up the pants,' displayed at two timesteps and from two viewpoints. The synthesized views faithfully reconstruct the scene by leveraging the ample camera views provided by \methodname{}.}
    \vspace{-1em}
    \label{fig:novel_view}
\end{figure}

\vspace{-1em}
\paragraph{Hand Motion Captioning for \methodname}
\label{sec:caption_within}
\Cref{fig:m2t_qual} shows examples of generated 3D motion captions. The model trained on \methodname{} generates diverse and accurate captions. Though objects are not present in the input, the model still generates captions which include reasonable object descriptions.
\Cref{tab:quant_m2t_metric} and \Cref{tab:quant_m2t_nlp} report the quantitative performance of models trained on different datasets~\cite{zhan2024oakink2, liu2024taco}.
The model trained on \methodname achieves superior motion-text alignment, particularly in Pairwise BLEU and distinct-n, which reflect caption diversity. While our accuracy is marginally lower compared to the simple triplet captioning in TACO, our approach produces richer and more varied captions.

\vspace{-1em}
\paragraph{3D Hand Motion Captioning for Unlabeled Datasets}
\label{sec:caption_inthewild}
\methodname's large variety motions and associated annotations allows models trained on it to caption other 3D hand motion datasets.
Fig.~\ref{fig:m2t_qual} shows some examples using the model trained on \methodname{} to caption motions from TACO and OakInk2 datasets after aligning their motion range with \methodname{}.
This model trained with \methodname{} generates finer-granularity captions for these other datasets.

\subsection{Dynamic Radiance Field Reconstruction}
\label{sec:reconstruction}
Unlike existing hand activity datasets, \methodname{} provides dense camera views.
This enables new applications, such as building dynamic radiance fields.
From the 51 views provided, we remove 12 camera views due to lighting issues and randomly select one view as the test view, resulting in 38 views used for training.
We segment consistent object and hand masks throughout the video as described in \Cref{sec:object_motion}.
We then fit 2DGS~\cite{huang20242d} for frame-wise radiance field reconstruction, initializing each frame with the previous frame for temporal consistency. \Cref{fig:novel_view} shows an example of synthesized test views for a motion clip.
Notably, the pants are a non-rigid object that cannot be easily tracked, but \methodname provides ample camera views for capturing non-rigid objects (more results in supp.~document).

\section{Conclusion}
We present \methodname{}, a massive annotated dataset of bimanual hand activities. The dataset contains 14k motion clips with 3D hand and object motions from 56 subjects interacting with 417 real-world objects, all captured from 51 camera views. It provides 183 million unique image frames and 84k textual descriptions, enabling a wide range of applications. We demonstrated how the scale and diversity of the dataset benefit text-driven motion synthesis and motion captioning, both within the dataset and on other data. Additionally, we have shown that \methodname{} enables dynamic radiance field reconstruction, opening possibilities for downstream tasks.

\noindent\textbf{Limitations and Future Directions.} Despite its scale and diversity, \methodname{} has limitations. The studio setting confines data collection to a limited space, making it challenging to accurately capture motions that require larger environments. While we can track rigid objects and object parts, fully automatic tracking of articulated and non-rigid objects remains challenging. 
Moreover, although we have showcased applications in motion synthesis and captioning, further research could explore how the dataset's scale and diversity can enhance robotic manipulation and human-computer interaction tasks. 

\section*{Acknowledgement}
This research was supported by AFOSR grant FA9550-21-1-0214, NSF CAREER grant \#2143576, and ONR DURIP grant N00014-23-1-2804. We would like to thank the OpenAI Research Access Program for API support and extend our gratitude to Ellie Pavlick, Tianran Zhang, Carmen Yu, Angela Xing, Chandradeep Pokhariya, Sudarshan Harithas, Hongyu Li, Chaerin Min, Xindi Qu, Xiaoquan Liu, Hao Sun, Melvin He and Brandon Woodard.

{
    \small
    \bibliographystyle{ieeenat_fullname}
    \bibliography{main}
}
\end{document}


\maketitle

The supplementary material for \methodname provides additional details and results to support the main paper. It includes supplementary videos, examples of data and annotations, comprehensive comparisons of dataset statistics, detailed explanations of our hand and object motion tracking methods, elaborations on text instructions and annotations (including prompts and examples), further experiments on text-driven motion synthesis, motion captioning, and dynamic reconstruction. We also provide additional applications on hand-object motion tasks, as well as detailed inspections of the dataset, such as object visualizations, the verb pool, and lists of scenarios and scenes.

\tableofcontents

\newpage

\section{Data and Annotation Example}
\Cref{tab:annotation} illustrates the types of annotations we provided for a single motion clip. These include the text instruction, text annotation, augmented annotations, original multi-view RGB videos, 2D and 3D keypoints, 3D hand meshes, hand motions, object masks, textured object meshes, and object motions. 

\begin{table}[ht]
\tabcolsep 0pt
\begin{tabular}{cccc}
\hline 
    type & view \#1 & view \#2 & view \#3 \\
    \hline 
    instruction  &
    \multicolumn{3}{c}{Place your left-hand fingers on the fretboard of the ukelele to form chords.}
     \\\hline 
     
     annotation  & 
\multicolumn{3}{c}{Place your left-hand fingers on the fretboard of the ukelele to form chords.}
     \\
     \hline
\multirow{5}{*}{augmented annotation} & \multicolumn{3}{c}{Position your left-hand fingers on the fretboard to shape chords.} \\
                      & \multicolumn{3}{c}{Rest your left-hand fingers on the fretboard to create chords.}  \\
                      & \multicolumn{3}{c}{Set your left-hand fingers on the fretboard to assemble chords.}   \\
                      & \multicolumn{3}{c}{Arrange your left-hand fingers on the fretboard to build chords.}   \\
                      & \multicolumn{3}{c}{Use your left-hand fingers on the fretboard to construct chords.}  \\\hline \\
     \small{video} &  
     \includegraphics[width=0.24\textwidth]{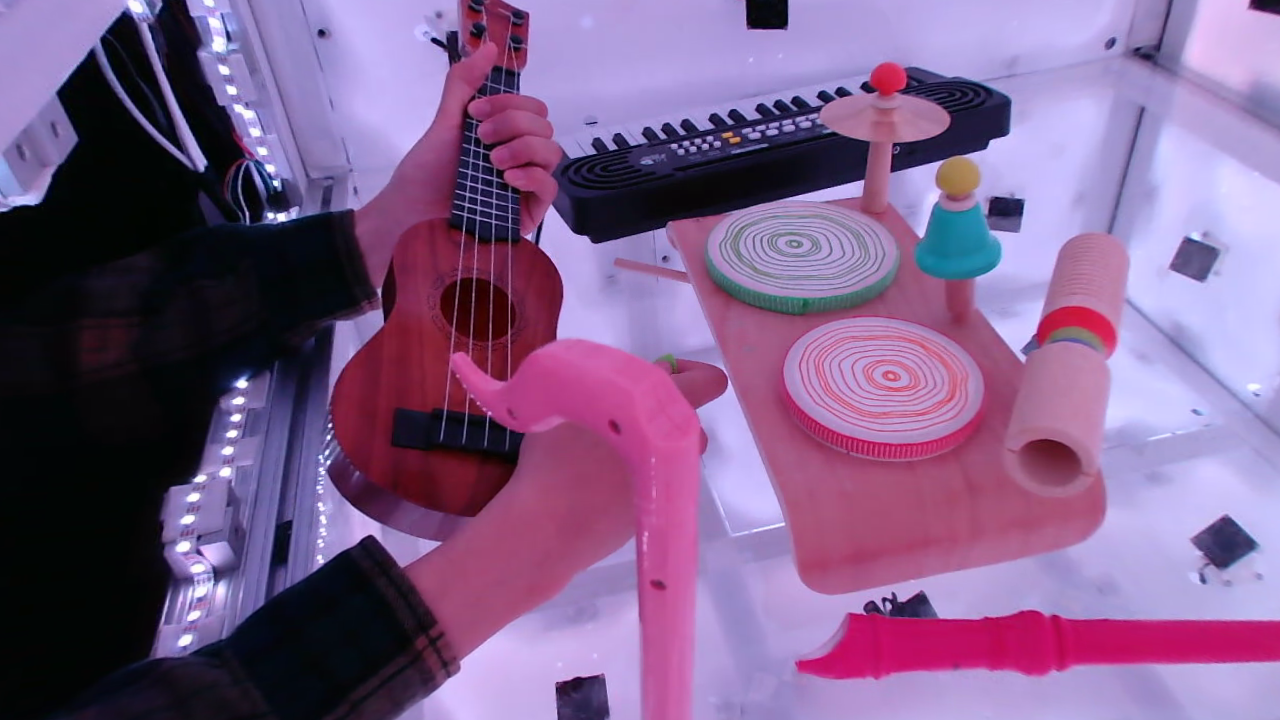}
     &
    \includegraphics[width=0.24\textwidth]{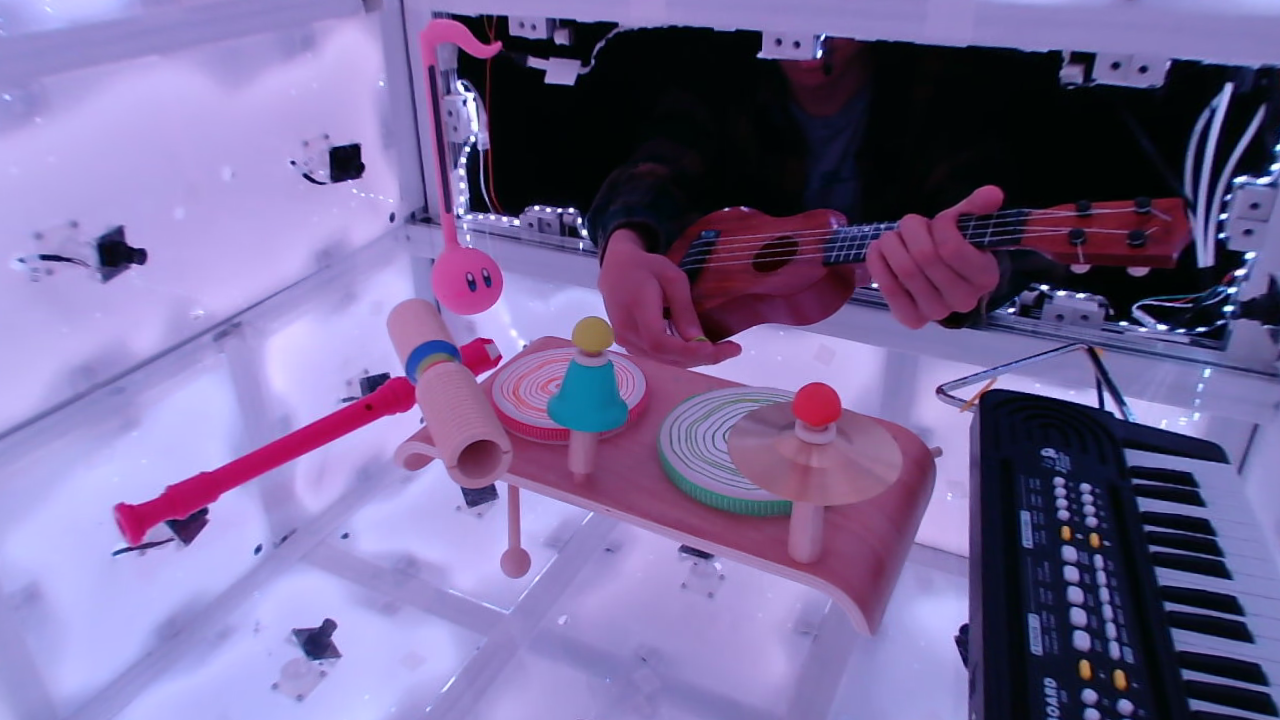}
    &
     \includegraphics[width=0.24\textwidth]{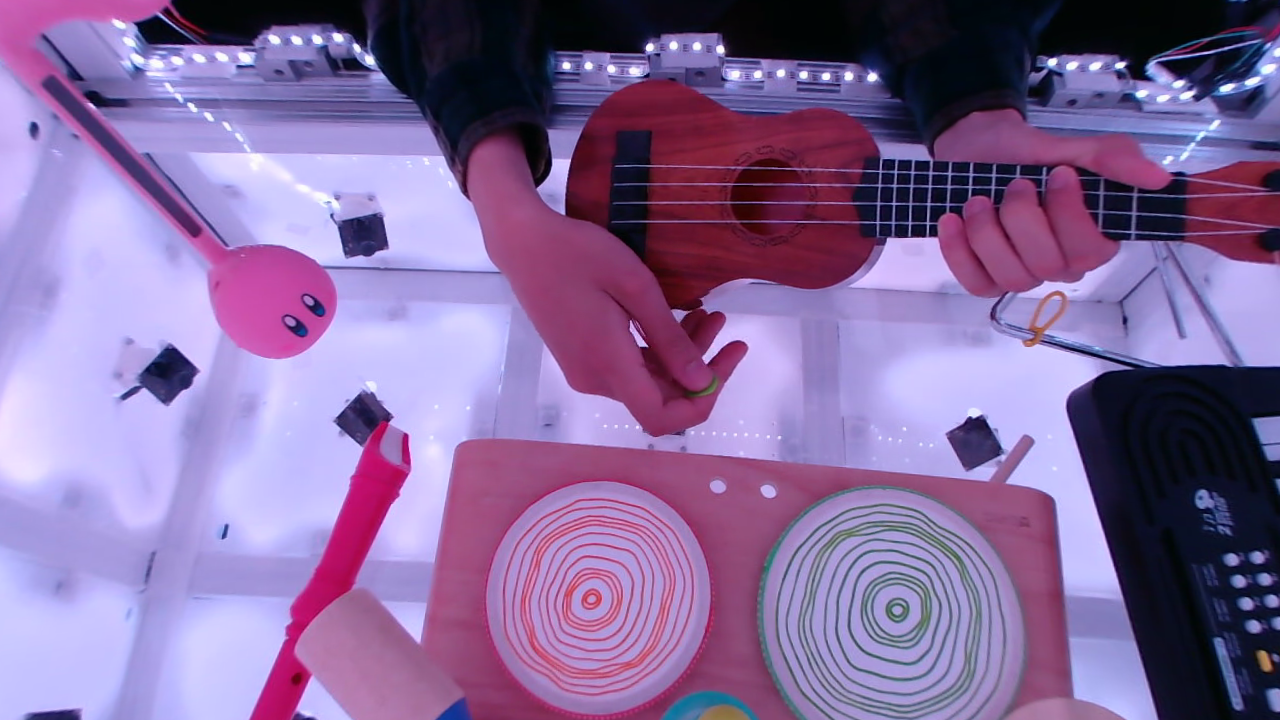}
    \\     
     \small{object masks} &  
     \includegraphics[width=0.24\textwidth]{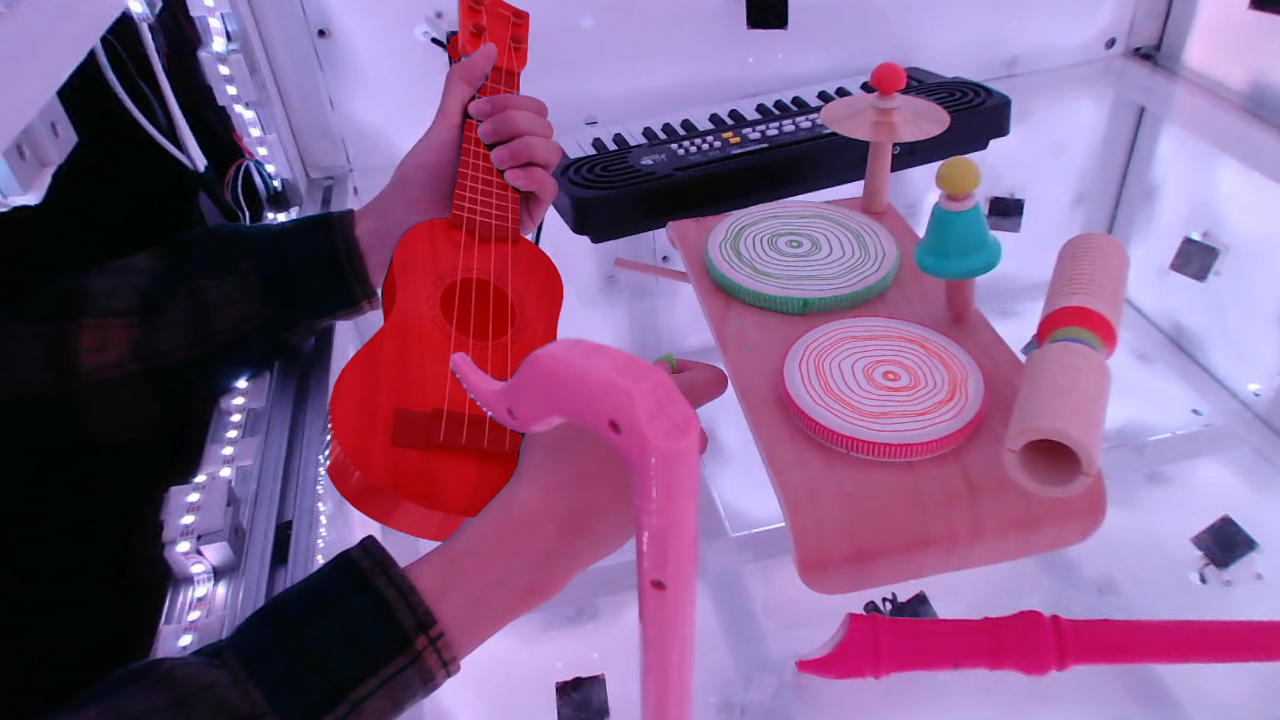}
     &
    \includegraphics[width=0.24\textwidth]{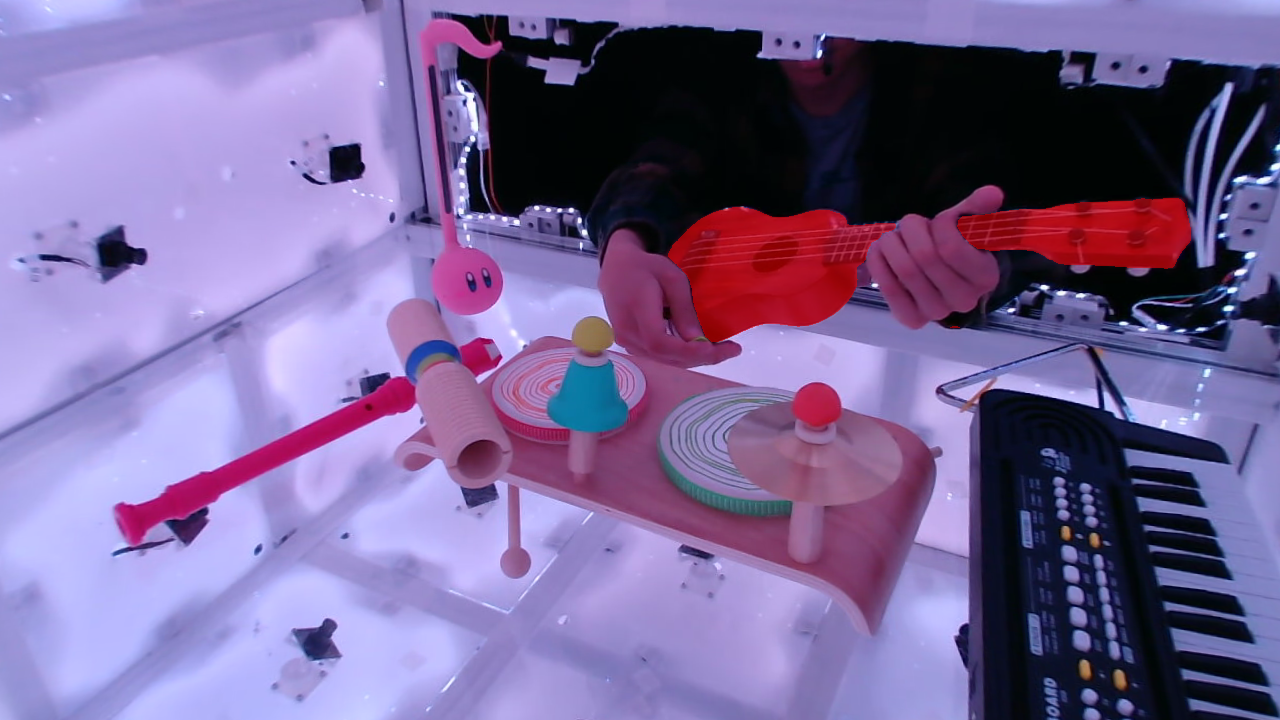}
    &
     \includegraphics[width=0.24\textwidth]{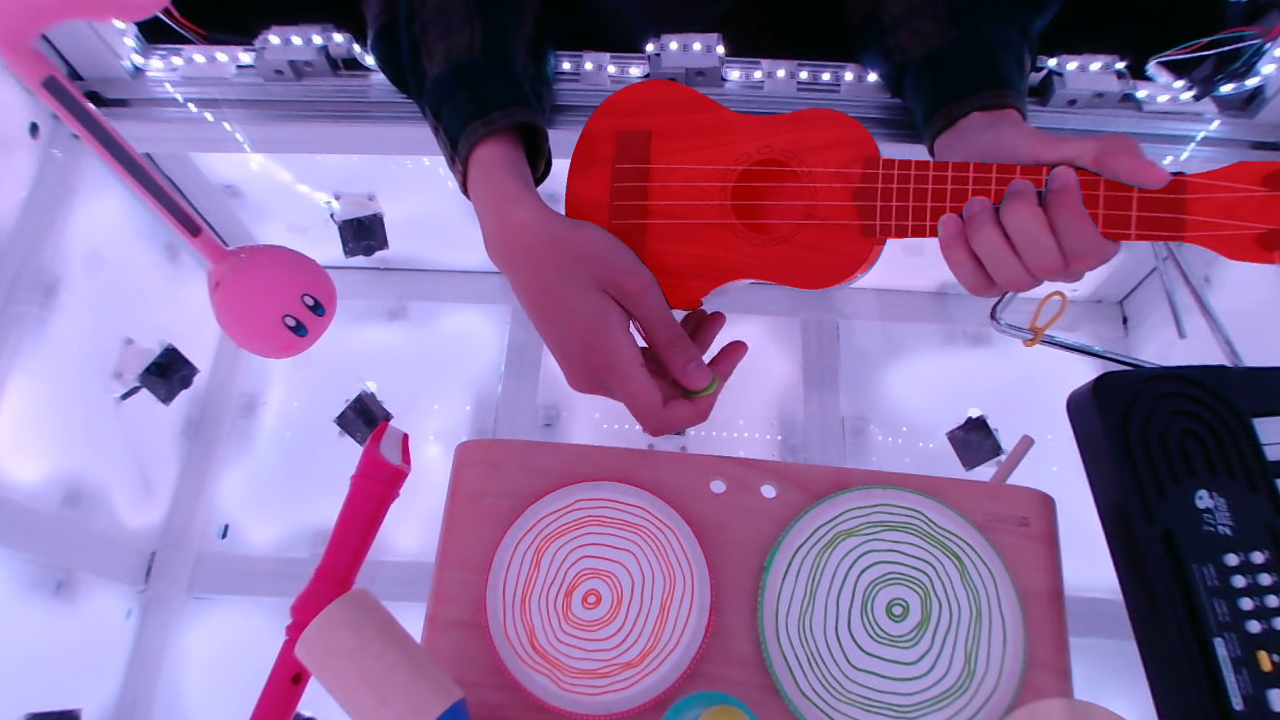}
    \\
     \small{2D hand keypoints} &  
     \includegraphics[width=0.24\textwidth]{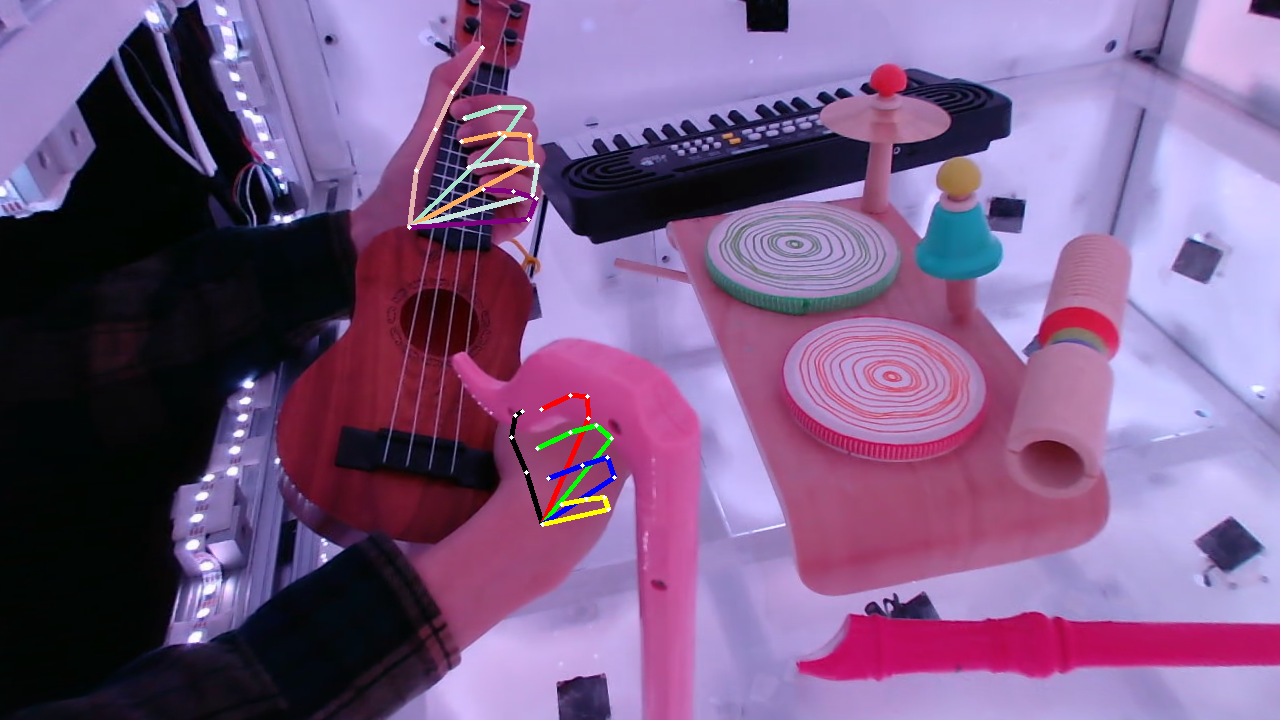}
     &
    \includegraphics[width=0.24\textwidth]{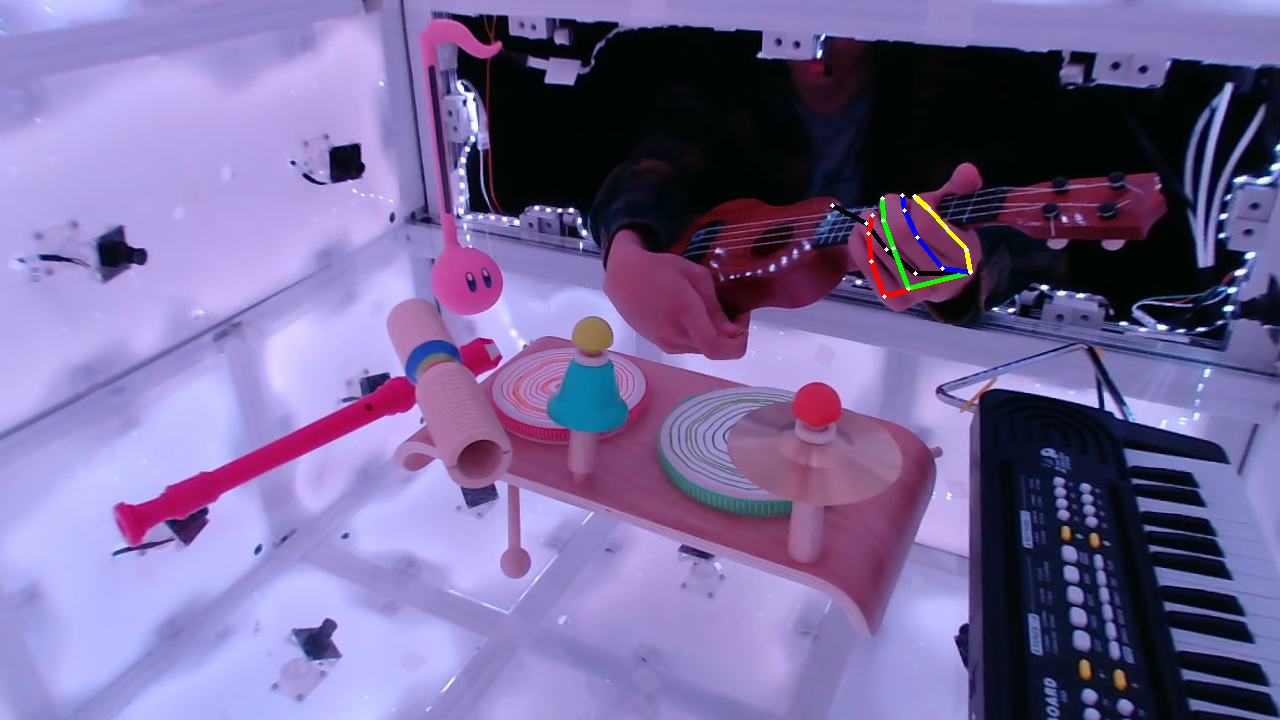}
    &
     \includegraphics[width=0.24\textwidth]{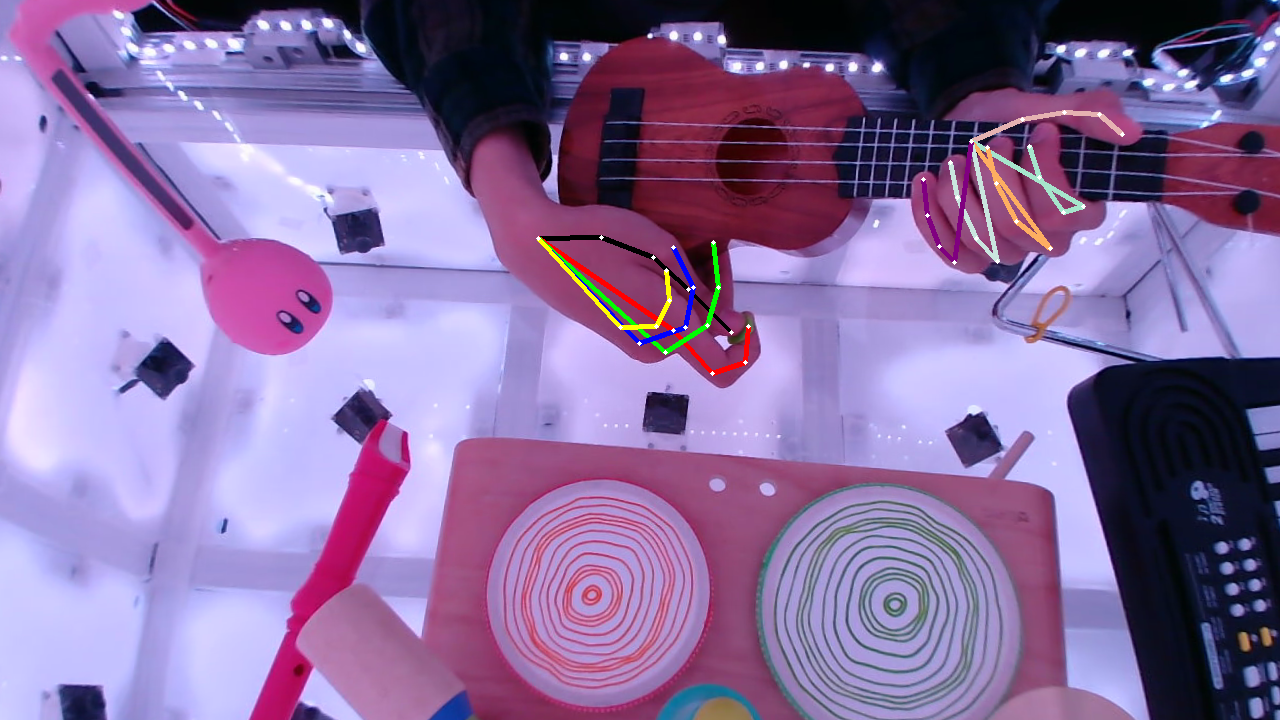}
    \\
     \small{3D hand keypoints} &  
     \includegraphics[width=0.24\textwidth]{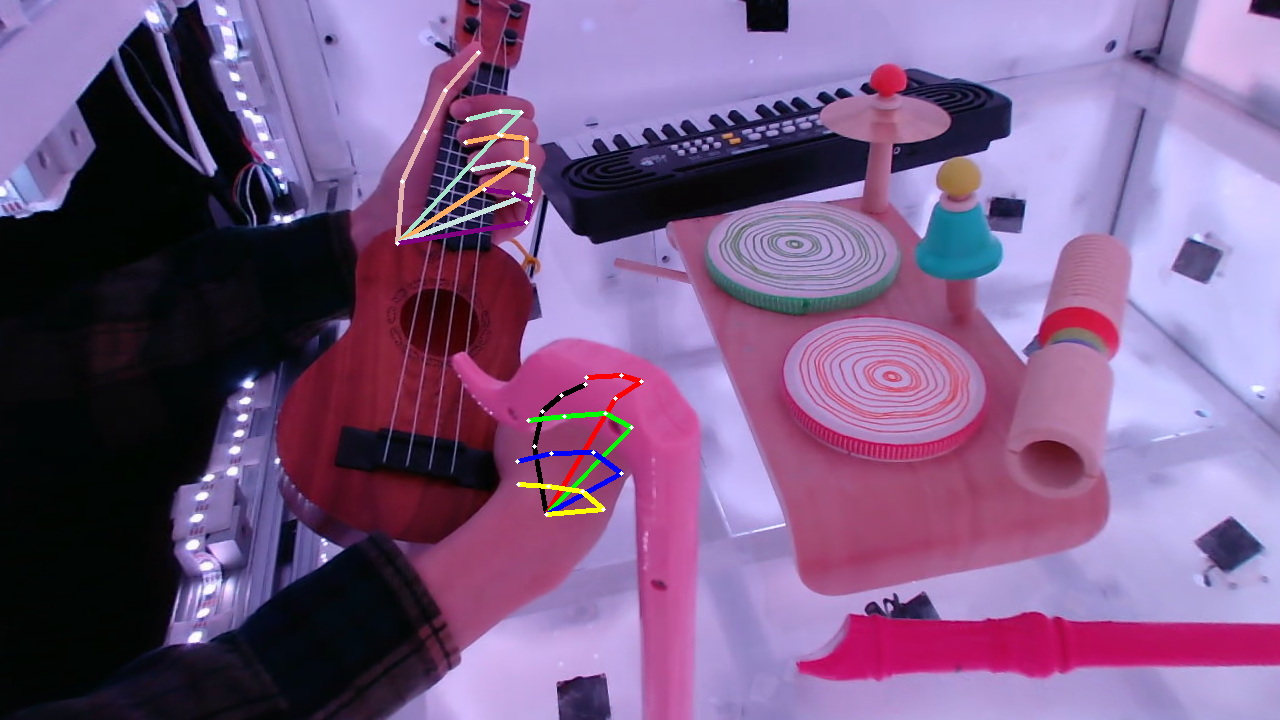}
     &
    \includegraphics[width=0.24\textwidth]{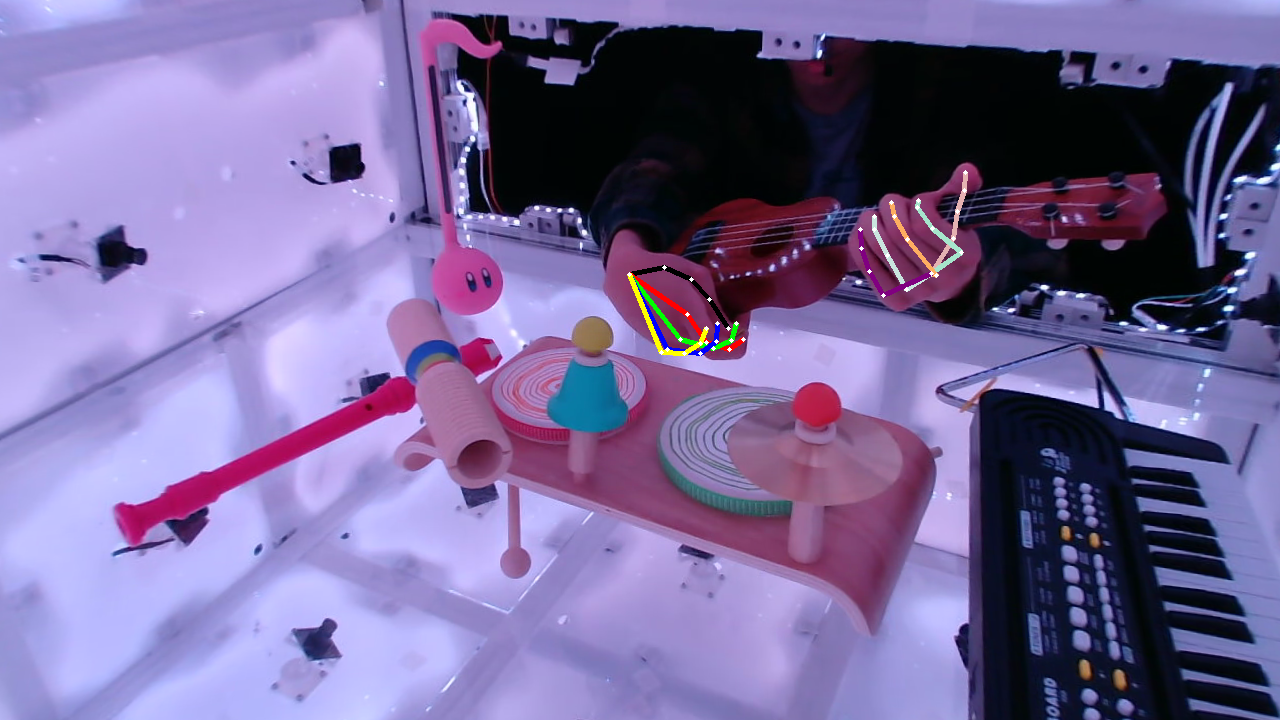}
    &
     \includegraphics[width=0.24\textwidth]{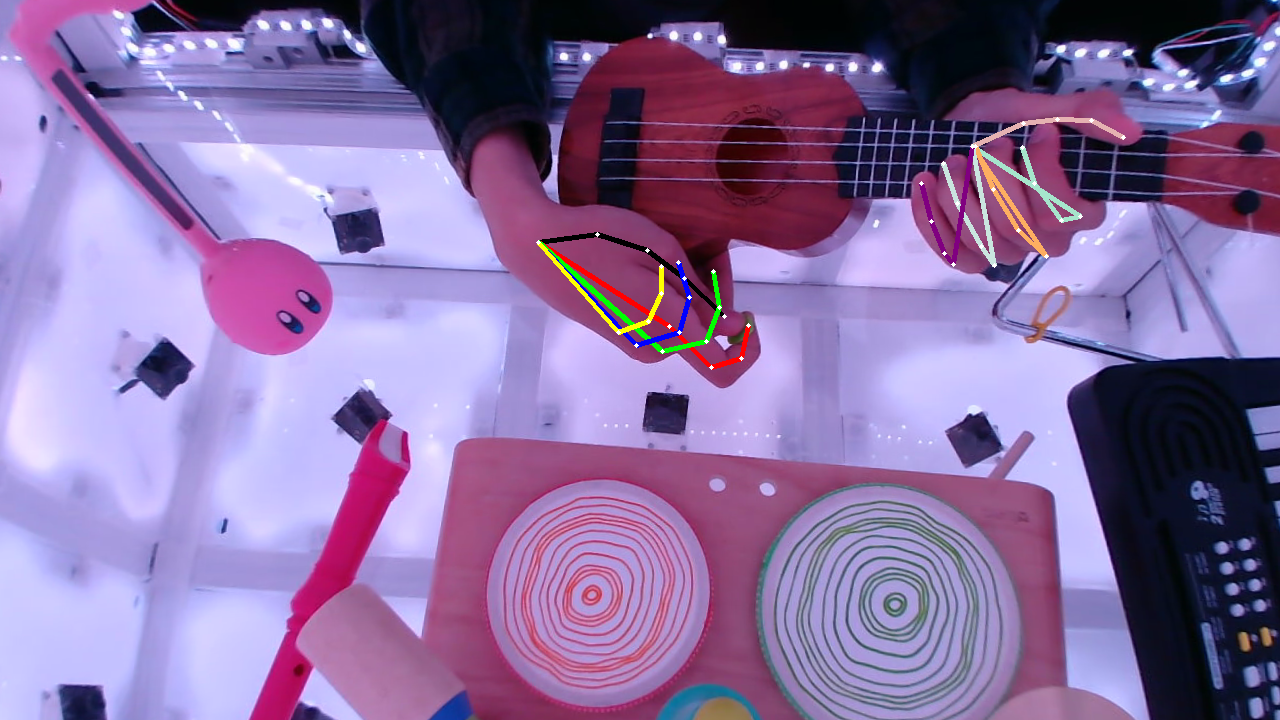}
     \\
     \small{3D hand mesh/motion} &  
     \includegraphics[width=0.24\textwidth]{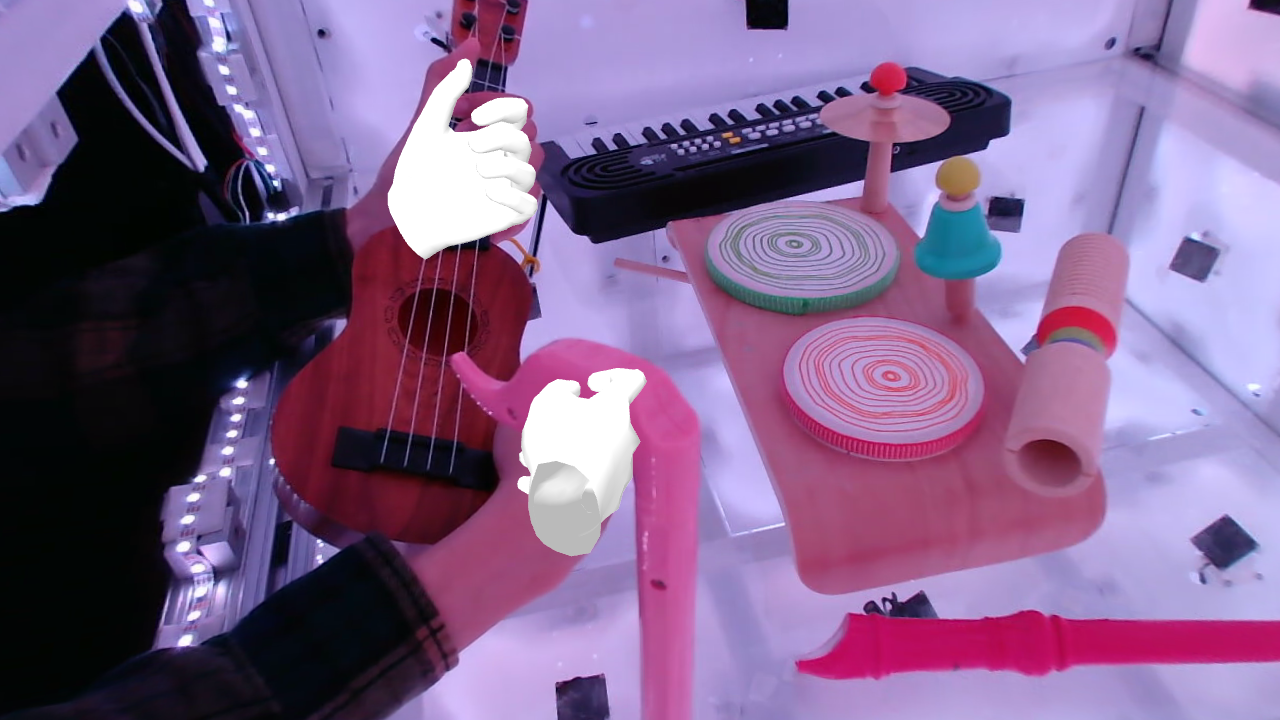}
     &
     \includegraphics[width=0.24\textwidth]{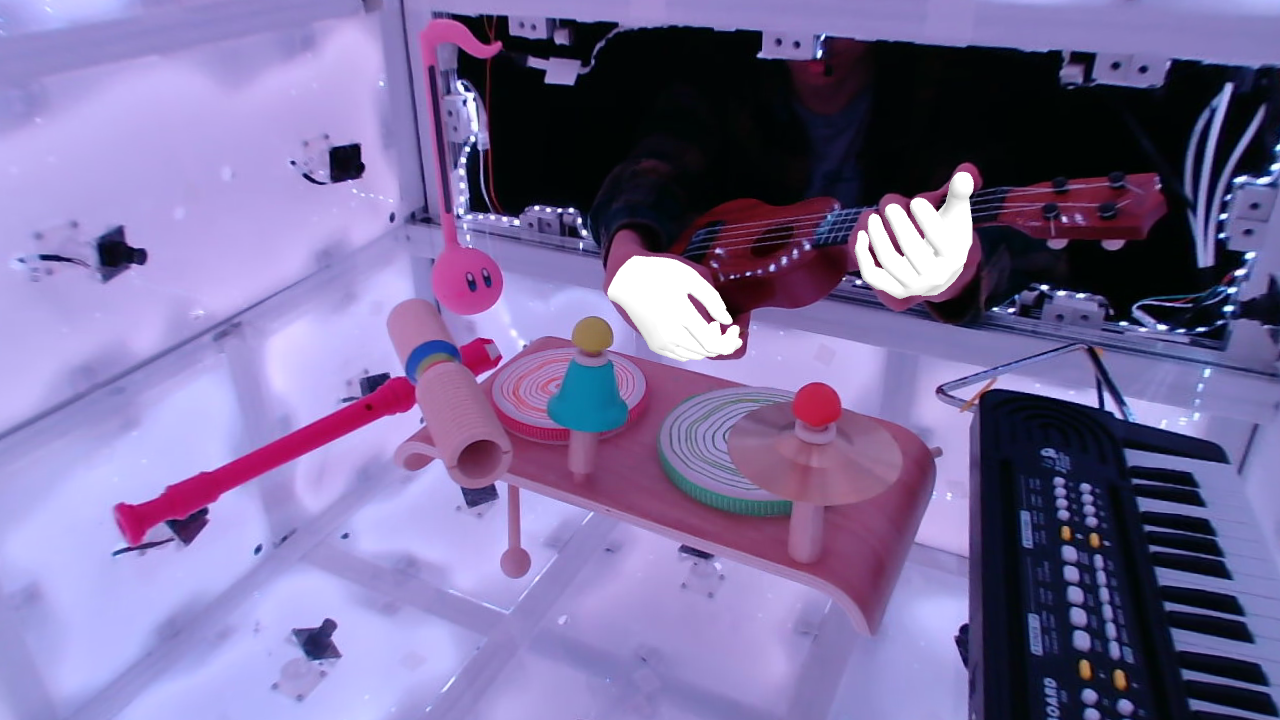}
     &
     \includegraphics[width=0.24\textwidth]{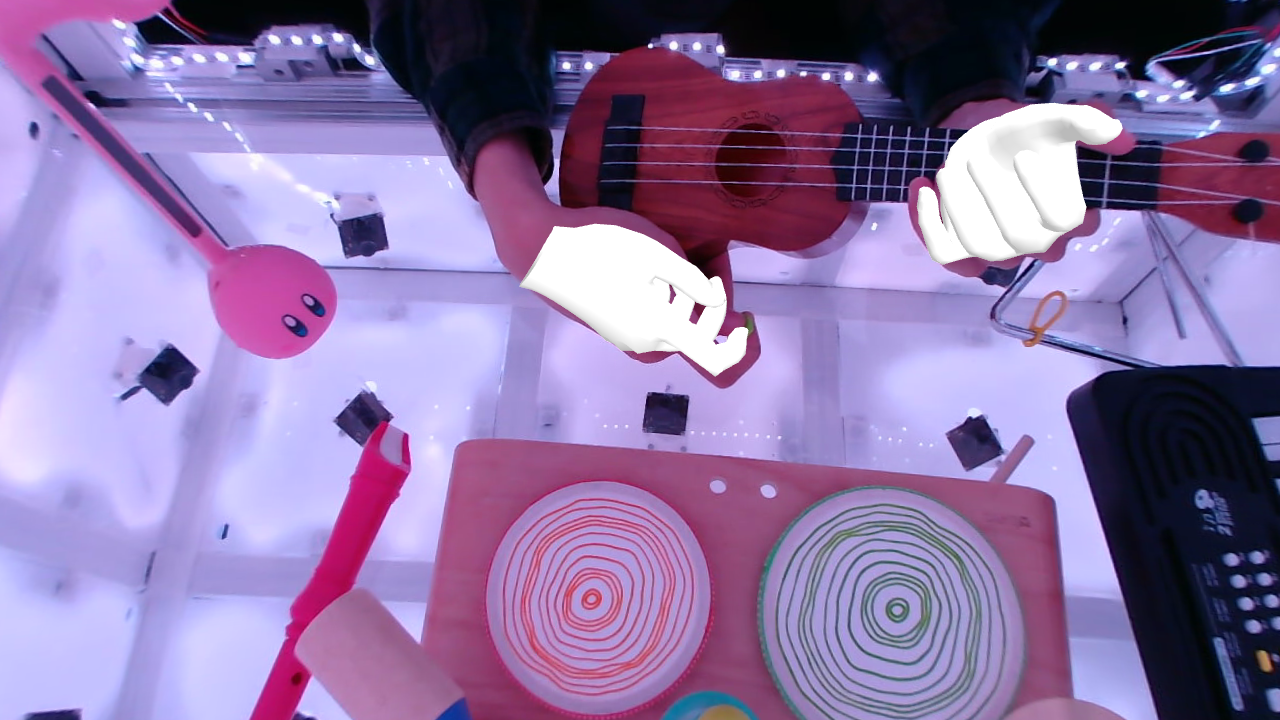}
     \\
     \small{3D object mesh/motion} &  
     \includegraphics[width=0.24\textwidth]{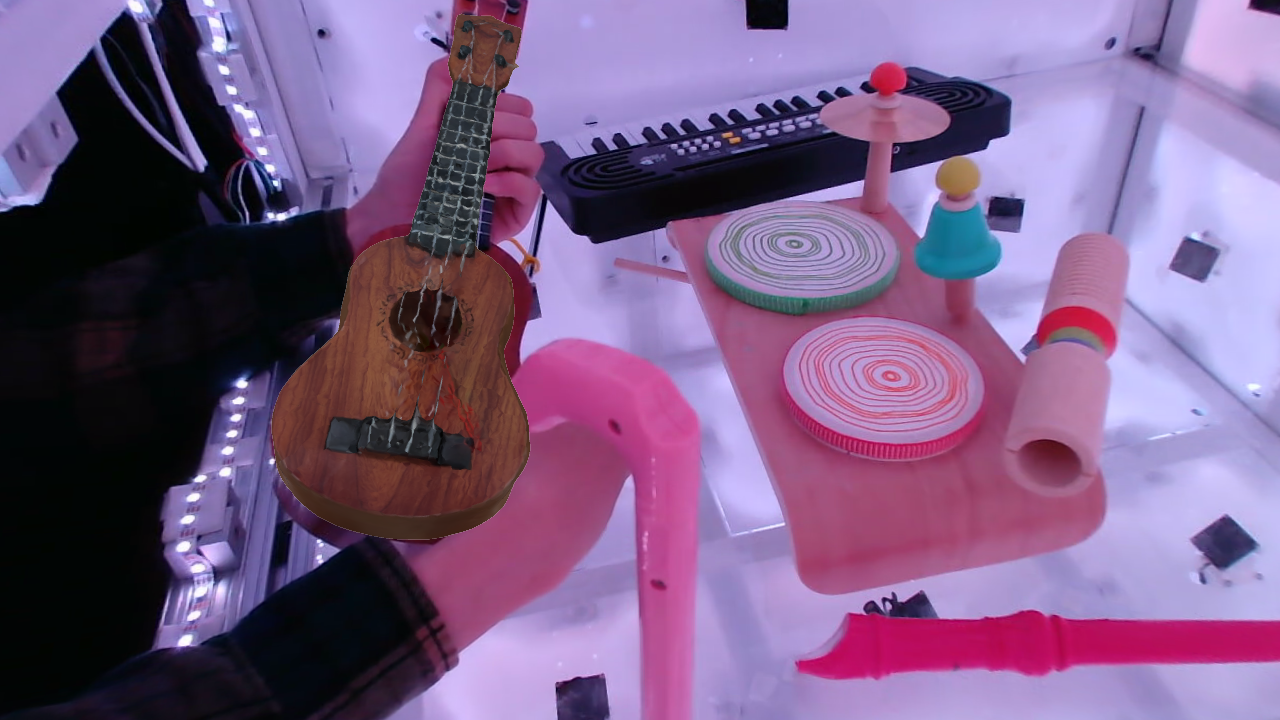}
     &
     \includegraphics[width=0.24\textwidth]{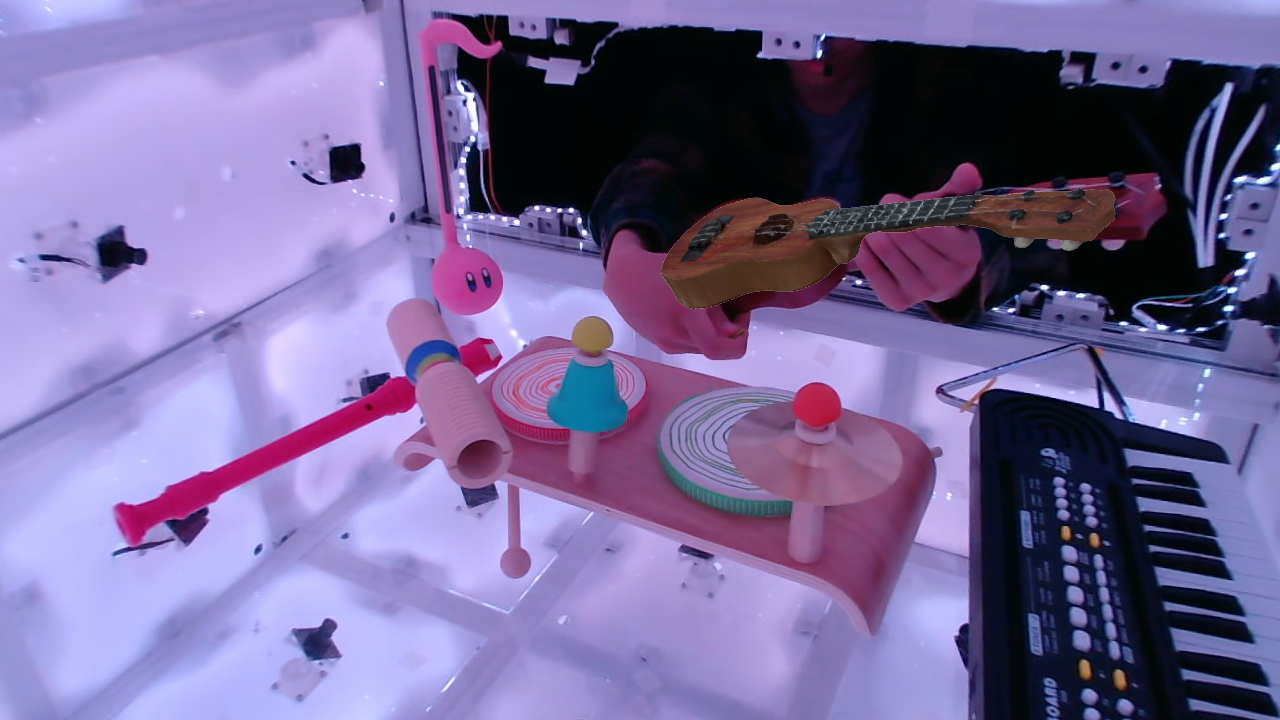}
     &
     \includegraphics[width=0.24\textwidth]{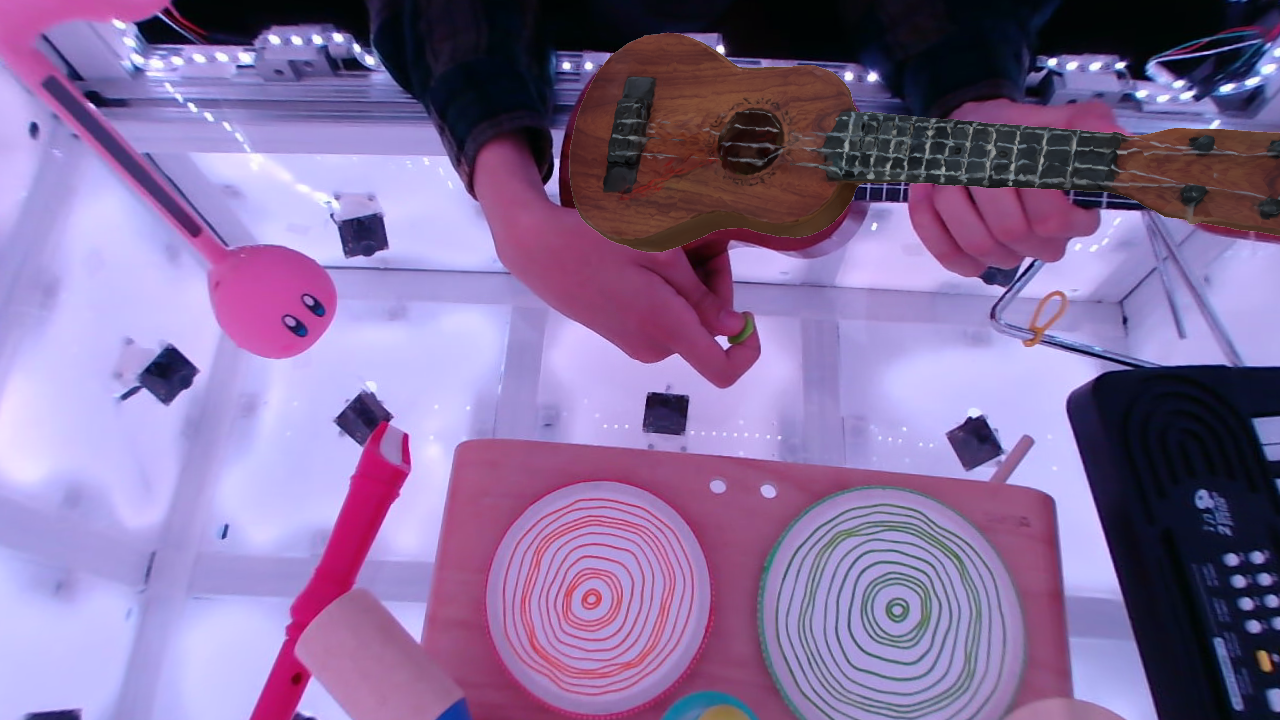}
     \\

\hline
\end{tabular}
\caption{Annotations for a single motion clip.}
\label{tab:annotation}
\end{table}

\section{Comparison of Dataset Statistics}
\begin{table}[ht]
\centering
\small
\tabcolsep 1pt
\caption{Comparisons of 3D bimanual motion datasets. Dataset names are highlighted with different colors if it has \textcolor{darkgray}{no text annotations (gray)}, \textcolor{mygreen}{action type (green)}, \textcolor{mypink}{sparse description (red)}, and \textcolor{mypurple}{dense description (blue)}.
}
\begin{tabular}{l|ccccccccccccc}
\hline
 Name&  setting & markerless & hand track. & object track. & \#mins &  \#motions &\# poses & \#views & \#frames & \#subjects & \#objects & \#verbs \\ \hline
\textcolor{mygreen}{AssemblyHands}~\cite{ohkawa2023assemblyhands}& studio & {\color{green}\ding{51}} &  semi-auto.& $/$ &630 & 62 & 203k & 12 & 3.03M & 34 & $/$ & 24\\
\textcolor{mypink}{Ego4D}\cite{jian2023affordpose} & in-the-wild & {\color{green}\ding{51}} & $/$ & $/$ & $/$ & $/$ & $/$ & 1 & $/$ & 931 & $/$ & 1452\\
 \textcolor{mypink}{Ego-Exo4D}~\cite{grauman2024ego}& in-the-wild & {\color{green}\ding{51}} & manual & $/$ &$/$ & $/$ & \textbf{4.4M} & 5-6 & $/$ & \textbf{740} & $/$  & 402\\
 \textcolor{mygreen}{HOI4D}~\cite{liu2022hoi4d}& in-the-wild & {\color{green}\ding{51}} & manual & manual &1,333 & 4k & 1.2M & 1 & 2.4M & 4 & \textbf{800} & 25\\
 \textcolor{darkgray}{ARCTIC}~\cite{fan2023arctic}& studio & {\color{red}\ding{55}} & mocap & mocap &121  & 339 & 218k  & 9 & 2.1M & 10 & 11 & {\color{red}\ding{55}}\\
 \textcolor{mygreen}{TACO}~\cite{liu2024taco}& studio & {\color{red}\ding{55}} &  mocap & mocap &202 & 2.3k & 363k & 13 & 4.7M & 14 &  196 & 13\\
 \textcolor{mypink}{OakInk2}~\cite{zhan2024oakink2}& studio & {\color{red}\ding{55}} & mocap & mocap & 557 & 2.8k & 993k & 4 & 4.01M & 9& 75 & 55\\
  \textcolor{darkgray}{HOT3D}~\cite{banerjee2024hot3d}& studio & {\color{red}\ding{55}} &  mocap & mocap &833  & 4.1k & 1.7M & 2-3 & 3.7M & 19 & 33 & {\color{red}\ding{55}} \\
 \textcolor{mypurple}{\methodname (Ours)} & studio & {\color{green}\ding{51}} & auto & auto &\textbf{2,034} & \textbf{13.9k} & {3.7M} & \textbf{51} & \textbf{183M} & {56} & 417 & \textbf{1467} \\

 \hline
\end{tabular}
\label{tab:supp_stats}
\end{table}

\Cref{tab:supp_stats} provides a comprehensive comparison of 3D bimanual motion datasets, including their capturing sources, annotation methods, and statistics across various features. The table applies the same method as Figure 2 in the main paper for verb counting. To compile the verb counts, we extracted verbs from several sources: fine-granularity labels from AssemblyHands~\cite{ohkawa2023assemblyhands}, atomic descriptions from Ego-Exo4D~\cite{grauman2024ego}, redacted narrations from Ego4D~\cite{grauman2022ego4d}, category labels and task definitions from HOI4D~\cite{liu2022hoi4d}, action types from TACO~\cite{liu2024taco}, task descriptions and affordance annotations from OakInk2, and our own augmented text descriptions from \methodname{}. For verb extraction, we parsed sentences using spaCy~\cite{spacy2} and collected the verb stems. Verb stems that were misspelled or not recognized as verbs in either WordNet~\cite{miller1995wordnet} or spaCy's `en\_core\_web\_sm' model were removed. Clearly, \methodname{} surpasses all other datasets in filming length, number of motion sequences, number of camera views, frame count, and verb count.

\section{Experiments on Text-driven Motion Synthesis}

\paragraph{Implementation Details}

Our framework builds upon the T2M-GPT architecture proposed by \cite{zhang2023generating} and utilizes their publicly available PyTorch codebase\footnote{\url{https://github.com/Mael-zys/T2M-GPT}}. The network comprises a motion VQ-VAE that learns a mapping between motion data and discrete code sequences and a T2M-GPT which generates code indices conditioned on the text description.
For the Motion VQ-VAE, the codebook size is set to $512\times 512$, with a downsampling rate $l=4$. The encoder and decoder is a simple convolutional architecture consisting of 1D convolutions, residual blocks \cite{he2016deep}, and ReLU activations. Temporal downsampling and upsampling are performed using strided convolutions (stride = 2) and nearest-neighbor interpolation, respectively. We use the AdamW optimizer \cite{loshchilov2017decoupled} with $[\beta_1, \beta_2]=[0.9,0.99]$, a batch size of 256, and an exponential moving constant $\lambda=0.99$. The model is trained for the first 200K iterations with a learning rate of $2\times 10^{-4}$, followed by 100K iterations with a reduced learning rate of $1\times 10^{-5}$. 
Based on our text annotations, we construct a custom word vectorizer utilizing pre-trained 300-dimensional word embedding vectors from GloVe \cite{pennington2014glove}.

For the T2M-GPT, we employ adopts a transformer architecture \cite{vaswani2017attention} consisting of 18 layers with a model dimensionality of 1,024 and 16 attention heads. The maximum length of the code index sequence is capped at 50, with an additional end token incorporated to indicate sequence termination. The optimization of the transformer is performed using the AdamW algorithm \cite{loshchilov2017decoupled}, configured with hyperparameters $[\beta_1, \beta_2]=[0.5,0.99]$ and a batch size of 128. The training process spans 150K iterations with an initial learning rate of $1\times 10^{-4}$, followed by a decay to $5\times 10^{-6}$ over an additional 150K iterations.

To ensure consistency across three datasets—TACO \cite{liu2024taco}, OakInk2 \cite{zhan2024oakink2}, and \methodname{}—we standardize the hand representations and motion ranges. First, we extract 3D hand keypoints from the MANO \cite{MANO:SIGGRAPHASIA:2017} parameters using the MANO Layer\footnote{\url{https://github.com/hassony2/manopth}}. Subsequently, we align the hand orientations such that the fingertips point in the positive z-axis direction. Additionally, we recenter each motion sequence by adjusting the hand positions so that the motion center of both hands is aligned to the origin.

\paragraph{Evaluation Metrics}

Given the lack of a standard hand motion feature extractor, we train a simple framework consisting of a motion extractor and a text extractor trained under a contrastive learning paradigm, following \cite{Guo_2022_CVPR}. The Motion and Text Feature Extractors are designed to learn geometrically close feature vectors for matched text-motion pairs while ensuring separation for mismatched pairs. Specifically, the input text and motion are encoded into semantic vectors $F_t$ and $F_m$, respectively, using two distinct bi-directional GRUs. To achieve this, we minimize a contrastive loss that enforces proximity for matched pairs and imposes a margin of separation $m$ for mismatched pairs:

\begin{equation} 
L_{cst} = y \cdot \left(\max(0, m - ||F_t - F_m||_2^2)\right)^2 + (1-y) \cdot ||F_t - F_m||_2^2, 
\end{equation}
where $y \in {0,1}$, with $y = 1$ indicating matched text-motion pairs and $y = 0$ otherwise. The margin $m$ is set to 10 across all datasets. These feature extractors are independently trained for each dataset to establish an upper bound respectively.

The trained text and motion feature extractors are utilized to evaluate text-to-motion generation using the metrics proposed in \cite{Guo_2022_CVPR}. We denote the ground-truth motion features, generated motion features, and text features derived from the aforementioned feature extractors as $F_{gt}$, $F_{gen}$, and $F_t$, respectively.

\textbf{R Precision.} 
For each generated motion, a text pool is formed with its ground truth text and 31 randomly selected mismatched texts. Euclidean distances between the description feature and motion features are computed and ranked. If the ground truth text ranks in the top-k positions (k=1, 2, 3), it counts as a successful retrieval. The average accuracy over all samples defines the top-k R-precision.

\textbf{Multimodal Distance (MM Dist.)}
MM Dist. evaluates the alignment between text embeddings and generated motion features. Given $N$ randomly generated samples, it calculates the average Euclidean distance between each text feature and the corresponding generated motion feature:
\begin{equation}
\text{MM Dist.} = \frac{1}{N}\sum_{i=1}^{N}\lVert F_{gen}^i - F_{t}^i\rVert
\label{eq:mm-dis}
\end{equation}
where $F_{gen}^i$ and $F_{t}^i$ are the features of the i-th text-motion pair. 

\textbf{Frechet Inception Distance (FID).} 
FID measures the distributional similarity between real and generated motion features. Features are extracted from ground-truth motions in the test set and generated motions from the corresponding descriptions. FID is computed as:
\begin{equation}
\text{FID} = \lVert \mu_{gt} - \mu_{gen}\rVert^2 - \text{Tr}(\sigma_{gt} + \sigma_{gen} - 2(\sigma_{gt}\sigma_{gen})^{\frac{1}{2}}),
\label{eq:fid}
\end{equation}
where $\mu_{gt}$ and $\mu_{pred}$ are mean of $F_{gt}$ and $F_{gen}$. $\sigma$ is the covariance matrix and $\text{Tr}$ denotes the trace of a matrix.

\textbf{Diversity.} 
Diversity quantifies the variance across all generated motion sequences in the dataset. From the generated motions, $S_{dis}$ pairs of motion features are randomly sampled, denoted as $F_{gen}^i$ and $F_{gen}^{i'}$. The diversity is then calculated as:
\begin{equation}
\text{Diversity} = \frac{1}{S_{dis}}\sum_{i=1}^{S_{dis}}||F_{gen}^i - F_{gen}^{i'}||,
\label{eq:diversity}
\end{equation}
In our experiments, we set $S_{dis} = 300$, as suggested in \cite{Guo_2022_CVPR}.

\textbf{MultiModality.} 
MultiModality assesses the diversity of hand motions generated from the same text description. For the $i$-th text description, 20 motions are generated, and two subsets, each containing 10 motions, are sampled. Denoting the features of the $j$-th pair for the $i$-th text description as $(F_{gen}^{i,j}, F_{gen}^{i,j'})$, MultiModality is computed as:
\begin{equation}
\text{MultiModality} = \frac{1}{10N}\sum_{i=1}^{N}\sum_{j=1}^{10}\lVert F_{gen}^{i,j} - F_{gen}^{i,j'}\rVert
\label{eq:mmodality}
\end{equation}

\begin{table}[t!]
    \caption{Quantitative results
    for text-driven motion synthesis with different backbones trained on our dataset. \emph{upper bound} indicates performance calculated with the ground truth. We repeat the evaluation 20 times and report the averge with 95\%  confidence interval.}
    \centering
    \begin{tabular}{l c c c c c c c}
    \toprule
     \multirow{2}{*}{Dataset}  & \multicolumn{3}{c}{R Precision(\%)$\uparrow$} & \multirow{2}{*}{MM Dist.$\downarrow$} & \multirow{2}{*}{FID$\downarrow$} & \multirow{2}{*}{Diversity$\rightarrow$} & \multirow{2}{*}{MultiModality$\uparrow$} \\

    \cline{2-4}
       ~ & @1 & @2 & @3 \\
    
    \midrule

    {upper bound} & \et{77.4}{.002} & \et{88.8}{.002} & \et{91.3}{.001} & \et{2.96}{.005} & \et{0.002}{.000} & \et{11.9}{.097} & - \\
    MDM~\cite{tevet2023human} & \et{22.5}{.004} & \et{42.7}{.005} & \et{50.2}{.005} & \et{7.81}{.082} & \et{5.60}{.126} & \et{9.8}{.088} & \et{8.52}{.103} \\
    TM2T~\cite{chuan2022tm2t} & \et{24.1}{.002} & \et{38.4}{.003} & \et{47.1}{.004} & \et{9.28}{.017} & \et{8.60}{.161} & \et{9.7}{.065} & \et{6.29}{.085} \\
    T2M-GPT~\cite{zhang2023generating} & \et{\textbf{31.2}}{.003} & \et{\textbf{44.7}}{.004} & \et{\textbf{53.1}}{.004} & \et{\textbf{6.68}}{.023} & \et{\textbf{4.70}}{.078} & \et{\textbf{10.5}}{.090} & \et{\textbf{9.11}}{.255} \\
    \bottomrule
    \end{tabular}

    \label{tab:quant_t2m_backbone}
\end{table}

\paragraph{Comparison of Different Backbones.}

We evaluate the text-to-hand motion synthesis performance on our dataset using three different backbone models: TM2T~\cite{chuan2022tm2t}, MDM~\cite{tevet2023human}, and T2M-GPT~\cite{zhang2023generating}, as summarized in \Cref{tab:quant_t2m_backbone}. TM2T comprises a motion VQ-VAE module for motion quantization and an attentive GRU-based model for text-to-motion generation. MDM employs a classifier-free diffusion generative approach. While these models were originally developed for human motion synthesis, we adapted them to generate hand motions on our dataset.

From \Cref{tab:quant_t2m_backbone}, it is evident that our dataset supports all three backbones effectively. Notably, the state-of-the-art T2M-GPT architecture for human motion generation also achieves the best performance on our dataset, demonstrating its superior capability for motion synthesis.

\paragraph{Ablations of Different Motion Representations and Text Annotations.}
\begin{table}[t!]
    \caption{Ablation study on
    different hand motion representations and text annotations for text-driven motion synthesis.\emph{KP} refers to the 3D hand keypoints representation, while \emph{6D} denotes the MANO pose parameters encoded in the 6D representation. Numbers in parentheses indicate the quantity of text scripts. \emph{upper bound} indicates performance calculated with the ground truth. We repeat the evaluation 20 times and report the averge with 95\%  confidence interval.}
    \centering
    \begin{tabular}{l c c c c c c c}
    \toprule
     \multirow{2}{*}{Dataset}  & \multicolumn{3}{c}{R Precision(\%)$\uparrow$} & \multirow{2}{*}{MM Dist.$\downarrow$} & \multirow{2}{*}{FID$\downarrow$} & \multirow{2}{*}{Diversity$\rightarrow$} & \multirow{2}{*}{MultiModality$\uparrow$} \\

    \cline{2-4}
       ~ & @1 & @2 & @3 \\
    
    \midrule

    {upper bound} & \et{73.3}{.003} & \et{87.0}{.002} & \et{92.1}{.001} & \et{3.29}{.006} & \et{0.002}{.000} & \et{12.8}{.067} & - \\
    6D (14k) & \et{26.5}{.003} & \et{43.3}{.004} & \et{52.3}{.003} & \et{7.56}{.051} & \et{5.11}{.183} & \et{9.7}{.078} & \et{7.28}{.059} \\
    6D (84k) & \et{29.5}{.003} & \et{\textbf{46.9}}{.004} & \et{\textbf{57.3}}{.004} & \et{7.04}{.041} & \et{\textbf{4.34}}{.221} & \et{\textbf{11.7}}{.068} & \et{9.08}{.060} \\
    \hline
    {upper bound} & \et{77.4}{.002} & \et{88.8}{.002} & \et{91.3}{.001} & \et{2.96}{.005} & \et{0.002}{.000} & \et{11.9}{.097} & - \\
    KP (14k) & \et{27.2}{.003} & \et{40.1}{.002} & \et{49.7}{.004} & \et{7.13}{.020} & \et{5.20}{.169} & \et{8.7}{.067} & \et{7.29}{.085} \\
    KP (84k)  & \et{\textbf{31.2}}{.003} & \et{44.7}{.004} & \et{53.1}{.004} & \et{\textbf{6.68}}{.023} & \et{4.70}{.078} & \et{10.5}{.090} & \et{\textbf{9.11}}{.255} \\
    \bottomrule
    \end{tabular}

    \label{tab:quant_t2m_ablation}
\end{table}

We further explore the performance of text-to-motion generation using different hand motion representations in \Cref{tab:quant_t2m_ablation}. One approach leverages the 3D hand keypoints derived from MANO parameters, represented as $X \in \mathbb{R}^{42 \times 3}$. The other employs the MANO hand pose parameters $\mathbb{\theta} \in \mathbb{R}^{16 \times 6}$, encoded in a 6D representation \cite{zhou2019continuity}. We also conduct an ablation study to evaluate the impact of the number of text descriptions on text-to-motion generation. In \methodname{}, each motion clip is paired with six distinct text descriptions, resulting in a total of 84k annotations. We compare the generation results using 14k annotations versus the full set of 84k annotations, as presented in \Cref{tab:quant_t2m_ablation}.

The results demonstrate that both the 6D representation and the 3D keypoint representation achieve comparable performance in text-driven hand motion synthesis. This finding suggests that both representations are equally capable of capturing the essential hand motion features required for this task. Nevertheless, incorporating additional text annotations significantly boosts performance across all evaluation metrics. This improvement underscores the importance of enriched textual descriptions in enhancing the semantic alignment between textual inputs and generated hand motions, irrespective of the chosen motion representation.

\begin{figure}[htbp]
    \centering
    \includegraphics[width=0.98\linewidth]{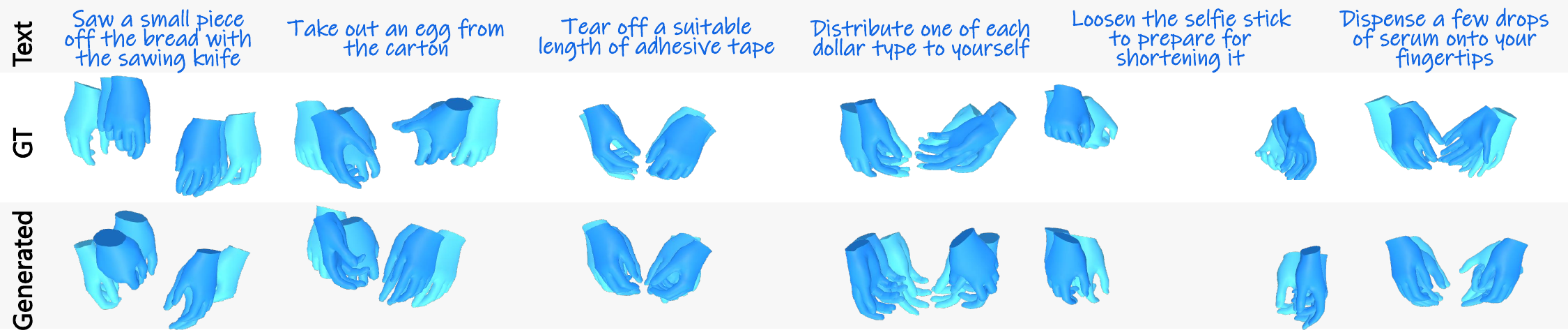}
    \includegraphics[width=0.98\linewidth]{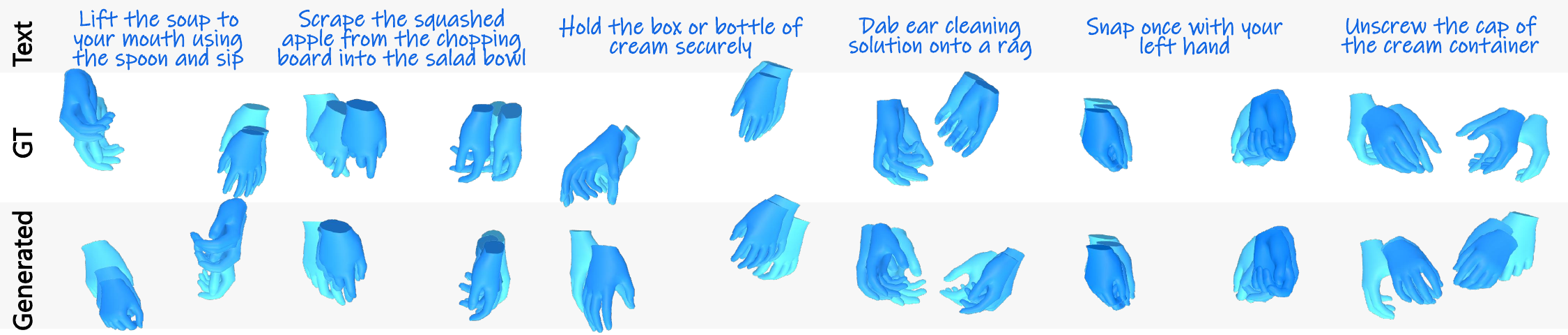}
    \caption{More qualitative results for text-driven motion synthesis on \methodname{}. Darker color indicates later frame in the sequence.}
    \label{fig:more_t2m_res}
\end{figure}

\paragraph{More Qualitative Results.}

We present additional qualitative results from our testing sets to demonstrate the effectiveness of text-driven hand motion synthesis in \Cref{fig:more_t2m_res}. These examples further highlight the ability of our approach to generate semantically aligned hand motions from diverse textual inputs.
\begin{table}[h]
    \centering
    \begin{tabular}{|c|c|c|c|}
        \hline
        \multirow{2}{*}{test\textbackslash train} & \multicolumn{3}{c|}{(FID / Diversity)} \\
        \cline{2-4}
        & GigaHands & Oakink2 & TACO \\
        \hline
        GigaHands & 6.61 / 10.5 & 44.8 / 4.01 & 52.1 / 7.11 \\
        \hline
        Oakink2 & 19.1 / 6.45 & 19.6 / 6.88 & 58.1 / 5.04 \\
        \hline
        TACO & 22.5 / 11.5 & 33.6 / 9.31 & 11.0 / 11.1 \\
        \hline
    \end{tabular}
    \caption{Cross-dataset evaluation of text-to-motion models trained on different datasets. Each column represents the training dataset, while each row shows the evaluation results on a different dataset. Metrics include FID and Diversity, computed on the test set feature extractor. The ground truth Diversity metrics for GigaHands, Oakink, and TACO are 11.9, 9.30, and 14.2, respectively.}
    \label{tab:t2m_cross}
\end{table}

\paragraph{Cross Validation on different datasets.} To validate the capacity of text-to-To evaluate the generalization ability of text-to-motion generation models trained on \methodname, we conduct cross-dataset validation. Specifically, we train T2M-GPT models on one dataset and evaluate their performance on others. Table \ref{tab:t2m_cross} presents the quantitative results, where each column indicates the dataset used for training, and each row corresponds to the dataset on which the model is tested. We report the FID and Diversity metrics, computed on the corresponding test dataset's feature extractor.

Since the triplet-based textual annotations in the TACO dataset are coarse, with each triplet corresponding to multiple motion sequences, we employ ChatGPT-4o \cite{achiam2023gpt} to refine the captions during testing and select the best-performing generation for each triplet. The results in Table \ref{tab:t2m_cross} demonstrate that our models achieve competitive performance on unseen datasets, despite being trained on a single dataset. However, due to the simplistic nature of TACO’s annotations, models trained on other datasets struggle to perform well when tested on TACO.

For OakInk2, object descriptions rely on material and texture distinctions rather than explicit shape descriptions. For example, "yellow striped bottle" refers to a large plastic jar, whereas "white bottle" describes a small bottle. This captioning style differs from the textual descriptions in our dataset, leading to a slight performance drop when models trained on our dataset are tested on OakInk2.

\section{Experiments on Motion Captioning}

\paragraph{Implementation Details.} Our motion captioning framework is also built upon the TM2T \cite{chuan2022tm2t} architecture, leveraging the same VQ-VAE for motion quantization as used in the text-driven motion synthesis task. For tokenized motion representation, we employ a transformer model to efficiently map hand motions to textual descriptions. The transformers have 4 attention layers, both with 8 attention heads with 512 hidden size. 

When evaluate captioning performance across datasets, we train the VQ-VAE and motion-to-text models separately on TACO \cite{liu2024taco}, OakInk2 \cite{zhan2024oakink2}, and \methodname{}. For testing the model's ability in in-the-wild motion captioning, we first train a VQ-VAE using all datasets combined, followed by training the motion-to-text model exclusively on \methodname{}. When generating captions for other datasets, tokenized motion sequences from the combined VQ-VAE are processed by our motion-to-text model, enabling consistent inference across diverse motion data.

\paragraph{Evaluation Metrics.} We use \textbf{R Precision} and \textbf{Multimodal Distance (MM Dist.)} to quantitatively measure the performance of our motion-to-text mapping. Unlike in text-to-motion tasks, where motion features are used to retrieve text, we reverse the process by using text features to retrieve the corresponding motion. This adaptation ensures the metrics effectively measure the alignment in motion-to-text tasks. For linguistic evaluation, we use the NLPEval codebase\footnote{\url{https://github.com/Maluuba/nlg-eval}} to compute BLEU~\cite{papineni2002bleu} and ROUGE~\cite{lin2004rouge}. To assess text diversity, we compute Distinct-n\cite{li2015diversity}, which evaluates diversity by counting the number of distinct unigrams and bigrams in the generated texts. Additionally, we measure Pairwise BLEU\cite{shen2019mixture} using the SacreBLEU \footnote{\url{https://github.com/mjpost/sacrebleu}}.

\begin{figure}[t!]
    \centering
    \includegraphics[width=0.95\linewidth]{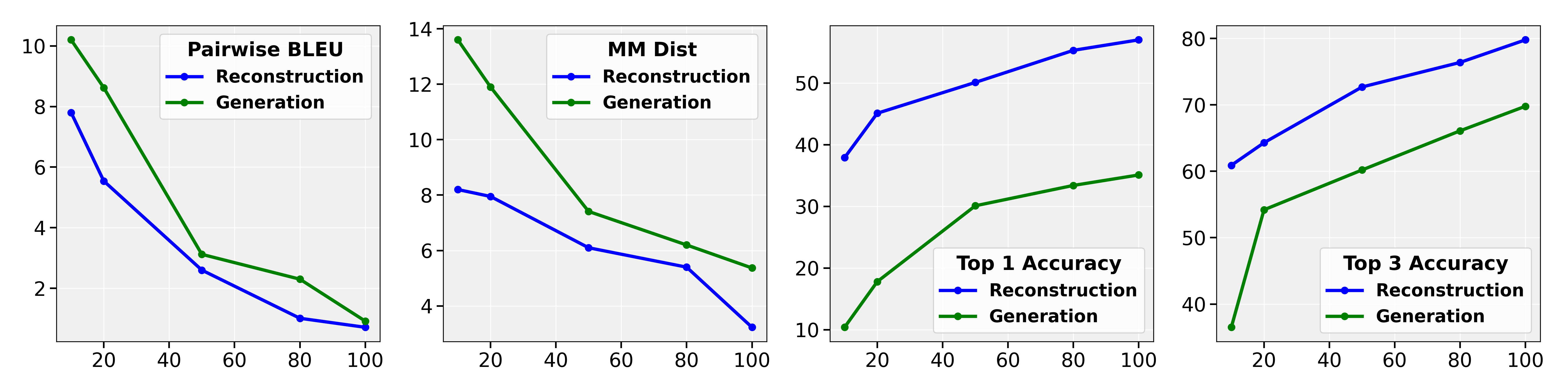}
    \caption{ Effect of dataset size on motion reconstruction and motion-to-text generation performance.
    The x-axis shows the percentage of training data used (10\%, 20\%, 50\%, 80\%, and 100\%), and the y-axis displays performance metrics: Pairwise BLEU, MM Dist., Top-1, and Top-3 accuracy. Larger datasets consistently improve performance across all metrics, highlighting the benefits of increased data scale.
    }
    \label{fig:m2t_scale}
    \vspace{-1em}
\end{figure}

\paragraph{Impact of Dataset Size.} 
We show the influence of dataset size on both motion reconstruction and text-to-motion tasks in the paper.
\Cref{fig:m2t_scale} further illustrates the impact of dataset size on motion reconstruction and motion-to-text generation. Based on the TM2T architecture, we train a motion VQ-VAE for reconstruction and a transformer-based motion-to-text model for captioning with varying proportions of the training set (10\%, 20\%, 50\%, 80\%, and 100\%), while evaluating performance on the same test set. We report Pairwise BLEU, MM Dist., and Top-1/Top-3 accuracies.

The results show consistent improvements across all tasks as the dataset size increases. Larger datasets lead to better motion-text alignment (lower MM Dist.), more diverse text generation, and improved retrieval accuracies. These findings emphasize the importance of large-scale data for enhancing performance in motion reconstruction, text-to-motion, and motion-to-text generation.

\paragraph{Ablations of Different Motion Representations and Text Annotations.}

\begin{table*}[ht!]
    \centering
    \caption{Ablation study on different hand motion representations and text annotations for motion captioning task.\emph{KP} refers to the 3D hand keypoints representation, while \emph{6D} denotes the MANO pose parameters encoded in the 6D representation. Numbers in parentheses indicate the quantity of text scripts. Upper bound indicates the metric performance calculated with the ground truth.}
    \begin{tabular}{l c c c c c c c c c c}
    \toprule
     \multirow{2}{*}{Datasets}  & \multicolumn{3}{c}{R Precision(\%)$\uparrow$} & \multirow{2}{*}{MM Dist$\downarrow$} & \multirow{2}{*}{\shortstack{Pairwise \\ BLEU}$\downarrow$} & \multirow{2}{*}{B@4$\uparrow$} & \multirow{2}{*}{ROUGE$\uparrow$}& \multirow{2}{*}{distinct-1(\%)$\uparrow$}& \multirow{2}{*}{distinct-2$\uparrow$(\%)}& \multirow{2}{*}{BScore$\uparrow$}\\

    \cline{2-4}
       ~ & @1 & @2 & @3 \\
    
    \midrule

    upper bound & 75.7 & 87.2 & 92.1 & 3.28 & - & - & - & - & - & - \\
    6D (14k) & 49.8 & 62.8 & 70.2 & 5.66 & 1.30 & 33.7 & 51.2 & 4.03 & 16.4 & 47.1 \\
    6D (84k) & 50.2 & 61.3 & 66.7 & 6.31 & \textbf{0.804} & 36.0 & 53.7 & 7.27 & 24.5 & 50.3 \\
    \hline
    upper bound & 75.3 & 89.1 & 93.9 & 2.87 & - & - & - & - & - & - \\
    KP (14k) & \textbf{57.2} & \textbf{68.9} & \textbf{74.4} & \textbf{4.69} & 1.21 & 41.3 & 56.8 & 6.28 & 21.4 & 52.4 \\
    KP (84k) & 57.0 & 66.1 & 69.8 & 5.37 & 0.916 & \textbf{43.1} & \textbf{57.7} & \textbf{15.3} & \textbf{36.9} & \textbf{55.4} \\
    
    \bottomrule
    \end{tabular}

    \label{tab:quant_m2t_ablation}

\end{table*}

We also conduct ablation study on different motion representation and number of text annotations for motion captioning task in \Cref{tab:quant_m2t_ablation}

The results indicate that the 6D representation slightly underperforms compared to the keypoint representation in the motion captioning task. However, the addition of extra text annotations significantly improves performance on linguistic metrics, boosting the diversity of generated motions. Also as the number of text annotations per motion clip increases, the retrieval accuracy for matching motions to corresponding text slightly decreases. This could be due to the fact that with more annotations, the matching process becomes more challenging, as the model may have to distinguish between a larger variety of possible descriptions for the same motion, leading to more potential mismatches in retrieval.

\paragraph{More Qualitative Results.}
\begin{figure}[htbp]
    \centering
    \includegraphics[width=0.98\linewidth]{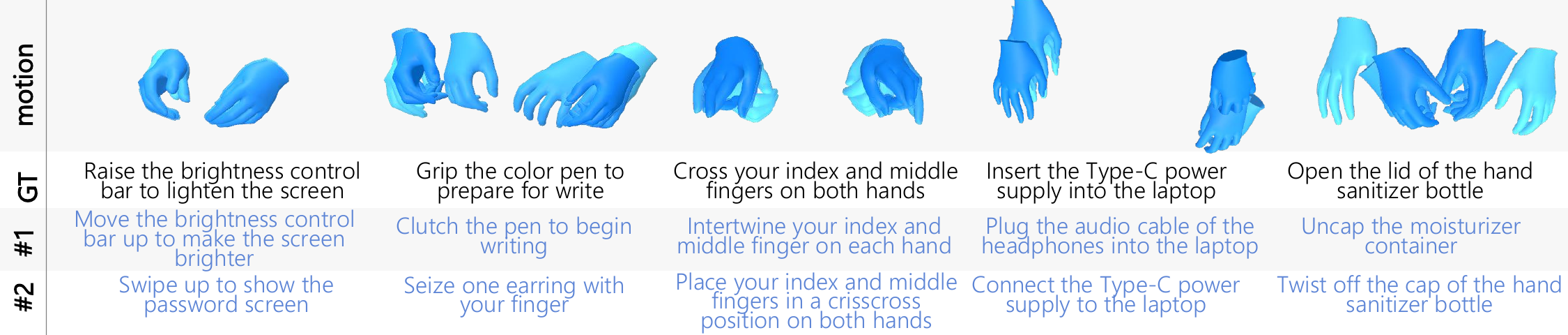}
    \includegraphics[width=0.98\linewidth]{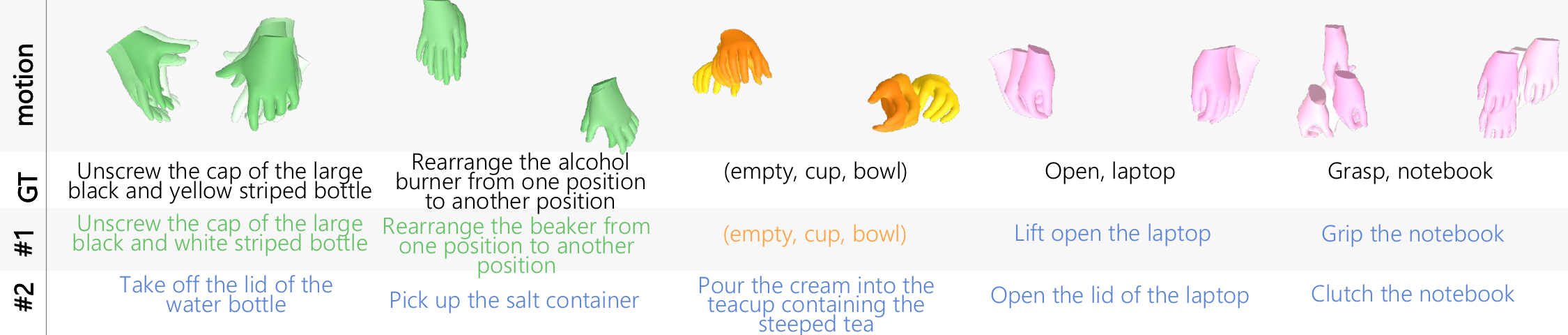}
    \caption{More qualitative results for motion caption. Each column shows a motion sequence, its ground truth text description, and two generated texts. Hand motions highlighted in \textcolor{green}{green}, \textcolor{orange}{orange}, \textcolor{pink}{pink} and \textcolor{blue}{blue} come from \textcolor{green}{OakInk2}, \textcolor{orange}{TACO}, \textcolor{pink}{Arctic} and \textcolor{blue}{\methodname}, respectively. Texts highlighted in these colors are generated by models trained on the corresponding datasets. The model trained on \textcolor{blue}{\methodname} generates diverse captions from a single motion (first row) and accurately captions motions from other datasets (second row). Darker color indicates later frame in the sequence.}
    \label{fig:more_m2t_res}
\end{figure}

We present additional qualitative results for motion captioning on our dataset and in-the-wild scenarios (\Cref{fig:more_m2t_res}). Captions in matching colors are generated by models trained on these datasets. Notably, the model on \methodname{} generates diverse descriptions from a single motion and accurately captions motions from other datasets, demonstrating its robustness and generalization ability. For the ARCTIC dataset \cite{fan2023arctic}, we apply a similar processing procedure to align orientation and motion range, ensuring consistent 3D hand keypoints. Additionally, we leverage verb annotations from \cite{cha2024text2hoi} to obtain semantically meaningful motion clips for training the combined VQ-VAE model.

\section{Experiments on Dynamic Reconstruction}

\begin{figure}[ht!]
    \centering
    \includegraphics[width=\textwidth]{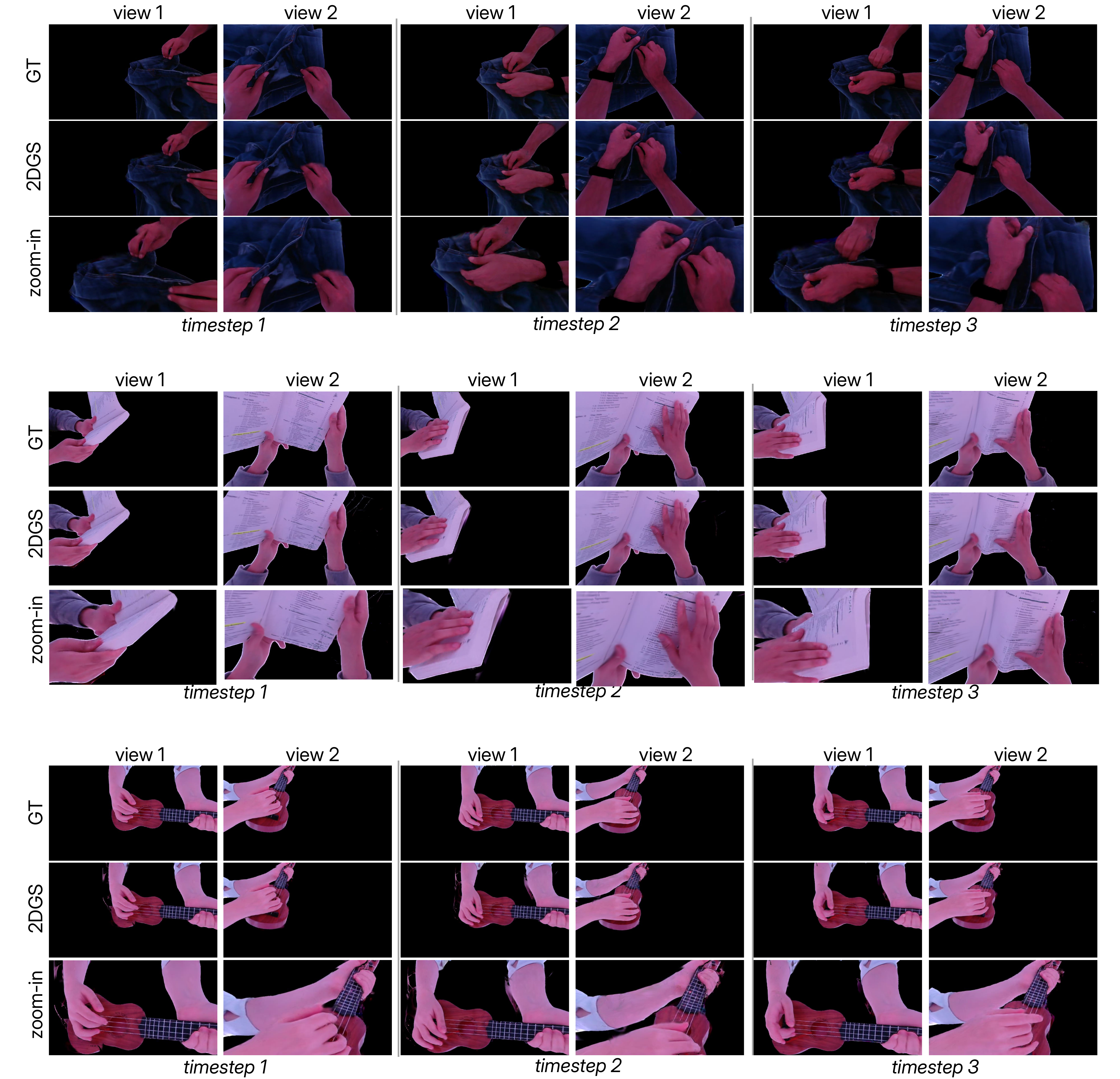}
    \caption{\textbf{Qualitative Results on Novel View Synthesis.} 
    }
    \label{fig:dynamic}
\end{figure}

\paragraph{More Qualitative Results.}
\Cref{fig:dynamic} presents additional qualitative results on novel view synthesis using \methodname{} data with 2D Gaussian Splatting~\cite{huang20242d}. Each example shows two test views at three different time steps. By utilizing 38 training views from \methodname{}, we are able to faithfully synthesize the test views, demonstrating the effectiveness of our method in capturing complex hand motions and interactions.

\paragraph{Quantitative Evaluations.}
\begin{table}[ht]
\centering
\begin{tabular}{lll}
\hline
PSNR  & SSIM  & LPIPS \\
\hline
29.50 & 0.963 & 0.063 \\
\hline
\end{tabular}
\caption{Quantitative Evaluation of Image Quality on Synthesized Views.}
\label{tab:dynamic}
\end{table}

\Cref{tab:dynamic} provides quantitative evaluations of the synthesized test views. We measure the rendering quality for 15 randomly selected motion sequences from \methodname{} using Peak Signal-to-Noise Ratio (PSNR), Structural Similarity Index Measure (SSIM)\cite{sara2019image}, and Learned Perceptual Image Patch Similarity (LPIPS)\cite{zhang2018unreasonable}. These metrics confirm the high fidelity and perceptual quality of our synthesized views, highlighting the effectiveness of our dataset in novel view synthesis.

\newpage

\section{Experiments on Hand-Object Interaction}
\begin{table}[t!]
    \caption{Quantitative results
    for text-driven motion synthesis with models trained on different datasets. \emph{upper bound} indicates performance calculated with the ground truth. We report the mean of 20 evaluations, and $\rightarrow$ means the closer to the upper bound the better. The model trained on \methodname performs best on most metrics.}
    \vspace{-1em}
    \centering
    \begin{tabular}{l c c c c c c c}
    \toprule
     \multirow{2}{*}{Dataset}  & \multicolumn{3}{c}{R Precision(\%)$\uparrow$} & \multirow{2}{*}{MM Dist.$\downarrow$} & \multirow{2}{*}{FID$\downarrow$} & \multirow{2}{*}{Div.$\rightarrow$} & \multirow{2}{*}{MM.$\uparrow$} \\

    \cline{2-4}
       ~ & @1 & @2 & @3 \\
    
    \midrule
    {upper bound} & 50.4 & 71.2 & 81.1 & 3.67 & 0.022 & 9.30 & - \\
    OakInk2 \cite{zhan2024oakink2} & 23.4 & 35.7 & 49.8 & 7.41 & 13.1 & 6.24 & 3.71 \\
    \hline
    {upper bound} & 77.4 & 88.8 & 91.3 & 2.96 & 0.002 & 11.9 & - \\
    \methodname{} & \textbf{27.2} & \textbf{46.2} & \textbf{54.6} & \textbf{6.12} & \textbf{5.91} & \textbf{10.2} & \textbf{9.73} \\
    \bottomrule
    \end{tabular}

    \label{tab:quant_tamf}
\end{table}

To assess the effectiveness of our dataset in modeling hand-object interactions, we evaluate its applicability across two object-related tasks inspired by prior benchmarks in Oakink and TACO. GigaHands provides rich and diverse annotations of objects and their interactions with human hands, making it a valuable resource for learning and predicting hand-object dynamics. The selected tasks aim to demonstrate the advantages of GigaHands in capturing fine-grained motion patterns and enhancing downstream applications.

\paragraph{Task-aware Motion Fulfillment.}  The Task-aware Motion Fulfillment task aims to generate realistic hand motion sequences that align with predefined object trajectories while adhering to a given textual task description.
To achieve this, we adapt the MDM baseline model \cite{tevet2023human} by incorporating dual-conditioning on both textual instructions and object trajectories. Given a task description and a sequence of object movements, the model predicts a corresponding sequence of hand motions that naturally interact with the object.
For evaluation, we employ feature extractors and metrics used in text-to-motion tasks above. Table \ref{tab:quant_tamf} provides a quantitative comparison between our approach and Oakink2. Our method outperforms Oakink2 across all evaluated metrics, highlighting its effectiveness in generating task-aware hand-object interactions.

\begin{table}[h]
    \centering
    \begin{tabular}{lccc}
        \toprule
        Method & $J_e (mm, \downarrow)$ & $T_e(mm, \downarrow)$ & $R_e(^\circ, \downarrow)$ \\
        \midrule
        TACO & 71.7 / 58.4 & 52.8 & 73.2 \\
        GigaHands & 69.3 / 62.3 & 47.6 & 67.5 \\
        \bottomrule
    \end{tabular}
    \caption{Quantitative comparison of hand-object motion forecasting across different datasets. The evaluation metrics include Mean Per Joint Position Error ($J_e$) for right / left hand pose prediction, translation error ($T_e$) and rotation error ($R_e$) for object motion. Lower values indicate better performance.}
    \label{tab:quant_homf}
\end{table}

\paragraph{Generalizable Hand-object Motion Forecasting.} Hand-object motion forecasting aims to predict future hand and object motions based on a short observed sequence. Given the poses of both hands and objects over $N$ consecutive frames, the objective is to forecast their poses over the subsequent $M$ frames. In our experiments, we set $N=10$ and $M=10$
We adapt the MDM model by conditioning it on the past N frames of hand and object poses to predict their future states over the next M frames. Following human-object forecasting evaluations \cite{xu2023interdiff, liu2024taco}, we access hand pose estimation using Mean Per Joint Position Error $J_e$, while object motion is evaluated with translation error $T_e$ and rotation error $R_e$.  For TACO’s triplet representation, we condition only on tool poses and compute the corresponding metrics. Table \ref{tab:quant_homf} presents a comparative analysis of models trained on different datasets. While both models achieve comparable performance in hand pose prediction, our approach outperforms TACO in object motion forecasting. However, hand-object motion forecasting still remains a challenging task. The inherent complexity and rapid variations in generative motion patterns make modeling motion distributions difficult. Future work could explore more effective strategies to address this challenge.

\section{Hand Motion Tracking}

\begin{figure}[ht!]
    \centering
    \includegraphics[width=\textwidth]{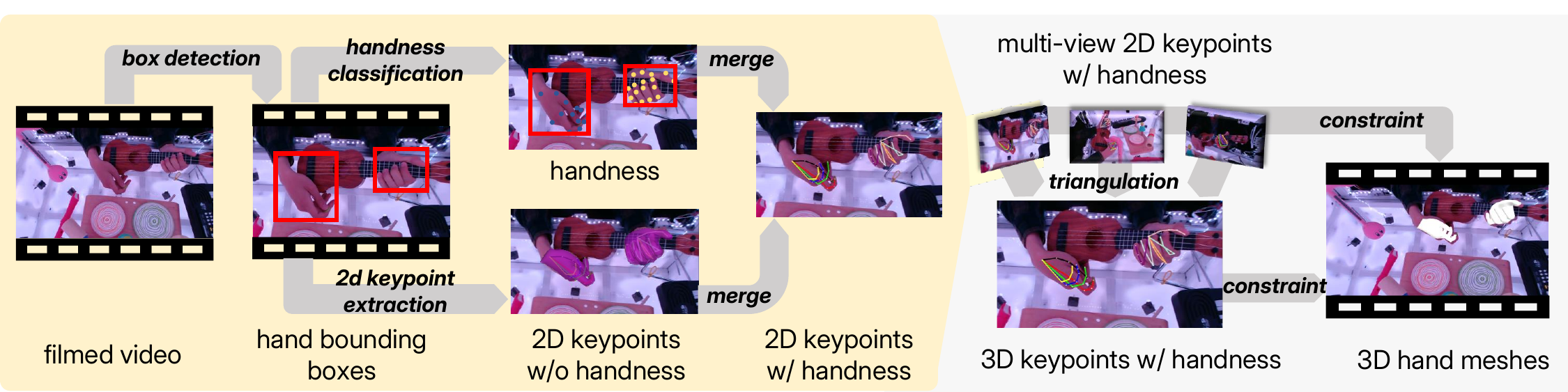}
    \caption{\textbf{Hand Motion Tracking Pipeline.} 
    }
    \label{fig:hand_tracking}
\end{figure}

\begin{table}[h]
\centering
\begin{tabular}{cc|cc}
\hline 
box detection                  & 2D keypoints detection  & time(150frames) & valid rate \\
\hline 
detectron2           & VITPose & 6.5min         & 91.7\%     \\
yolov9c/track        & VITPose & 3.3min         & 89.6\%     \\
yolov9c/detect batch & VITPose & 1.1min         & 85.4\%     \\
yolov9c/detect batch & HaMeR   & 3.0min         & 97.9\%    \\
\hline 
\end{tabular}
\caption{Comparison of different bounding box detection and 2D keypoint estimation methods.}
\label{tab:hand_evaluation}
\end{table}

\Cref{fig:hand_tracking} illustrates the complete pipeline for hand motion tracking. The process begins with view-wise hand detection, tracking, handedness classification, and 2D keypoint extraction. Using multi-view 2D keypoints for both hands, we then triangulate each hand separately to obtain 3D hand keypoints. Finally, with the 2D keypoints and triangulated 3D keypoints, we fit the MANO~\cite{romero2022embodied} parameters under these constraints. To validate our choices for hand detection, handedness classification, and 2D keypoint estimation, we calculate the number of valid 3D triangulated frames. We consider a 3D frame valid if, for both hands: (1) there are no missing keypoints; (2) the hand kinematics are normal, with bone length variations across frames within a certain threshold; and (3) the root of the hand moves temporally consistently. Frames that do not meet these criteria are considered invalid 3D triangulated keypoints.\Cref{tab:hand_evaluation} compares the valid frame rate among 60 motion clips and the inference time for processing 150 frames (5 seconds).

\paragraph{Hand Detection and Tracking.}
For hand detection, we choose YOLO-v9~\cite{reis2023real} as the backbone because it provides reliable detection results, runs efficiently compared to other backbones such as Detectron2\cite{wu2019detectron2}, and yields consistent temporal results. Instead of using box tracking, we batch frames across multiple time steps (up to 256 images) and apply the detection function simultaneously, which accelerates inference time without significantly reducing the valid rate. During the filming process, we instructed subjects to keep both of their hands within the scene. Therefore, by default, we extract the two most confident bounding boxes labeled as 'hand' from the detection results.

\paragraph{Handness (left or right).}
To determine the handedness (left or right) of each detected hand, we follow the method adopted in HaMeR~\cite{pavlakos2024reconstructing}. We use a side-aware checkpoint of ViTPose to detect keypoints within each bounding box. We classify a hand as 'right' if more than 60\% of the detected keypoints correspond to the right hand, and 'left' if they correspond to the left hand.

\paragraph{2D Keypoint Extraction.}
We use HaMeR\cite{pavlakos2024reconstructing} for 2D keypoint extraction for two main reasons. First, it estimates a parametric representation of the hand from a cropped image, ensuring that all keypoints can be extracted from the output. Second, it provides more reliable results that ensure the hand kinematics are reasonable. We believe this is the primary reason that keypoints extracted from HaMeR lead to the most valid triangulated 3D keypoints. Although using HaMeR decreases inference speed, it significantly improves the valid rate.

\paragraph{3D Keypoint Triangulation.}
With the camera parameters and 2D keypoints extracted from multiple views, we triangulate the 3D keypoints for each hand. We remove outliers using RANSAC to improve accuracy.

\paragraph{Parameter Fitting. }
For MANO parameter fitting, we follow the EasyMoCap~\cite{easymocap} pipeline, using both 2D and 3D keypoints as supervision. To capture fine-grained finger motions, we disable the PCA components and use a flat mean shape during the fitting process. This approach allows for more detailed and accurate hand mesh reconstruction.

\section{Object Motion Tracking}

\begin{figure}[ht!]
    \centering
    \includegraphics[width=\textwidth]{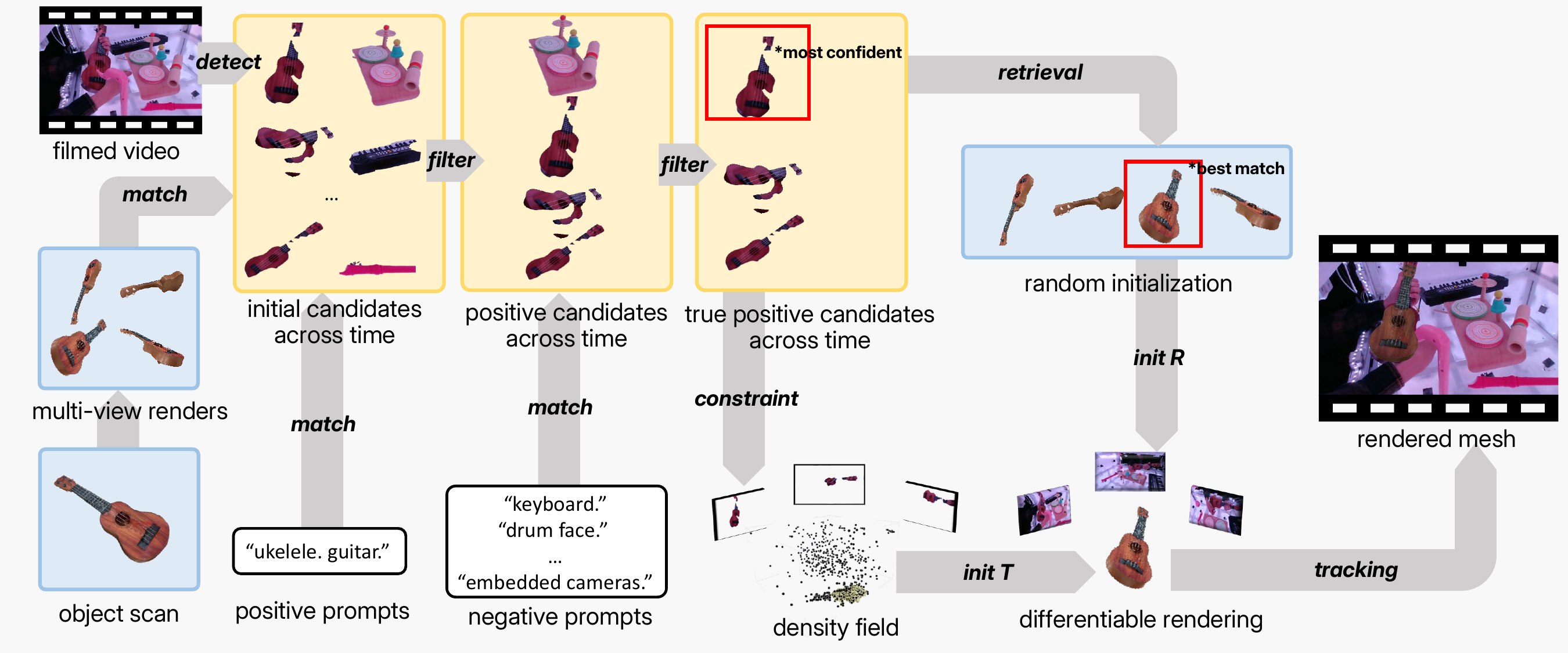}
    \caption{\textbf{Hand Motion Tracking Pipeline.} 
    }
    \label{fig:object_tracking}
\end{figure}

\begin{table}[h!]
\centering
\begin{tabular}{c|c}
\hline
stage                         & mask coverage \\
\hline
coarse estimation             & 45.2\%         \\
pose refinement (first frame) & 91.1\%         \\
pose refinement (sequence)    & 78.5\%        \\
\hline
\end{tabular}
\caption{Evaluation of Mask Coverage Rates Across Different Stages of Object Motion Tracking.}
\label{tab:object_tracking}
\end{table}

The tracking process aims to estimate the 6DoF (six degrees of freedom) pose of the pre-scanned or generated mesh over time using multi-view RGB videos as constraints. Since our capturing system only contains RGB information without depth, pose estimation poses significant challenges. \Cref{fig:object_tracking} illustrates the complete object tracking pipeline. However, this can be mitigated by utilizing 51 camera views that provide 360-degree coverage of the scene. The tracking process generally consists of three stages: (1) extracting object masks across views as constraints; (2) providing a coarse pose estimation for the mesh as an initialization for the following refinement process; (3) refining the object pose using multi-view pixel-level constraints to achieve accurate pose estimation. We use the mask coverage rate as an evaluation metric for tracking quality. The mask coverage rate is defined as the percentage of true positive segmented masks across frames that overlap with the rendered object's silhouette during differentiable rendering. \Cref{tab:object_tracking} provides this evaluation across different stages.

\paragraph{Object Segmentation.}
To use the multi-view RGB videos as constraints, accurate object masks are necessary. During the mask extraction process, we aim for three objectives: (1) obtaining masks from as many views as possible to guarantee more accurate results; (2) avoiding false positives in the segmentation masks, as they can be very distracting; (3) ensuring the masks are consistent across views to provide stable supervision. To achieve the first goal, for each view, we use DINOv2~\cite{oquab2023dinov2} to detect as many object candidates as possible across time steps. We detect objects across time steps because an object might only appear during certain time period throughout the video. With multiple candidates that cover all objects across time steps, we address the second goal by using both image and text prompts to identify true positive candidates through feature matching methods using CLIP~\cite{o2023open} and Grounding DINO~\cite{liu2023grounding}. Image prompts are generated by rendering object scans from random views on a sphere, while text positive prompts are manually designed. To remove false positives, we design negative prompts and compare them with the positive candidates. If an object candidate is more similar to the negative prompt than to the positive prompt, we regard it as a false positive. From all the true positive candidates, we select the most confident one using text-image similarity evaluated by CLIP~\cite{o2023open}. With the most confident candidate, we can track the object across the video using SAM2~\cite{ravi2024sam}.

\paragraph{Coarse Pose Estimation}
Given the segmented object masks across views, we can initialize the coarse pose of the object. We first find the time step with the largest summed mask area of views. We use the multi-view masks at this time step to reconstruct the density field using Instant-NGP~\cite{muller2022instant}. We threshold the reconstructed density field and find the largest cluster inside the field, shown as the yellow points in \Cref{fig:object_tracking}. We then use the center of this density field as the coarse translation of the object. If the density field reconstruction fails, we use the center of the capture system with five random offsets as the coarse translations. Since the reconstructed density field is noisy, we cannot reliably use ICP (Iterative Closest Point) to obtain the coarse rotation. Additionally, the geometric symmetry of the meshes requires us to use appearance information to initialize the rotation. We follow FoundPose~\cite{ornek2023foundpose} by randomly rotating the template mesh and retrieving the top five matched rotations using DINOv2 features. We use these as the initialization for rotation. \Cref{tab:object_tracking} row 1 reports the mask coverage of the coarse estimation at this stage.

\paragraph{Pose Refinement.}
Given the coarse estimation of translation and rotation, we refine the pose estimation using differentiable rendering supervised with multi-view silhouette loss. The refinement consists of three stages. In the first stage, we select the best coarse initialization. In the previous step, we might have multiple translation and rotation candidates due to density field reconstruction failure and ambiguity in object symmetry. Therefore, we first optimize the pose starting from multiple coarse pose candidates for 200 steps. In the second stage, we select the coarse pose candidate with the minimal silhouette loss and further optimize the pose for 500 steps. \Cref{tab:object_tracking} row 2 reports the mask coverage of the pose refinement for this frame after these two stages. In the third stage, we optimize the whole sequence, using the frame from the previous stage as initialization. Note that this time step is selected based on the maximal mask area, so it could be in the middle of a sequence. If this is the case, we need to optimize the loss temporally in both directions. For each next time step to optimize, we use its nearest optimized time step as initialization. We optimize 500 steps for each frame. \Cref{tab:object_tracking} row 3 reports the final mask coverage across the sequence. Note that the mask coverage drops compared to the first frame; this is because for sequences with fast motion or where the object is severely occluded due to manipulation, the pose might lose track and accumulate errors.

\section{Text Instructions and Annotations}
In this section, we provide the prompts and examples used to create the instruction scripts and for annotation augmentation, along with details about the annotation interface.

\lstset{
  basicstyle=\ttfamily,
  breaklines=true
}

\subsection{Prompts and Examples for scenario grouping}
In the scenario grouping phase, we start with verb pools extracted from multiple hand datasets. Our goal is to find objects corresponding to these verbs within various scenarios, enabling us to act out the associated actions. We prompt the LLM with specific instructions to generate verb-object pairs. From the LLM's outputs, we select reasonable verb-object pairs that can be filmed on a tabletop setting. Here is an example to find objects associate with the verb `beat'.
\begin{lstlisting}
Task: Using the verb "beat," generate a list of objects that can be acted upon with this verb in various scenarios. Ensure the objects are relevant and specific to the given context. 
Scenarios
1. Cooking: Identify objects or ingredients commonly associated with "beat" in culinary activities.
2. Entertainment: List objects or tools that can be "beat" in entertainment or recreational contexts.
3. Housework: Suggest items or surfaces that are "beat" during cleaning or household chores.
4. Crafting: Find materials or tools that involve "beat" as part of a creative or crafting process.
5. Office Work: Consider any metaphorical or literal uses of "beat" with objects in an office or work environment.

Requirements:
1. Ensure that each object aligns with the context of the scenario.
2. Provide a diverse and creative range of examples for each scenario.
3. Be specific about the relationship between "beat" and the object.
\end{lstlisting}

\subsection{Prompts and Examples for Scene structuring}
In the scene structuring phase, we have already associated each scene with verbs and objects, organizing them through activities. However, in the manually designed raw activity scripts, some verbs related to the activities might be missing. To address this, we provide the LLM with a prompt to add the missing verbs to the scenes. For example below, we augmented the "playing cards" part within the "Playing Monopoly, Cards, Coins, and Knucklebones" scene. After obtaining the output from the LLM, we perform a sanity check to ensure accuracy and coherence.
\begin{lstlisting}

**Refined Prompt for Playing Cards Script**

You are tasked with refining a script for playing cards. The script must follow the format: **[verb-ing]: detailed description**. Several actions are missing, and you need to incorporate them into the sequence of actions in the correct position while ensuring the script is clear, organized, and detailed enough for a hand actor to act out.

Instructions:
1. Follow the format: Each action is described with a verb ending in "-ing" and its detailed description.
2. Include the following missing actions: scatter, slide, split, swap, wave.
3. Ensure the scenario is consecutive: The sequence of actions should flow logically without gaps.
4. Provide detailed hand action descriptions: Describe how the hands move, grip, and interact with the cards for clarity and precision.
5. This script is for a hand actor to act out. The quality of your addition and organization will determine the final output.
6. Ensure the sequence of actions forms a coherent and consecutive scenario.

If you revise the script well, I will reward you $20.

---

Original Script:
Play Cards
- [Bridge Shuffling]: Hold the deck with your right hand. Use your thumb and middle finger to grip the deck on the short sides, with your index finger resting along the long edge of the deck. Do the bridge shuffle.
- [Regular Shuffling]: Shuffle the deck of cards by interweaving them with your hands, mixing them thoroughly.
- [Cutting + Dealing]: Cut the deck of cards in half using a quick motion, separating it into two smaller decks. Deal the cards to the players one by one, distributing them evenly.
- [Flipping]: Flip the top card of the deck face-up, revealing its value or suit.
- [Fanning + Checking + Sorting]: Fan out the cards in your hand, creating a spread of cards that can be easily viewed. Check the value of your cards by looking at them without revealing them to others. Sort the cards in your hand according to their suits or numerical order.
- [Drawing + Discarding]: Draw a card from the deck and add it to your hand, increasing the number of cards you hold. Discard a card from your hand by placing it face-down on a designated discard pile.
- [Collecting + Stacking]: Collect the cards from all players after a round of the game has ended. Stack the cards on top of each other, forming a neat pile.

---


\end{lstlisting}

\subsection{Prompts and Examples for instruction scripting}
In the instruction scripting phase, we already have an activity script for each scene, where each activity is associated with a list of verbs that occur sequentially. Our goal here is to expand each verb into a complete instruction that helps fulfill the activity. We provide the LLM with a prompt to achieve this expansion. For instance below, we applied this process to an activity in the "Making and Drinking Tea" scene. After receiving the output from the LLM, we conducted three rounds of sanity checks to ensure the instructions were accurate and coherent.

\begin{lstlisting}

You are tasked with refining and structurizing the given hand action script. The final script must follow these guidelines:

1. Format: Each action must follow the format `[verb-ing]: detailed description of the action.`
2. Retain All Verbs: Do not delete or remove any verbs provided in the brackets.
3. Separate Multiple Verbs: If a bracket contains multiple verbs, split them into individual actions, each with its own description.
4. Expand Missing Verbs: If a verb is implied but not explicitly described in the action, add it with an accurate and detailed description.
5. Verb Format: Change all verbs into the present participle format (`-ing` form).
6. Clarity and Detail: Ensure the descriptions are clear, precise, and detailed enough for a hand actor to perform the actions.

---

Example:
Original Action:
`[grip, set, lift]: Grip the teapot lid with the right hand, lift it, and set it aside.`

Refined Actions:
- [Gripping]: Grip the teapot lid firmly with the right hand.  
- [Lifting]: Lift the teapot lid straight up, keeping it steady.  
- [Setting]: Set the teapot lid down gently on a flat surface.  
\end{lstlisting}

\subsection{Annotation Interface}
\begin{figure}[ht!]
    \centering
    \includegraphics[width=\textwidth]{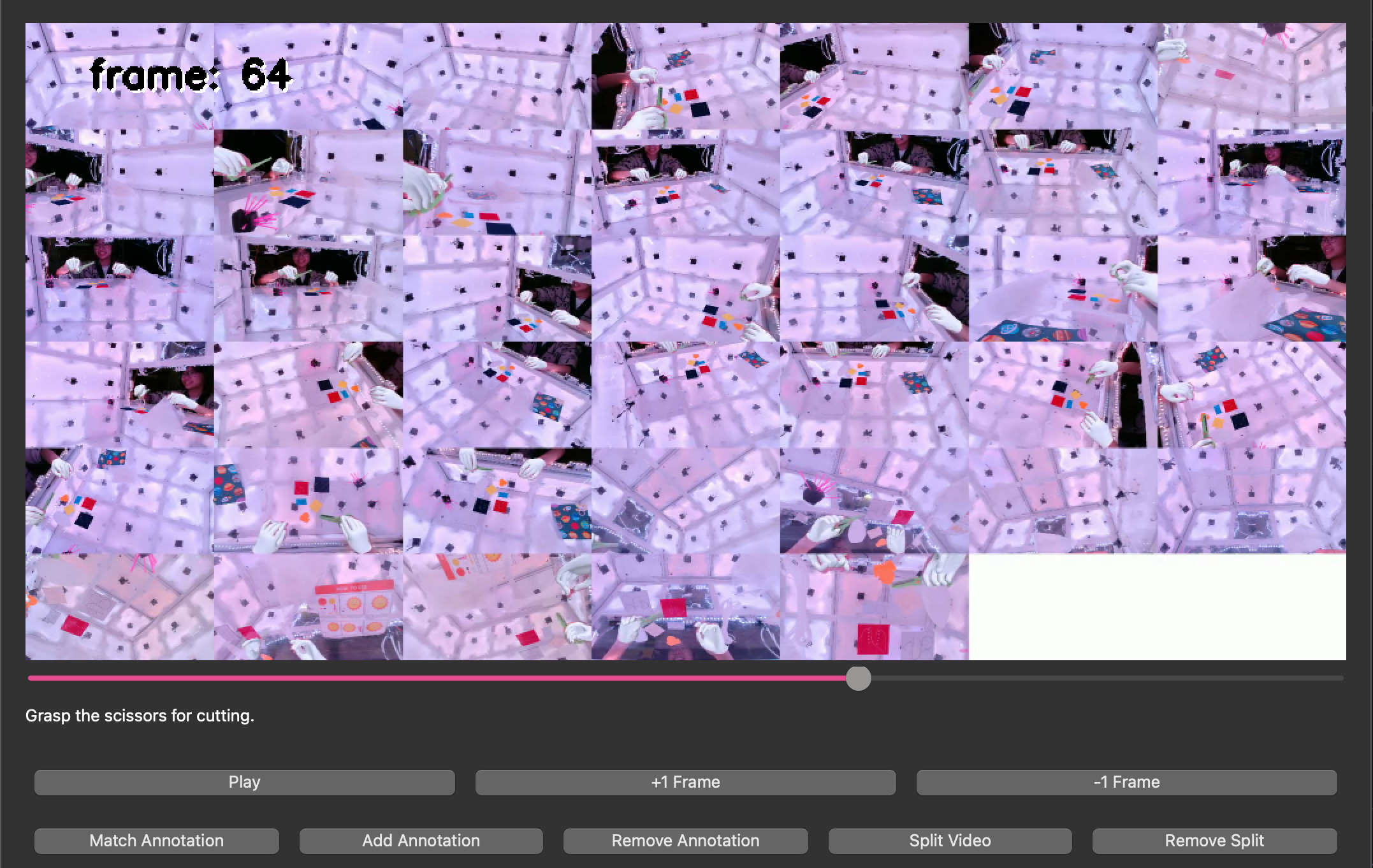}
    \caption{\textbf{Annotation Interface.} 
    }
    \label{fig:annotation_interface}
\end{figure}

For the annotation phase, we already have the instruction scripts and the motion sequences filmed accordingly. Our task is to annotate any actions that were not mentioned in the original instruction scripts. \Cref{fig:annotation_interface} illustrates the annotation interface we used. This interface loads the filmed multi-view videos with the hand mesh rendered on top for visual clarity. The original instruction is displayed below the video. Features such as a progress bar and controls like ``Play," ``+1 Frame," and ``-1 Frame" buttons are provided to accurately segment the motion sequence. Additionally, buttons like ``Match Annotation," ``Add Annotation," ``Remove Annotation," ``Split Video," and ``Remove Split" are available to correct or refine the original instructions.

\subsection{Prompts and Examples for text augmentation}
In the text augmentation phase, we have the motion clips along with their text annotations. To enhance the diversity of descriptions, we aim to augment these annotations since one action can be described in multiple ways. We feed the LLM with a specific prompt for text augmentation. Outputs such as system errors or phrases like ``I'm sorry'' are removed to maintain data quality and consistency.
\begin{lstlisting}
You are a sentence rewriter. Your task is to rewrite the provided input sentence 
into five different variations. You can vary the verbs and descriptions but 
**do not add any new actions or change the original meaning**.

Instructions:
- Begin the output with: "Rewritten Sentences:"
- Each sentence should be separated by the symbol "$"
- Use different verbs or phrasing to achieve natural, varied expressions 
  without changing the action or intent of the original sentence.

Example:
Input: "Take out an egg from the carton.\n"
Output: Rewritten Sentences: Remove an egg from the carton. $ Pick out an egg from the carton. $ Pull an egg from the carton. $ Grab an egg from the carton. $ Lift an egg from the carton.
\end{lstlisting}

\section{Dataset Inspection}

\subsection{Object Visualization.}
\begin{figure}[ht!]
    \centering
    \includegraphics[width=0.60\textwidth]{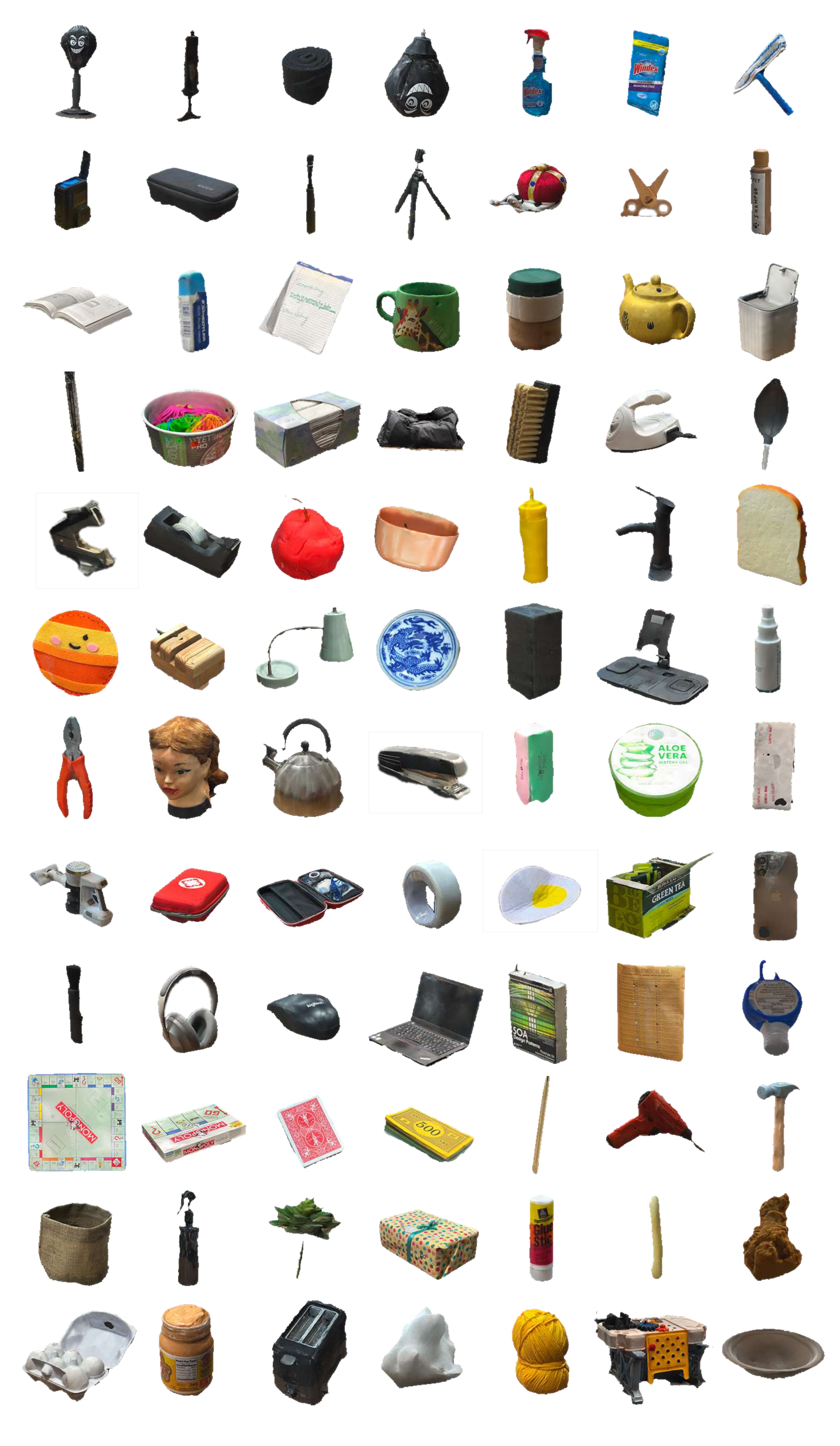}
    \caption{\textbf{Randomly Sampled Objects.} 
    }
    \label{fig:random_object}
\end{figure}

\Cref{fig:random_object} shows 96 of the objects in \methodname by randomly select a few objects each scene. The objects spans diverse scenarios and functionalities. 

\newpage
\subsection{Verb Pool.}
\methodname contains a total of 1467 verbs. \Cref{tab:most_verb}
and \cref{tab:least_verb} shows the most frequent and least frequent verbs in the text annotation.  

\begin{table}[ht]
\centering
\begin{tabular}{c | c | c | c | c | c | c | c}
 \hline
place & take & use & put & press & set & open & remove \\
 \hline
position & move & hold & pull & seize & apply & pick & lift \\
 \hline
secure & slide & grab & lay & turn & grasp & tap & push \\
 \hline
extract & clutch & insert & shift & detach & fasten & adjust & grip \\
 \hline
rest & twist & rotate & arrange & snatch & hit & spread & return \\
 \hline
close & retrieve & employ & release & loosen & squeeze & drop & flip \\
 \hline
glide & rub & make & click & fold & attach & touch & clean \\
 \hline
cut & get & leave & raise & sweep & wipe & switch & collect \\
 \hline
shut & seal & transfer & cover & snap & draw & withdraw & keep \\
 \hline
spin & swipe & select & separate & strike & perform & shake & toss \\
 \hline
bring & utilize & pour & extend & activate & compress & stretch & smooth \\
 \hline
pinch & ease & execute & unseal & pluck & tighten & lock & dab \\
 \hline
gather & uncap & throw & scoop & roll & break & create & hoist \\
 \hline
give & brush & flick & change & reduce & wrap & elevate & fill \\
 \hline
trim & clasp & work & engage & distribute & swivel & slice & increase \\
 \hline
drag & deposit & slip & fetch & unfold & split & let & flatten \\
 \hline
pat & pass & fit & add & divide & bend & peel & dispense \\
 \hline
decrease & mix & deliver & connect & agitate & trace & organize & choose \\
 \hline
undo & stick & form & guide & gesture & clip & continue & swing \\
 \hline
play & blend & disconnect & tear & stir & wiggle & carry & tilt \\
 \hline
alter & direct & affix & coat & run & modify & unzip & uncover \\
 \hline
clench & dry & shape & present & mark & maintain & depress & sketch \\
 \hline
slow & massage & speed & lessen & replace & rip & wind & join \\
 \hline
reach & submerge & unplug & wave & capture & immerse & bind & reveal \\
 \hline
point & crush & accelerate & diminish & fix & swirl & straighten & navigate \\
 \hline
snip & carve & ensure & restore & free & strip & combine & unroll \\
 \hline
obtain & handle & store & tie & wash & sway & repeat & strum \\
 \hline
intensify & lower & disengage & aim & plug & chop & quicken & unwind \\
 \hline
catch & jiggle & reverse & go & thread & revolve & stroke & identify \\
 \hline
display & pound & jostle & render & highlight & coil & knock & pretend \\
 \hline
stack & relocate & scroll & untangle & rinse & show & amplify & clap \\
 \hline
beat & start & enhance & wrench & spoon & tidy & align & pop \\
 \hline
zip & acquire & unravel & shave & punch & widen & indicate & enter \\
 \hline
write & dump & assemble & remover & launch & loop & knot & crack \\
 \hline
curl & encircle & weave & plunge & do & opt & soak & suspend \\
 \hline
enclose & lengthen & examine & drizzle & dispose & fle & sprinkle & pump \\
 \hline
expand & unhook & grind & send & build & clamp & nudge & smack \\
 \hline
relax & operate & stow & pack & hasten & tickle & shear & buckle \\
 \hline
interlock & whisk & act & dunk & lean & stop & hurl & shuffle \\
 \hline
hand & caress & flex & envelop & slot & drive & shoot &  introduce\\
 \hline
\end{tabular}
\caption{Most Frequent 320 Verbs.}
\label{tab:most_verb}
\end{table}

\begin{table}[ht]
\centering
\begin{tabular}{c | c | c | c | c | c | c | c}
 \hline
hydrate & nourish & gain & closed & travel & document & interchange & submit \\
 \hline
disorganize & investigate & convert & decorate & opener & imperfect & realign & be \\
 \hline
disarrange & toggle & tab & purify & enfold & seek & tension & container \\
 \hline
crisscross & come & systematize & chuck & tangle & slurp & bunch & relay \\
 \hline
stem & filter & wield & impact & protect & achieve & slam & narrow \\
 \hline
remain & dish & null & trickle & segregate & distinguish & rely & assist \\
 \hline
disjoin & fuse & fish & disintegrate & heave & patch & file & waver \\
 \hline
buff & stab & excise & shoelace & swell & practice & uncrease & notch \\
 \hline
avoid & estimate & confirm & probe & unlink & disassemble & underline & unbend \\
 \hline
orient & crash & haul & redo & enact & tense & discharge & disburse \\
 \hline
amass & annotate & state & incorporate & fire & rid & reproduce & modulate \\
 \hline
diffuse & collide & assort & invite & bash & beetle & blower & wound \\
 \hline
lash & mount & swish & actuate & doodle & help & suck & absorb \\
 \hline
center & revert & need & ascertain & demonstrate & lodge & deflate & shove \\
 \hline
delve & repack & stuff & wriggle & cascade & stri & imbibe & encapsulate \\
 \hline
freshen & back & scrap & sieve & screen & ignite & ajar & cool \\
 \hline
swinge & defend & exhale & trill & uptick & augment & cuddle & inspire \\
 \hline
exame & belt & srew & sit & quench & nil & isolate & copy \\
 \hline
overcross & authenticate & reaccess & register & poke & flood & calm & remit \\
 \hline
burrow & bottle & engrave & glaze & speckle & jitter & strain & stash \\
 \hline
encompass & commence & erase & structure & delineate & envelope & pamper & neaten \\
 \hline
mention & designate & jump & muster & signify & aid & distill & swarm \\
 \hline
overflow & infest & glitch & unblock & rise & weigh & hover & amp \\
 \hline
categorize & establish & hinge & emphasize & specify & assert & rebuild & seem \\
 \hline
fragment & saw & retie & conclude & finalize & scribe & acknowledge & portray \\
 \hline
tumble & perfect & reinforce & temper & define & flaunt & amalgamate & tend \\
 \hline
recompose & customize & water & irrigate & coax & uplift & magnify & grap \\
 \hline
sud & prompt & jingle & aside & leftover & peal & infuse & nest \\
 \hline
bury & quarter & mop & cap & heft & nearer & expel & dress \\
 \hline
interleave & cooperate & crawl & intermingle & rummage & briskly & hurry & float \\
 \hline
reform & nab & bake & amend & shade & unsecure & harvest & discard \\
 \hline
sample & recite & advise & grace & drum & repress & reapply & rectify \\
 \hline
hop & leverage & ring & substitute & issue & believe & channel & object \\
 \hline
tease & crease & snuff & draft & survey & bounce & test & hack \\
 \hline
pipe & refit & prefer & style & duplicate & echo & shorn & fluctuate \\
 \hline
twitch & pressurize & toast & melt & latticework & twiddle & impress & disturb \\
 \hline
tremble & recreate & snug & envision & think & interweave & redirect & paste \\
 \hline
scald & permit & tip & interconnect & improve & experience & refill & stock \\
 \hline
resort & impart & involve & claw & pry & drap & chip & hone \\
 \hline
allot & deck & trap & juggle & clarify & retreat & unfurl & disband \\
 \hline
\end{tabular}
\caption{Least Frequent 320 Verbs.}
\label{tab:least_verb}
\end{table}

\newpage
\subsection{List of Scenarios and Scenes.}
\Cref{tab:scenes} presents all of the scenarios and scenes in \methodname.
\begin{table}[h]
\begin{tabular}{l|l}
\hline
scenarios & scenes \\
\hline
\multirow{6}{*}{Cooking, Cleaning, Eating.} & Making and Drinking Tea                                        \\
                                            & Making, Eating and Cleaning Instant Noodle                     \\
                                            & Eating and Packing Delivered Food                              \\
                                            & Making an open egg and spam sandwich                           \\
                                            & Making a Bowl of Salad                                         \\
                                            & Baking and Frosting a Bread                                    \\
                                            \hline
\multirow{6}{*}{Office Working}             & Drawing Mind Map                                               \\
                                            & Receiving and Sending a File                                   \\
                                            & Testing the laptop                                             \\
                                            & Testing Tablet DigitalPen Phone SmartWatch on Charging Station \\
                                            & Filming gestures with GoPro                                    \\
                                            & Unwrapping and Wrapping a present                              \\
                                            \hline
\multirow{4}{*}{Crafting}                   & Seal Carving and Smoothing                                     \\
                                            & Assemble and Disassemble Kids Tool Bench                       \\
                                            & Cultivating and uproot a Plant                                 \\
                                            & Sewing kit, Crocheting, Knitting and Finger Knitting           \\
                                            \hline
\multirow{4}{*}{Entertaining}               & Playing monopoly, cards, coins and knucklebones                \\
                                            & Playing mini instruments                                       \\
                                            & Desktop Boxing                                                 \\
                                            & Paying with a Puppy Dog                                        \\
                                            \hline
\multirow{5}{*}{Houseworks}                 & Applying Makeups                                               \\
                                            & Packing a gym bag                                              \\
                                            & Massaging Your Own Hands                                       \\
                                            & First Aid                                                      \\
                                            & Cleaning Glass Tabletop                                        \\
                                            \hline
\end{tabular}
\caption{All 5 Scenarios and 25 Scenes.}
\label{tab:scenes}
\end{table}

{
    \small
    \bibliographystyle{ieeenat_fullname}
    \bibliography{main}
}